\documentclass[10pt,twocolumn,letterpaper]{article}

\usepackage[pagenumbers]{iccv} % To force page numbers, e.g. for an arXiv version

\usepackage[utf8]{inputenc}
\usepackage{microtype}
\usepackage{amsmath}
\usepackage{amssymb}
\usepackage{xcolor}
\usepackage{soul}
\newcommand{\todo}[1]{\textcolor{red}{\textbf{[TODO]} #1}}
\usepackage{bm}
\usepackage{multirow}
\usepackage{colortbl}
\definecolor{gblue}{RGB}{66,133,244}
\definecolor{ggreen}{RGB}{52,168,83}
\definecolor{gyellow}{RGB}{251, 188, 5}
\definecolor{gred}{RGB}{234, 67, 53}
\definecolor{gorange}{RGB}{255,109,1}
\usepackage{float}
\usepackage{makecell}
\usepackage{array}
\usepackage{graphics}
\usepackage{graphicx}
\usepackage{comment}

\usepackage{listings}
\usepackage{pifont}% http://ctan.org/pkg/pifont
\newcommand{\rev}[1]{#1}

\definecolor{iccvblue}{rgb}{0.21,0.49,0.74}
\usepackage[pagebackref,breaklinks,colorlinks,allcolors=iccvblue]{hyperref}

\def\paperID{571} % *** Enter the Paper ID here
\def\confName{ICCV}
\def\confYear{2025}

\title{UniRes: Universal Image Restoration for \rev{Complex Degradations}}

\author{
Mo Zhou$^{1,2}$\thanks{Work done during internship at Google LLC.}
\,
Keren Ye$^1$
\,
Mauricio Delbracio$^1$
\,
Peyman Milanfar$^1$
\,
Vishal M. Patel$^2$
\,
Hossein Talebi$^1$
\\
$^1$Google
\qquad
$^2$Johns Hopkins University\\
\,\newline
}

\begin{document}

\maketitle

\begin{abstract}
% !! challgenges of image restoration in the wild -- unknown degradation mixture.
Real-world image restoration is hampered by diverse degradations stemming from varying capture conditions, capture devices and post-processing pipelines.
% !! how existing works handle the challenge? (1) simulation pipelines (2) generative prior
Existing works \rev{make improvements} through simulating those degradations and leveraging
image generative priors, however generalization to in-the-wild data remains an unresolved problem.
% !! how we handle the issue? 
In this paper, \rev{we focus on complex degradations, \emph{i.e.}, arbitrary mixtures of multiple types of known degradations, which is frequently seen in the wild.}
%In this paper, we propose that the in-the-wild degradation can be approximately decomposed into a combination of known degradations, and hence it can be dealt with a weighted combination of corresponding restoration tasks.
% !! how we handle this issue? (2) core idea and method
%Based on this, we propose a simple yet flexible diffusion-based framework to transfer the knowledge from a set of known in-domain restoration tasks to out-of-domain image restoration, through a weighted combination of latent domain diffusion predictions.
\rev{A simple yet flexible diffusion-based framework, named UniRes, is proposed
to address such degradations in an end-to-end manner.
It combines several specialized models during the diffusion sampling
steps, hence transferring the knowledge from several well-isolated 
restoration tasks to the restoration of complex in-the-wild degradations.
}
% !! characteristics of our method? (1) weight comb flexibility
%These weights can be flexibly adjusted at inference time on a per-image basis to adapt to arbitrary mixture of degradations.
%\rev{This is achieved by only leveraging existing well-isolated training data for several degradation types.}
\rev{This only requires well-isolated training data for several degradation types.}
% !! characteristics of our method? (2) extension flexibility
The framework is flexible as extensions can be added through a unified formulation, \rev{and the fidelity-quality trade-off can be adjusted through a new paradigm}.
% !! evidence and conclusion
Our proposed method is evaluated on \rev{both complex-degradation and single-degradation} image restoration datasets. Extensive qualitative and quantitative experimental results show consistent performance gain \rev{especially} for images \rev{with complex degradations}.
\end{abstract}

\begin{figure}[t]
    %\centering
    \resizebox{\columnwidth}{!}{%
    \setlength{\tabcolsep}{1pt}%
    \renewcommand{\arraystretch}{0.5}%
    \large%
    \begin{tabular}{ccccc}
    
        & Real60~\cite{supir} & DiversePhotos$\times$1 &
        DiversePhotos$\times$4 & DiversePhotos$\times$4 \\
        
        & 
        ($512p\rightarrow 512p$) &
        ($512p\rightarrow 512p$) &
        ($128p\rightarrow 512p$) &
        ($128p\rightarrow 512p$) \vspace{.2em}\\ 
        
        \rotatebox{90}{\hspace{0.15\linewidth} LQ} &
        \includegraphics[width=0.40\linewidth]{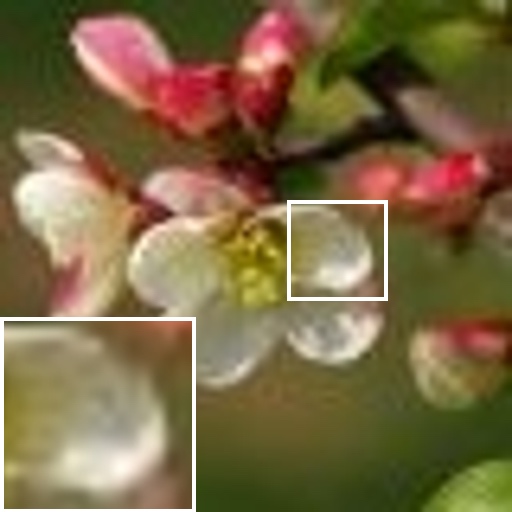} &
        \includegraphics[width=0.40\linewidth]{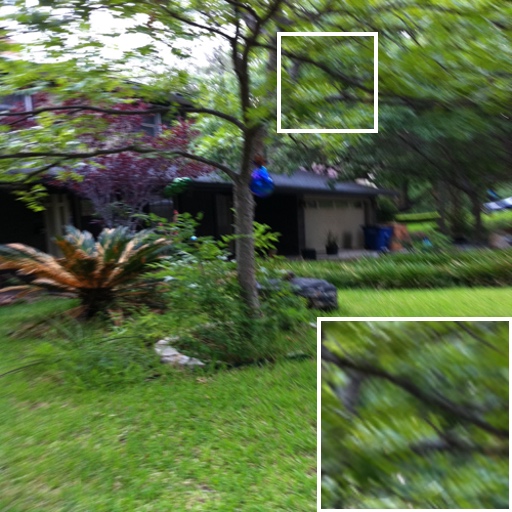} &
        \includegraphics[width=0.40\linewidth]{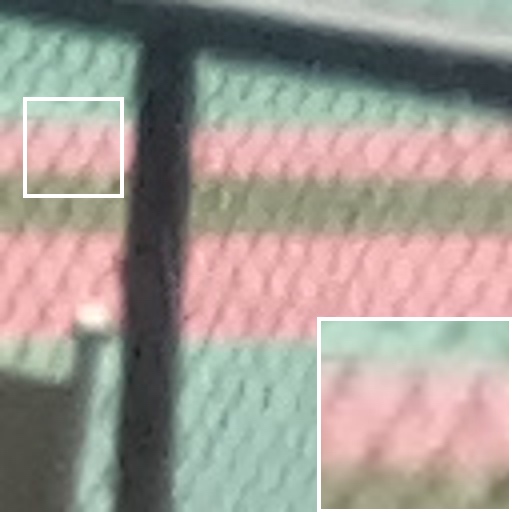} &
        \includegraphics[width=0.40\linewidth]{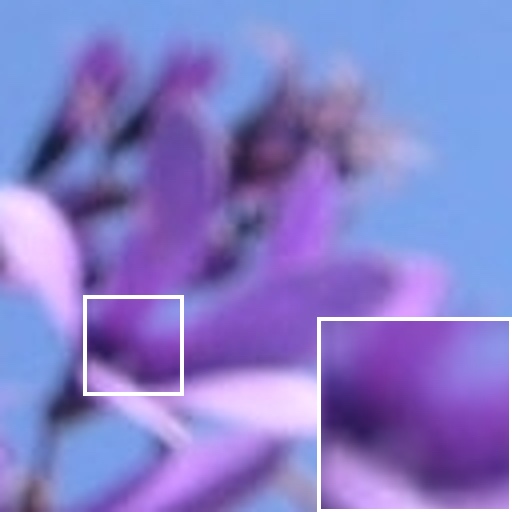} \\
        
        \rotatebox{90}{\hspace{0.04\linewidth} StableSR~\cite{stablesr}} &
        \includegraphics[width=0.40\linewidth]{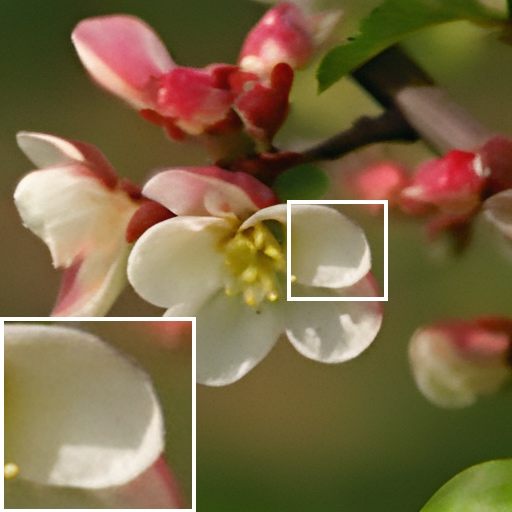} &
        \includegraphics[width=0.40\linewidth]{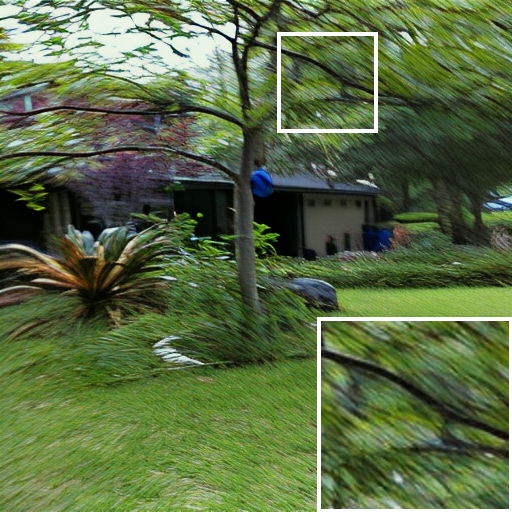} &
        \includegraphics[width=0.40\linewidth]{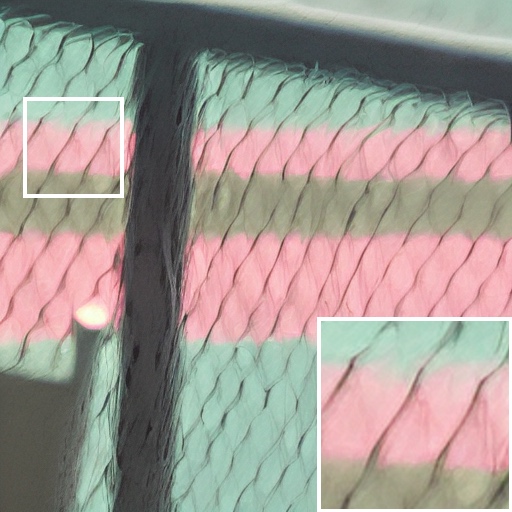} &
        \includegraphics[width=0.40\linewidth]{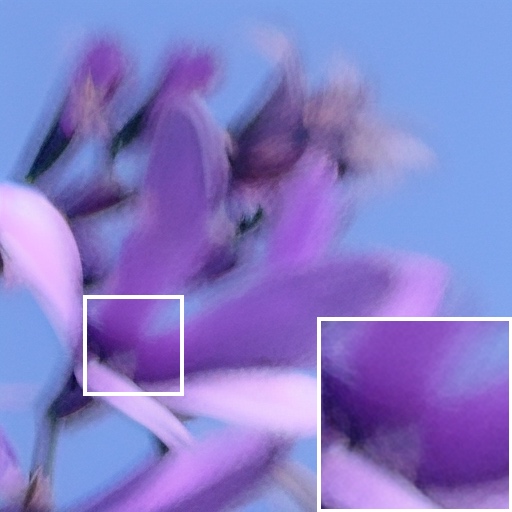} \\
        
        \rotatebox{90}{\hspace{0.05\linewidth} DiffBIR~\cite{diffbir}} &
        \includegraphics[width=0.40\linewidth]{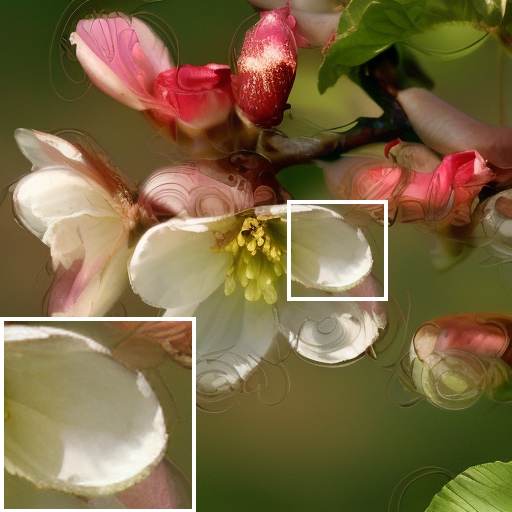} &
        \includegraphics[width=0.40\linewidth]{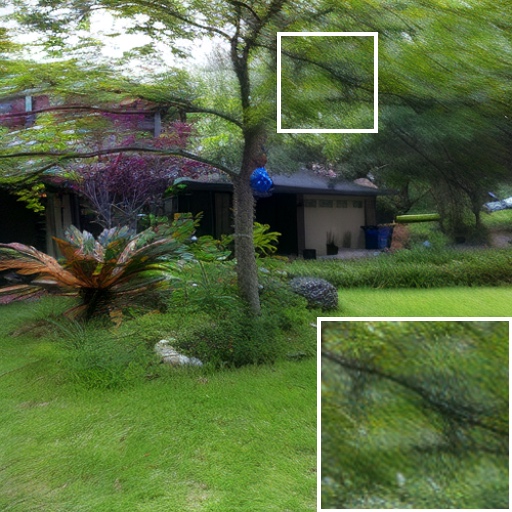} &
        \includegraphics[width=0.40\linewidth]{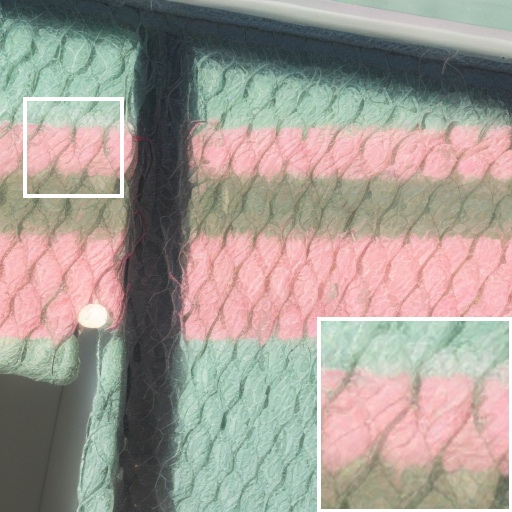} &
        \includegraphics[width=0.40\linewidth]{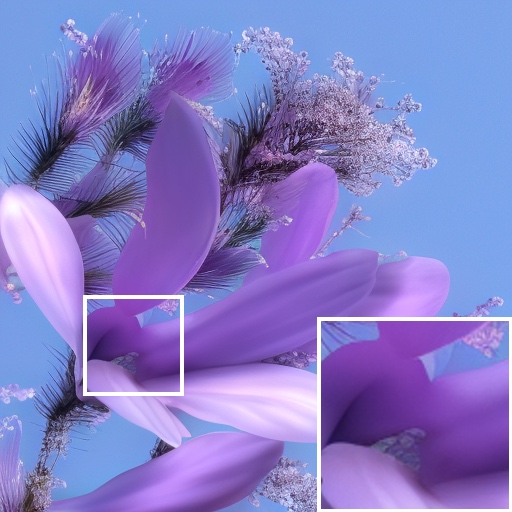} \\
        
        \rotatebox{90}{\hspace{0.08\linewidth} SUPIR~\cite{supir}} &
        \includegraphics[width=0.40\linewidth]{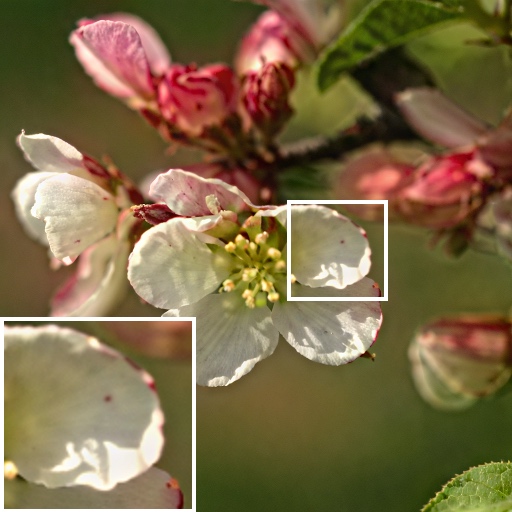} &
        \includegraphics[width=0.40\linewidth]{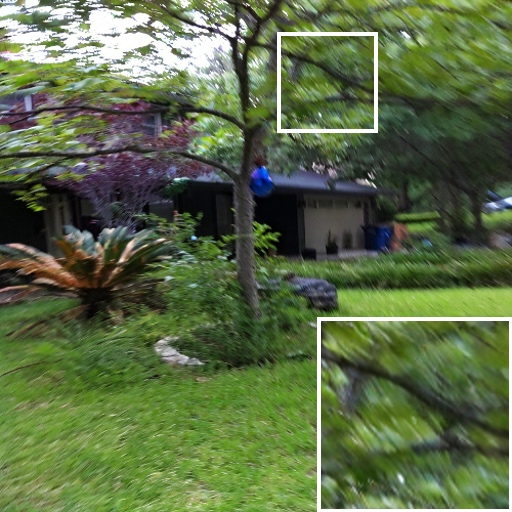} &
        \includegraphics[width=0.40\linewidth]{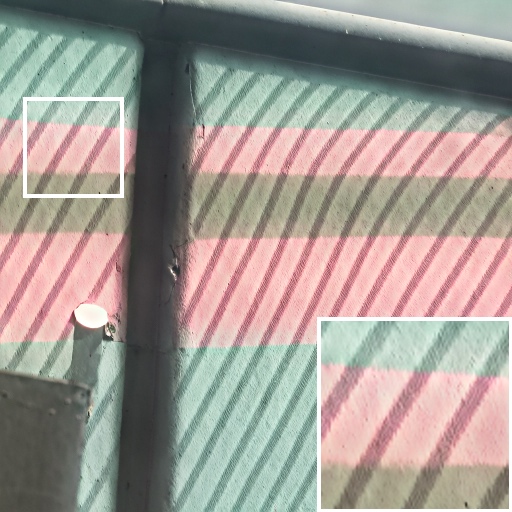} &
        \includegraphics[width=0.40\linewidth]{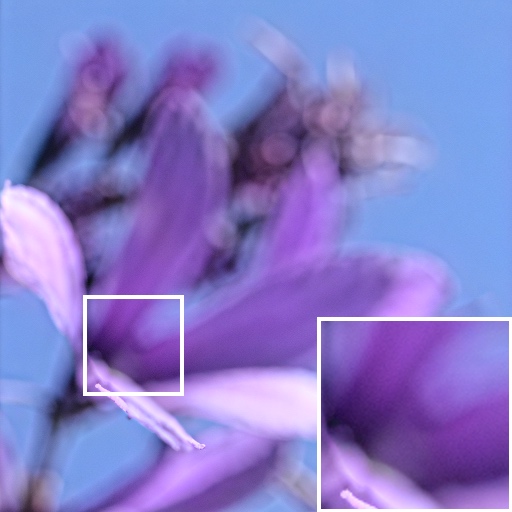} \\
        
        \rotatebox{90}{\hspace{0.02\linewidth} DACLIP-IR~\cite{daclip-uir}} &
        \includegraphics[width=0.40\linewidth]{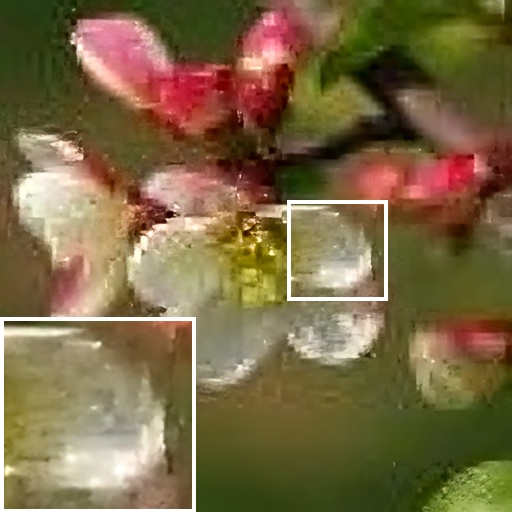} &
        \includegraphics[width=0.40\linewidth]{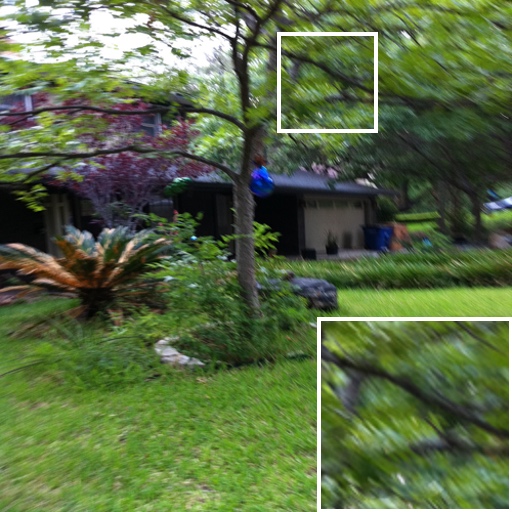} &
        \includegraphics[width=0.40\linewidth]{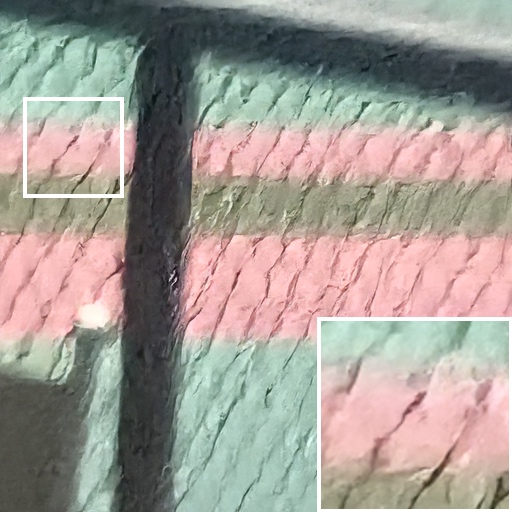} &
        \includegraphics[width=0.40\linewidth]{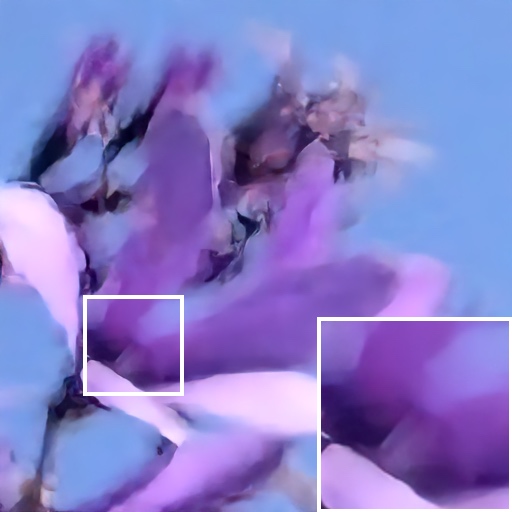} \\
        
        \rotatebox{90}{\hspace{0.14\linewidth} \textbf{Ours}} &
        \includegraphics[width=0.40\linewidth]{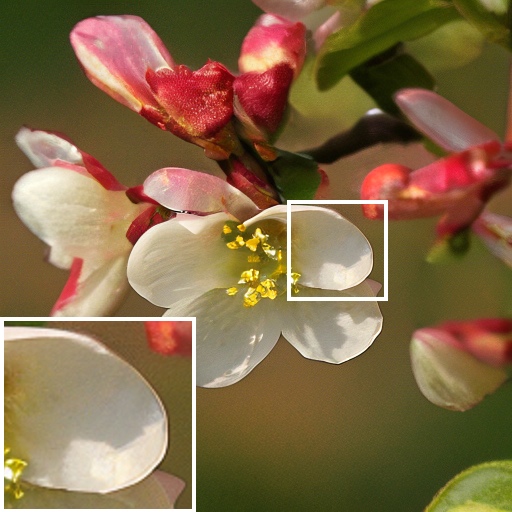} &
        \includegraphics[width=0.40\linewidth]{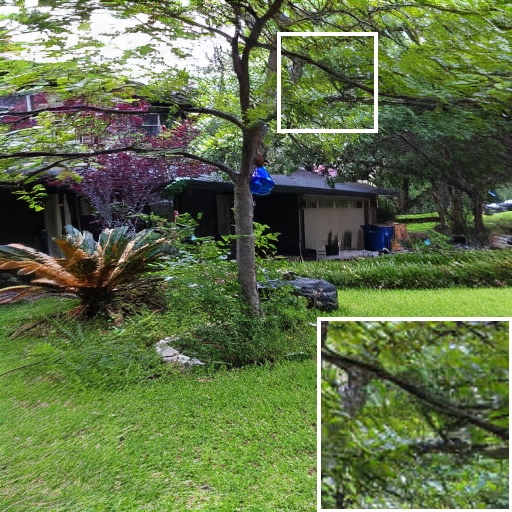} &
        \includegraphics[width=0.40\linewidth]{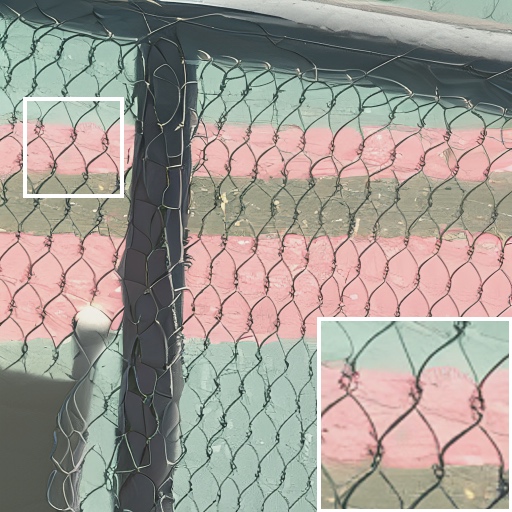} &
        \includegraphics[width=0.40\linewidth]{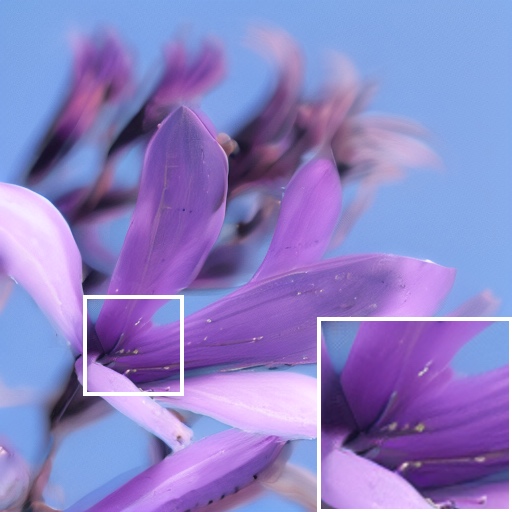}
        
    \end{tabular}}
    \vspace{-0.9em}
    \caption{Image restoration demonstration for \rev{complex} degradations.
    \rev{The ``complex degradations'' means arbitrary combinations of several
    fundamental image degradations caused by capture condition, capture device, and/or post-processing.}
    Our method is compared with several
    related state-of-the-art image restoration methods~\cite{stablesr,diffbir,supir,daclip-uir} on
    \rev{test images including} Real60~\cite{supir}, DiversePhotos$\times$1 and DiversePhotos$\times$4.
    The later two ``DiversePhotos'' test sets are challenging in-the-wild degradated images
    curated from \rev{several image quality assessment datasets}~\cite{live,spaq,koniq}, focusing
    on properly representing the real-world \rev{complex} photo degradations (see Sec.~\ref{sec:4} for detail).
    The first row of ``LQ'' image is the low-quality input image.
    %Zoom in for detail.
    }
    \vspace{-2em}
    \label{fig:teaser}
\end{figure}

\section{Introduction}
\label{sec:1}

% !! challgenges of image restoration in the wild -- unknown degradation mixture.

Real-world image restoration~\cite{stablesr,diffbir,esrgan,realesrgan,supir} is challenging due to very diverse and unpredictable degradations.
These degradations stem from various factors, including capture conditions,
capture devices,
and post-processing pipelines.
For example, object motion and slow shutter speeds
(along with camera shake~\cite{shakeblur} and the absence of vibration
reduction) can cause motion blur~\cite{gopro};
large apertures or focusing errors lead to defocus blur~\cite{dpdd};
high ISO settings produce noise~\cite{sidd};
and low JPEG quality factor for better storage leads to compression artifacts~\cite{realesrgan}.
Even worse, these degradations can co-appear \rev{on the same image as a complex degradation}.
%Such complexity make it particularly difficult to develop an effective image restoration algorithm for the real-world applications.
%This complexity makes developing effective image restoration algorithms for real-world applications particularly challenging.
\rev{This is inevitable in real-world applications, and remains very challenging to image restoration algorithms.}
%These degradations stem from various factors, including the specific conditions under which the image was captured (such as motion blur from object or camera movement), the type of capture devices used (including both hardware and software), and any subsequent post-processing pipelines (such as compression, retouching, and editing),

%
% !! how existing works handle the challenge? (1) simulation pipelines

An intuitive solution for addressing complex degradations, is to create training datasets that include pairings of high-quality (HQ) and low-quality (LQ) images. 
However, \rev{curating such image pairs with complex degradation is very difficult. On the other hand,} existing datasets are often limited by a lack of scene diversity~\cite{gopro} and realistic degradations~\cite{realsr,drealsr,rsblur,gopro}.
Recent methods employ synthetic degradation pipelines~\cite{realesrgan,daclip-uir}, typically involving Gaussian blur, Gaussian/Poisson noise, resizing, and JPEG compression.
%
% However, a significant generalization gap persists between synthetic and real degradations~\cite{sidd,realsr,gopro,dpdd}, leading to suboptimal performance of image restoration models on real-world data.
Despite advancements in degradation diversity within the training data, these methods still suffer from a significant generalization gap -- models trained on simulated data often under-perform when facing real-world images~\cite{sidd,realsr,gopro,dpdd}. This challenge urges further research to improve the generalization capabilities of image restoration models.

% !! how existing works handle the challenge? (2) generative prior

To bridge the generalization gap, recent methods leverage pre-trained image generative priors~\cite{supir,diffbir,stablesr,seesr,pasd}. These approaches operate under the assumption that a proficiently trained generative model can consistently generate clear and sharp images for reference. Numerous methods utilize pre-trained priors and fine-tune adapters to achieve blind image restoration~\cite{stablesr,diffbir,supir}, and are further boosted with multi-modal language models~\cite{llava,supir} for semantic consistency. 
However, as a side-effect of the frozen prior, adapter-based methods are susceptible to content inconsistencies and hallucination~\cite{pasd,supir,stablesr,diffbir}, where restored pixel structures differ significantly from the input.
%
% Despite progress, 
In other words, these methods remain insufficiently robust for effectively addressing the real-world \rev{complex} degradations.

% !! how we handle the issue? (1) assumption

This paper focuses on real-world image restoration,
particularly for images \rev{with complex} degradations, \rev{beyond} the ones in the training data distribution \rev{well-isolated in terms of degradation type} (see Fig.~\ref{fig:teaser}).
A real-world image \rev{may} potentially \rev{exhibit} a complex mixture of \rev{four types of} degradations \rev{casued by capture condition, capture device, and/or post-processing, including low resolution, motion blur, defocus blur,
and real noise, as aforementioned.}
%We assume this complex (out-of-domain) degradation can be approximately represented by a combination of four known (in-domain) degradation types, \emph{i.e.}, low resolution, motion blur, defocus blur, and real noise --all stemming from practical factors such as capture condition as discussed previously.
\rev{Such cases pose real challenges to existing methods~\cite{stablesr,diffbir,supir,daclip-uir} including
all-in-one restoration methods~\cite{airnet,autodir,restoreagent}.}

% !! how we handle this issue? (2) core idea and method

\rev{To tackle this challenge}, we propose a universal framework for transferring knowledge from \rev{several} known degradation restoration \rev{tasks} to handle \rev{complex} degradations in real-world images.
Our model leverages the generative prior of a pre-trained Latent Diffusion Model (LDM)~\cite{ldm}. %, drawing inspiration from works such as~\cite{stablesr,diffbir,supir,pasd,seesr}.
%
% We fine-tune it with multi-task training.
% % , with the training data representing diverse in-domain degradation restoration tasks, 
% The training data for each task is focused on a single domain. These domains include super-resolution~\cite{div2k,flickr2k,lsdir}, motion deblur~\cite{gopro}, defocus deblur~\cite{dpdd}, and real denoise~\cite{sidd}. We develop text-based prompting to guide the model to differentiate different tasks. Hence, our multi-task training is akin to co-training~[][], where a diverse set of specialized models are cultivated in an alternating fashion in a single model.
The model is enhanced by training it on multiple tasks \rev{for well-isolated degradation types} simultaneously, where each task focuses on improving a specific aspect of image quality, such as increasing resolution, removing \rev{motion or defocus} blur, or reducing noise. To help the model distinguish between these tasks, we use text prompts as guidance. This approach is similar to a co-training strategy, where we essentially cultivate a team of specialized experts within a single model, each expert excelling in a particular image enhancement task.
Our model's diffusion inference stage offers significant flexibility by combining diffusion latents from the individual expert models.
%The diffusion inference stage of our model can flexibly combine diffusion latents generated by the individual expert models.  
This adaptability allows us to effectively address real-world degradations, which may be composed of the various degradations encountered during our multi-task training.
%
% Thus, the in-domain knowledge, as represented by the latent domain diffusion predictions, can be adjusted to dynamically address various mixtures of degradations, even if out-of-domain with respect to the training data.
%
The combination weights can be determined by the user, or through optimization on a per-image basis.
%
%The proposed framework offers flexibility as the combination of in-domain knowledge, represented by the latent domain diffusion predictions, can be adjusted dynamically to address various mixtures of degradations, which are out-of-domain with respect to the training data.
%
% The framework also provides flexibility in potential extensions through its unified formulation.

% !! how to properly evaluate the effectiveness of our model

To properly represent real-world \rev{complex} image degradations, we curate ``DiversePhotos'',
% \footnote{Details for reproducing it are presented in the supplementary material.},
a set of test images originated
from SPAQ~\cite{spaq}, KONIQ~\cite{koniq}, and LIVE~\cite{live}.
\rev{Every curated image contains at least two types of real degradations.}
This collection covers a wide range of device types and real degradations,
with a balance in the number of images in each major degradation type,
effectively reflecting the diversity and complexity encountered in real-world.
\rev{To the best of our knowledge, there is no alternative
image restoration benchmark for such complex degradations.}
%
% how does it perform in evaluations
%
Extensive qualitative and quantitative experiments demonstrate the effectiveness of our  method, especially for \rev{complex degradations}.

\vspace{.2em}
\noindent The contributions of this paper are as follows:
\begin{itemize}
\item We propose \rev{UniRes}, a universal \rev{end-to-end} diffusion-based image restoration framework, targeting \rev{complex degradations}.
It can \rev{leverage} the knowledge from \rev{isolated} restoration tasks to 
\rev{address complex degradations}.
%\item 
%Our framework provides fine-grained control over the restoration process; users can adjust weights assigned to individual diffusion predictions, enabling the model to adapt to any mix of real-world degradations.
%
%Based on the framework, the latent domain diffusion predictions can be flexibly combined through arbitrary weights controlled by the user, and hence adapt to arbitrary mixture of real-world degradations.
%
\item We introduce DiversePhotos, a set of test images encompassing diverse \rev{complex} degradations, properly representing real-world image restoration challenges \rev{as a benchmark for complex degradation is missing from literature}.
\end{itemize}

\section{Related Work}
\label{sec:2}

%\noindent \textbf{Diffusion Models.} Image generation is a fundamental problem in machine learning and computer vision~\cite{ldm,muse,imagen,gan,cyclegan,guided-diffusion}.  Recently, diffusion models have emerged as a powerful approach for generating high-quality, high-resolution images~\cite{ddpm,ddim,vprediction,reprodiffusion,ldm,dit,imagen,imagen3}.  Latent Diffusion Models (LDMs)~\cite{ldm}, in particular, have been highly influential. They operate in the latent space of a Variational Autoencoder (VAE)~\cite{vae}, allowing for efficient generation, and incorporate cross-attention layers to enable conditional image generation, such as guiding the synthesis with text prompts.  Further advancements, like DiT~\cite{dit}, which replaces the U-Net architecture~\cite{unet} in LDMs with a Transformer~\cite{transformer}, have improved the performance of these models.  The success of text-to-image synthesis with diffusion models has led to their widespread adoption in various applications, including image editing~\cite{ip2p}, conditional image generation~\cite{controlnet,controlnext}, and, notably, image restoration~\cite{stablesr,diffbir,supir}.

\noindent \textbf{Diffusion Models}
%
%Image generation is a fundamental problem in machine learning and computer vision~\cite{ldm,muse,imagen,gan,cyclegan,guided-diffusion}. 
%
have emerged as a powerful approach for generating high-quality images~\cite{ddpm,ddim,vprediction,reprodiffusion,ldm,dit,imagen}.
Among these, Latent Diffusion Models (LDMs)~\cite{ldm,dit}, in particular, have been highly influential. They operate in the latent space of a (Variational) Autoencoder (AE)~\cite{vae}, allowing for efficient generation, and incorporate cross-attention layers to enable conditional image generation, such as guiding the synthesis with text prompts.
%
%Further advancements, like DiT~\cite{dit} replaces the U-Net architecture~\cite{unet}  in LDMs~\cite{ldm} with a Transformer~\cite{transformer} for better performance of these models.
%
The success of text-to-image synthesis models~\cite{ldm,t2irlhf,muse} has led to their widespread adoption in various downstream applications, including image editing~\cite{ip2p}, conditional image generation~\cite{controlnet}, and, notably, image restoration~\cite{stablesr,diffbir,supir}.
%
%Our method also leverages the image generative prior
%to image restoration from a pre-trained LDM~\cite{ldm}.

\begin{figure*}[t]
\centering

\includegraphics[width=1.0\linewidth]{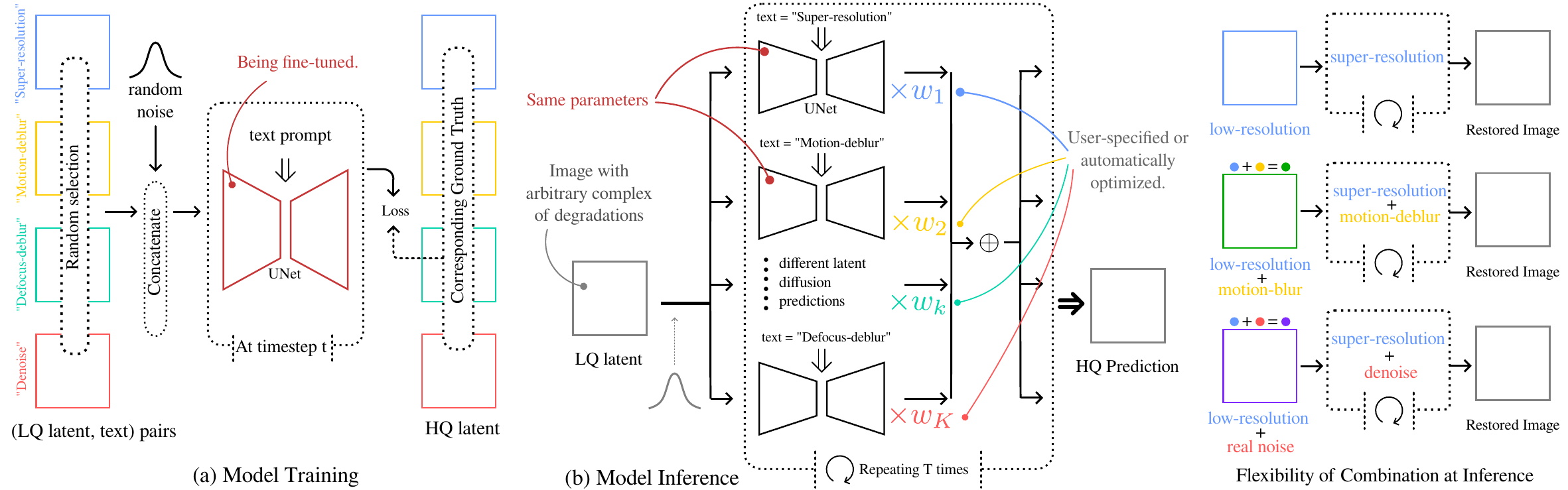}\\
%{\footnotesize
%\hspace{0.18\linewidth} (a) Training Stage \hspace{0.4\linewidth} (b) Inference Stage \hfill
%}
\vspace{-0.5em}
\caption{Diagram of our proposed UniRes framework, \rev{which is designed for the complex degradations as detailed in Sec.~\ref{sec:1}}. (a) We fine-tune a pre-trained
text-to-image LDM~\cite{ldm} (see Sec.~\ref{sec:31} for architecture), on a set of image restoration tasks. (b) At inference time, we flexibly combine the knowledge \rev{from several image restoration tasks}, in order to restore the arbitrary complex degradations in the real-world (see Sec.~\ref{sec:32}).
By combining the latent diffusion predictions with different weights, our framework effectively handles arbitrary degradations.
}
\vspace{-0.5em}
\label{fig:diag}
\end{figure*}

\vspace{0.2em}
\noindent \textbf{Image Restoration}~\cite{esrgan,maxim,hq50k,delbracio2023inversion,spire,codi} 
involves a wide range of low-level vision tasks, such
as super-resolution~\cite{esrgan,realesrgan,realbsr,saharia2022image,facesr,metasr}, deblurring~\cite{realdeblur,su2017deep,gopro,whang2022deblurring,dpdd}, and denoising~\cite{sidd}.
Recently, diffusion models~\cite{ldm,sdxl,add} have emerged as a popular backbone for image restoration~\cite{stablesr,diffbir,supir} for their promising image generative prior.
For instance,
StableSR~\cite{stablesr} fine-tunes adapters on pre-trained diffusion priors without explicit degradation assumptions.
SeeSR~\cite{seesr} preserves the semantic fidelity as low-quality image could be semantically ambiguous.
XPSR~\cite{xpsr} acquires semantic conditions 
with MLLM~\cite{llava} to mitigate incorrect
contents.
DiffBIR~\cite{diffbir}
decouples blind image restoration into degradation
removal and information regeneration steps.
SUPIR~\cite{supir} uses a degradation-robust latent encoder, a large-scale
adapter, and
multi-modal language models~\cite{llava} for photo-realistic image restoration
in the wild.
Notably,
for most diffusion-based methods, the latent diffusion prediction is conditioned on the LQ image input -- through either cross attention with adapter~\cite{controlnet} and frozen backbone~\cite{stablesr,supir,diffbir};
or by concatenating~\cite{ldm,ip2p} the LQ image latent with
the noisy latent $\bm{z}_t$.
While the former mechanism is popular among recent works~\cite{stablesr,supir,diffbir},
ControlNet-based~\cite{pasd} restoration has been found susceptible~\cite{pasd} to
``inconsistency'' issues, where
the restored output exhibits noticeable pixel-level structure discrepancies from the the LQ input.
Unlike recent works, we employ the concatenating~\cite{ldm,ip2p} mechanism for LQ image condition.

%\vspace{0.3em}
\noindent \textbf{All-in-one Restoration}.
Apart from %individual
models designed for specific restoration tasks~\cite{maxim,ipt,realdeblur}, all-in-one models can address multiple degradations simultaneously~\cite{spire,autodir,promptir,restoreagent,airnet,rfir,daclip}.
For instance, 
AutoDIR~\cite{autodir} determines the degradation type, % through CLIP-based model~\cite{clip}, 
and then iteratively restores the image over different restoration operations.
RestoreAgent~\cite{restoreagent} uses LLMs~\cite{instructgpt,llava} to organize  the restoration
sequence, while PromptFix~\cite{promptfix} uses LLMs to enable the use of human instructions.
%model with the capability of human-instructed restoration.
%
PromptIR~\cite{promptir} introduces a degradation-aware prompt learning-based method. % for all-in-one restoration.
DACLIP-IR~\cite{daclip-uir} presents a synthetic degradation pipeline to learn image restoration in the wild, built on top of DA-CLIP~\cite{daclip}.
%
%Unlike typical all-in-one models,
Our method resembles Mixture of Experts~\cite{moe}, and
handles complex degradations in an end-to-end manner, instead of iterating
over several different degradation types~\cite{autodir,restoreagent} like typical all-in-one models.
Moreover, our method only focuses on camera-based degradations, different 
from all-in-one methods~\cite{autodir,restoreagent} that also involve adverse weather conditions.

\noindent \textbf{Generalization} towards data \rev{beyond the training distribution} is a persistent challenge in deep learning.  Techniques such as meta-learning~\cite{maml},
domain adaptation~\cite{domainadaptationsurvey}, test-time adaptation~\cite{testtimeadaptationsurvey}, inference-time model parameter prediction~\cite{calidet,metasr}, and model interpolation~\cite{esrgan,diffusionsoup} have been proposed to address this.
Our method focuses on improving generalization \rev{to complex degradation by using only training data in several well-isolated degradation types.} %by transferring in-domain knowledge via a flexible combination of latent predictions.

\section{Our Approach}
\label{sec:3}
In this paper, we introduce a universal framework for real-world image restoration that leverages the power of pre-trained text-to-image LDMs. Our approach specifically targets the challenging problem of \rev{complex degradation}, where restoration models must effectively handle degradations significantly different from those encountered during training. As illustrated in Figure~\ref{fig:diag}, our method employs a multi-task training strategy on diverse image restoration tasks. This process allows the model to acquire robust image priors and generalized restoration capabilities, enabling effective transfer to novel, \rev{in-the-wild} images during inference.

%In this paper, we propose a universal framework for real-world image restoration based on a pre-trained text-to-image LDM~\cite{ldm}, particularly focusing on out-of-domain generalization.
%%
%The overall pipeline of our proposed method can be found in Fig.~\ref{fig:diag}.
%
%The proposed model is trained in a multi-task manner for several in-domain image restoration tasks, and the knowledge acquired through this process is transferred to out-of-domain image restoration during the inference stage.

%

\subsection{Latent Diffusion for Image Restoration}
\label{sec:31}

\noindent \textbf{Background.}
Diffusion models~\cite{ddpm,ddim,vprediction} are generative models that learn to synthesize data by reversing a gradual noising process.
%where the latents are of the same dimensionality as the data
%$\bm{x}_0\sim q(\bm{x}_0)$.
%
During the forward process, Gaussian noise is progressively added to the data $\bm{x}_0$ according to a fixed Markov chain $q(\bm{x}_t|\bm{x}_{t-1})$, where $t=1,2,\ldots,T$ represents the time step. This process gradually transforms the data into pure noise.
%\begin{equation}
%q(\bm{x}_t|\bm{x}_{t-1}) =
%\mathcal{N}(\bm{x}_t;
%\sqrt{1 - \beta_t} \bm{x}_{t-1}, \beta_t\bm{I}),
%\end{equation}
%%
%where $\beta_t$ is a pre-defined variance schedule.
%
%
The reverse process aims to learn another Markov chain $p_\theta(\bm{x}_{t-1} | \bm{x}_t) $ that progressively removes the noise and recovers the clean data distribution. This is achieved by training a neural network to predict the noise component at each time step. The training objective typically involves minimizing a simplified variational bound~\cite{ddpm}:
\begin{equation}
L(\theta) = \mathbb{E}_{t,\bm{x}_0,\bm{\epsilon}\sim \mathcal{N}(0,1)}
\big[
\| \bm{\epsilon} - \bm{\epsilon}_\theta(\bm{x}_t, t) \|^2
\big],
\end{equation}
where $\bm{\epsilon}_\theta(\cdot)$ denotes the trained model that predicts the noise.

% LDM and DiT

Training diffusion models is computationally demanding. %towards high-resolution image
%generation is computationally demanding.
%
To tackle this, LDMs~\cite{ldm,dit} 
operate on compressed perceptual image representations $\bm{z}_t$ (latent space) instead of directly on the pixels.
LDMs also introduce a text conditioning mechanism
through cross-attention~\cite{ldm} allowing the use of text prompts $\bm{s}$ to guide the synthesis process. Thus, the latent prediction can be
extended to $\bm{\epsilon}_\theta(\bm{z}_t, \bm{s})$,
where the time step $t$ is omitted
for brevity.

\vspace{.5em}
\noindent \textbf{Model Architecture.}
%
%There are two primary approaches to condition the noise prediction
% on the LQ image input, \emph{i.e.}, $\bm{\epsilon}_\theta(\bm{z}_t, \bm{z}_\text{LQ}, \bm{s})$ -- through either cross attention with adapter~\cite{controlnet} and frozen backbone~\cite{stablesr,supir,diffbir};
%or by concatenating~\cite{ldm,ip2p} the LQ image latent $\bm{z}_\text{LQ}$ with
%the noisy latent $\bm{z}_t$.
%
As discussed in Sec.~\ref{sec:2}, unlike adapter-based methods~\cite{stablesr,supir,diffbir}
that have been found susceptible to inconsistency issues~\cite{pasd,supir,stablesr},
we adopt the less-explored latent concatenation~\cite{ldm,ip2p} for LQ image conditioning.
Namely, the LQ image latent $\bm{z}_\text{LQ}$ is concatentated with the noisy latent $\bm{z}_t$
to achieve latent prediction conditioned on LQ image, further extending the notation as
$\bm{\epsilon}_\theta(\bm{z}_t, \bm{z}_\text{LQ}, \bm{s})$.

%Conversely, we adopt the latent concatenation~\cite{ldm,ip2p} as our image condition mechanism.
%
Yet less popular, we note that
(1) it introduces only a small set of new parameters (for the first convolution layer
of UNet~\cite{unet});
(2) it could better preserve the pixel structure from the LQ image, as
all UNet parameters are fine-tuned with LQ-HQ image pairs and penalized for
inconsistencies, instead of being frozen.
The inconsistency issue is often \rev{involved in} ``fidelity-quality trade-off''~\cite{supir,stablesr,diffbir} \rev{mechanisms}.
\rev{The details} will be revisited in the following sections.

%Building upon this assumption, we propose a novel approach that leverages the knowledge learned from individual in-domain restoration tasks — namely, super-resolution~\cite{esrgan,realesrgan}, motion deblurring~\cite{gopro}, defocus deblurring~\cite{dpdd}, and real image denoising~\cite{sidd} — to address out-of-domain image restoration.  By flexibly combining the expertise of these specialized tasks, we aim to achieve robust restoration even for unknown degradation types.

% \subsection{Flexible Combination of Noise Predictions}

% Images captured in the wild often exhibit complex degradations that extend beyond the specific types encountered during training. We hypothesize that these out-of-domain degradations can be effectively approximated by combining various in-domain degradations with different intensity levels.  As discussed in Section~\ref{sec:1}, we focus on four fundamental degradation categories prevalent in camera capture and post-processing: low-resolution, motion blur, defocus blur, and noise.

\subsection{Flexible Combination of Latent Predictions}
\label{sec:32}

Images captured in the wild often exhibit complex degradations that extend beyond those specific
and well-isolated types encountered during training. 
As discussed in Sec.~\ref{sec:1}, 
we \rev{focux on complex degradation, 
an arbitrary combination of four known} degradations with different strengths, including
low-resolution, motion blur, defocus blur, and noise -- all stemming from the camera capture and post-processing pipelines.
%
%Namely, we assume such out-of-domain degradation can be handled by
%transferring the knowledge from the corresponding in-domain restoration tasks
%(\emph{i.e.}, super-resolution~\cite{esrgan,realesrgan},
%motion-deblur~\cite{gopro},
%defocus-deblur~\cite{dpdd},
%and real denoise~\cite{sidd})
%in different intensity levels, to out-of-domain image restoration.
% ALTERNATIVE:
\rev{Based on this}, we propose a novel approach that leverages the knowledge learned from individual \rev{well-isolated} restoration tasks (super-resolution~\cite{esrgan,realesrgan}, motion deblurring~\cite{gopro}, defocus deblurring~\cite{dpdd}, and real image denoising~\cite{sidd}) to address restoration \rev{of images with complex degradation in an end-to-end manner}.  By flexibly combining the expertise of these specialized tasks, we aim to achieve robust restoration for \rev{arbitrary complex} degradations.
%%%%%
% how it is trained

% Preview source code for paragraph 1

\begin{table}
\resizebox{\columnwidth}{!}{%
\setlength{\tabcolsep}{2pt}%
\begin{tabular}{ccccc}
\toprule 
\multirow{2}{*}{\textbf{Task Description}} & \multicolumn{2}{c}{\textbf{Condition Inputs}} & \multirow{2}{*}{\textbf{Output}} & \multirow{2}{*}{\textbf{Training Data}}\tabularnewline
\cmidrule{2-3}
 & \textbf{Image} & \textbf{Text} &  & \tabularnewline
\midrule
\midrule 
Super resolution & LQ & ``Super-resolution'' & HQ & DF2K~\cite{div2k,flickr2k}, LSDIR~\cite{lsdir}\tabularnewline
Motion deblur & LQ & ``Motion-deblur'' & HQ & GoPro~\cite{gopro}, OID-Motion\tabularnewline
Defocus deblur & LQ & ``Defocus-deblur'' & HQ & DPDD~\cite{dpdd}\tabularnewline
Denoise & LQ & ``Denoise'' & HQ & SIDD~\cite{sidd}\tabularnewline
%Text-to-image synthesis & N/A & {[}Per-image caption{]} & HQ & WebLI-v3\tabularnewline
\bottomrule
\end{tabular}

}

\caption{Multi-task training detail of the proposed method.
``LQ'' and ``HQ'' denote low-quality and high-quality images, respectively.
``DF2K'' is a combination of DIV2K~\cite{div2k} and Flickr2K~\cite{flickr2k}.
The OID-Motion is Open Images Dataset~\cite{oid} with simulated camera shake
blur~\cite{shakeblur}, as detailed in Sec.~\ref{sec:4}.
}

\label{tab:trainlist}
\end{table}

\vspace{0.5em}
\noindent \textbf{Model Training.}
To let the model acquire knowledge for \rev{different} image restoration
tasks, we fine-tune a pre-trained LDM in the manner of multi-task learning
using DDPM~\cite{ddpm} formulation.
As summarized in Tab.~\ref{tab:trainlist}, these tasks are decoupled with per-task constant text prompts. During the training process, each training sample is randomly sampled from the those discussed image restoration tasks.
%

%
%
%In particular, our model learns four image restoration tasks:
%super-resolution~\cite{esrgan,realesrgan},
%motion-deblur~\cite{gopro},
%defocus-deblur~\cite{dpdd}, and denoise~\cite{sidd}.
%

During the training process, we randomly drop the image
and the text conditions following \cite{ip2p}, in order to enable
classifier-free guidance~\cite{cfg} using a single diffusion model, since it has proven to be effective~\cite{ip2p,supir}.
This also means our model implicitly learns to do ``blind restoration''
$\bm{\epsilon}_\theta(\bm{z}_t,\bm{z}_\text{LQ},\varnothing)$, \rev{without specifying} the
degradation type. % (reasonable since we need to handle unknown degradations),
%and fully unconditional generation
%$\bm{\epsilon}_\theta(\bm{z}_t,\varnothing,\varnothing)$.
%

\vspace{.5em}
\noindent \textbf{Model Inference.}
As we assume the degradation in image $\bm{x}$ can be \rev{a mixture of four aforementioned} degradations with arbitrary strengths,
we formulate the corresponding restoration problem as
a weighted combination of the corresponding latent diffusion predictions (from the different restoration tasks).
The knowledge acquired from \rev{those isolated} restoration tasks can hence
be transferred to \rev{complex degradation} image restoration.
Let $K$ denote the number of restoration tasks, and
$\bm{w}\triangleq[w_1,\ldots,w_K]$ denote the noise
combination weights for the respective $K$ latent predictions.

During a sampling step (e.g., using DDIM~\cite{ddim}),
the weighted combination of the noise predictions for
the $K$ different tasks forms the full latent diffusion prediction:
\begin{align}
    \tilde{\bm{\epsilon}}_\theta(\bm{z}_t,\bm{z}_\text{LQ};\bm{w}) & \triangleq
    \sum_{k=1}^K w_k \cdot \bm{\epsilon}_\theta(\bm{z}_t,\bm{z}_\text{LQ},\bm{s}_k ),
    \label{eq:core}
\end{align}
where $w_k\in \mathbb{R}$, $\sum_{k=1}^K w_k = 1$, and $\bm{s}_k$ is the text prompt
corresponding to the $k$-th prediction task.
This formulation extends the concept of classifier-free guidance~\cite{cfg}, commonly used in related works~\cite{ip2p,supir}, by allowing for a more flexible and adaptive combination of task-specific knowledge through latent diffusion predictions.

The adjustable weights $(w_k)$ in Eq.\eqref{eq:core} enable our model to tailor the restoration process to individual images. For instance, a blurry night photo might benefit from a high weight for motion deblurring and a small weight for denoising.
This adaptability highlights our framework's ability to dynamically select and utilize relevant knowledge.
\rev{See Fig.~\ref{fig:wild1x} and Fig.~\ref{fig:wild4x} for examples.}
Importantly, our formulation is not limited to those tasks discussed earlier (blind restoration and the four explicit tasks). It can be readily extended to incorporate latent predictions from other restoration or image manipulation tasks within a unified framework.

\begin{figure}[t]
    %\centering
    \resizebox{\columnwidth}{!}{%
    \setlength{\tabcolsep}{1pt}%
    \renewcommand{\arraystretch}{0.5}%
    \begin{tabular}{cccccc}
\includegraphics[width=0.3\columnwidth]{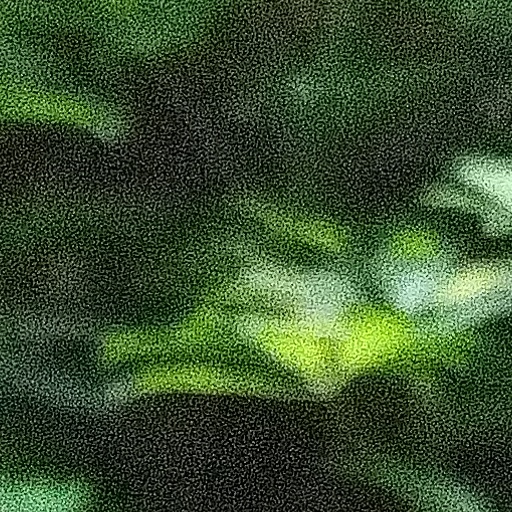} &
\includegraphics[width=0.3\columnwidth]{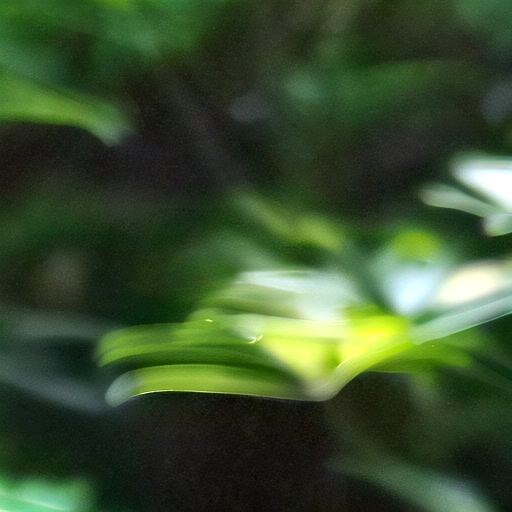} &
\includegraphics[width=0.3\columnwidth]{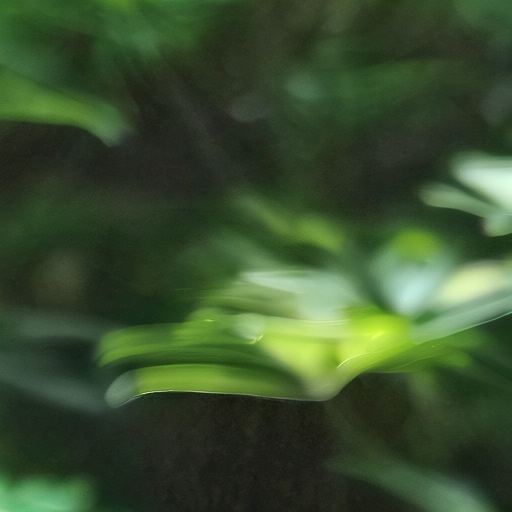} &
\includegraphics[width=0.3\columnwidth]{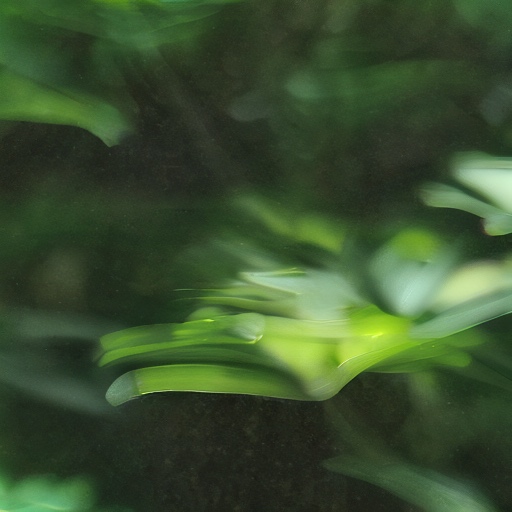} &
\includegraphics[width=0.3\columnwidth]{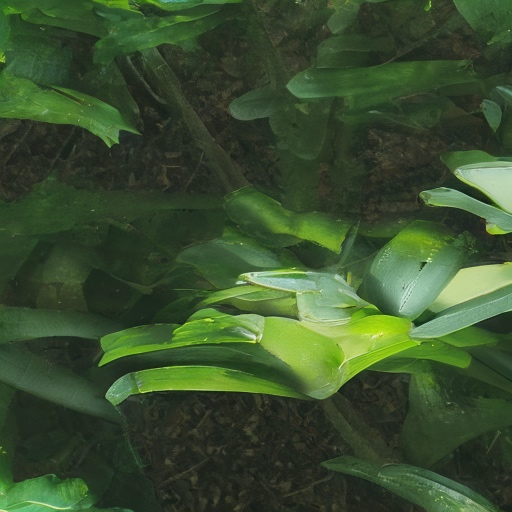} &
\includegraphics[width=0.3\columnwidth]{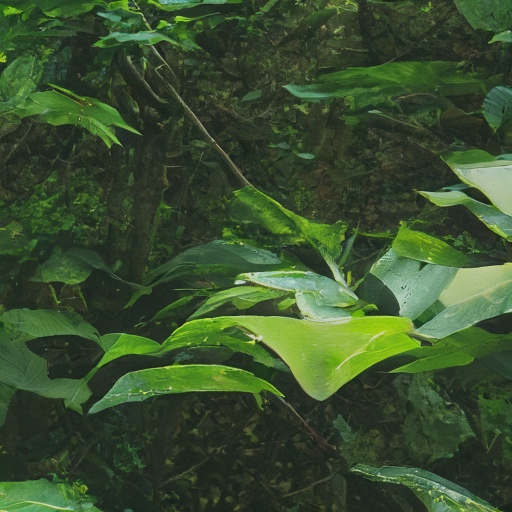} \\
{\large LQ} & $512\times512$  ($\times 1$) & $256\times256$ ($\times 2$)  & $128\times128$ ($\times 4$)  &
$64\times64$ ($\times 8$)  & $32\times32$ ($\times 16$)
\end{tabular}}
    \vspace{-7pt}
    \caption{Effect of the ``DownLQ'' term with different downscaling factors. The result images refer to super-resolution on a pre-processed LQ image, where the LQ input image is downscaled by a factor (as annotated below each image) and then \rev{bicubic}-upscaled back to its original resolution. Larger downscaling factors elicit the model to generate more details.
    The ``DownLQ'' term is used by our model to control the fidelity-quality trade-off. }
    \vspace{-1em}
    \label{fig:downlq}
\end{figure}

% Mauricio Stopped here. 10h28PM 112Nov.
\noindent \textbf{Fidelity \emph{vs.} Quality.}
Many adapter-based restoration models~\cite{stablesr,diffbir,supir} introduce a fidelity-quality trade-off mechanism to pull the prediction towards the LQ image.
However, we empirically observe that our concatenation-based model (discussed in Sec.~\ref{sec:31})
is inherently more conservative in terms of generating details compared to adapter-based approaches.
%
%
%Hence, different from other proposed trade-off mechanisms
%that pulls the prediction towards the LQ image, we introduce a mechanism for inducing controlled hallucinations in order to improve image quality by introducing a ``DownLQ'' noise prediction.
Since restoration model tends to generate more details when more information is lost
from the LQ image~\cite{mengwei},
a very natural way to further improve our model is to add an additional
``DownLQ'' inference task with dedicated weight to Eq.~\ref{eq:core} (and hence $K$ is increased by $1$).
``DownLQ'' refers to the super-resolution prediction task $\bm{\epsilon}_\theta(\bm{z}_t,\bm{z}_\text{DownLQ},\bm{s}_\text{SR} )$,
conditioned on a pre-processed LQ input ($\bm{z}_\text{DownLQ}$) that is down-sampled by a constant factor in pixel space, and then \rev{bicubic-}upscaled back to its original resolution.
As shown in Fig.~\ref{fig:downlq}, our model generates more details with a larger down-sampling factor. 
We empirically observe that $\times 4$ down-sampling provides visually higher-quality details, while not excessively hallucinating, and thus adopt DownLQ with $\times 4$ factor as our fidelity-quality trade-off mechanism.
%This also shows the flexibility of our framework for extensions.

\vspace{0.2em}
\noindent\textbf{Other Potential Extensions},
such as positive and negative text prompts~\cite{diffbir,supir},
and semantic text prompts~\cite{supir,seesr} are reported effective
for image restoration.
While our Eq.~\ref{eq:core} is flexible enough to incorporate them,
we refrain from doing so in our experiments, because they
are not our contribution and dilute the performance gain of our proposed method.
Our framework is ``universal'' in the sense that
different restoration tasks can be incorporated and combined using the same formulation \rev{for arbitrary complex degradations}.

\subsection{Optimal Combination Weights}
\label{sec:33}

During inference, our model allows arbitrary weighted combination
of different \rev{latent diffusion} predictions.
%
%What values to use for those combination weights becomes spontaneous question.
These weights can be determined by the user per their preference,
or automatically calculated based on certain criterion.

Let $g(\bm{x},\bm{w})$ be our model which generates a
restored image through DDIM~\cite{ddim} algorithm.
Since the optimal weights $\bm{w}^*$ varies across
different inputs depending on the complex degradations, a simple method to figure out the
optimal combination weights for an image is grid search.
Let $Q(\cdot)$ be an image quality assessment
function such as~\cite{vila,musiq,nima}, which approximates human perceptual preference.
Then, the optimization is done by grid search within
a pre-defined range $[\gamma,\delta]^K$, where $\gamma\leqslant\delta$ and $\gamma,\delta\in\mathbb{R}$:
\begin{align}
\bm{w}^* &= \arg\max_{\bm{w}\in\Omega} \, Q(g(\bm{x},\bm{w})) \\
\text{where}\quad\Omega &\triangleq \Big\{
\bm{w} \in [\gamma,\delta]^K
\Big|
\sum_{k=1}^K w_k = 1
\Big\}
\end{align}
In this paper, we empirically adopt MUSIQ~\cite{musiq} as
the $Q(\cdot)$ function.
The range $[\gamma,\delta]$ as well as the grid search interval are
hyper-parameters provided by the user.

The optimal weights are expected to correlate with the intensity of
different degradation types.
For instance, if motion blur is the dominating degradation, we expect the $w_k$ for motion
deblur to be higher than the rest values.
Furthermore, the average optimal combination weights over a set of images
is expected to reflect the \rev{overall degradation types}. %image distribution characteristics in terms of degradation.
%
%The proposed framework is ``universal'' also in the sense that the combination weights can be adjusted for arbitrary mixtures of degradationsfor better restoration.
%
%Our framework's universality extends to its adaptability to arbitrary mixtures of degradations.
By adjusting the combination weights, the model can be optimized for \rev{arbitrary complex} degradations. %, leading to improved restoration across diverse scenarios.

\section{Experiments}
\label{sec:4}

% To validate the effectiveness of the proposed method, we quantitatively and qualitative evaluate it on several sets of test images.
% %
% Our model is trained on $32$ TPU-v5 with Jax~\cite{jax}.
% %
% Quantitative metrics are calculated using IQA-PyTorch~\cite{pyiqa}.

%\textbf{OID-Motion}.
%
%To create a diverse dataset of degraded images, we adopted the camera shake blur simulation method from~\cite{shakeblur}. This involved generating random blur kernels with a range of intensities and sizes, which were then applied to high-quality images from the Open Image Dataset to simulate per-object motion blur.  We further degraded these images by introducing lens blur (using Gaussian blur kernels), shot noise, read-out noise, and JPEG compression. By randomly sampling the parameters for each degradation, we created a dataset that encompasses a wide spectrum of image quality, from heavily degraded to pristine

\noindent\textbf{Implementation Details.}
%
% We use the LDM~\cite{ldm} ($865$M parameters) backbone pre-trained on the WebLI~\cite{pali}. Then, we
% %
% % It is fine-tuned on multiple tasks simultaneously on several datasets,
% % as listed in Tab.~\ref{tab:trainlist}.
% employ the datasets and predefined tasks outlined in Tab.~\ref{tab:trainlist} to fine-tune the model.
The WebLI~\cite{pali} pre-trained text-to-image LDM~\cite{ldm} model, with 865M parameters, is fine-tuned using tasks outlined in Tab.~\ref{tab:trainlist}.
%Simulation algorithms are used to generate LQ images for data sources in Tab.~\ref{tab:trainlist} that lack them.
Specifically, the Real-ESRGAN degradation pipeline~\cite{realesrgan} is adopted for the super-resolution task.
For the motion-deblur task, we simulated camera shake blur~\cite{shakeblur} (by applying random blur kernels of different intensities and sizes, to individual objects within an image, mimicking the blur that occurs due to motion movement) using images from Open Image Dataset~\cite{oid}.
%
% ``OID-Motion'', it refers to the images from the Open Image Dataset~\cite{oid}, but we paired it with simulated camera shake blur~\cite{shakeblur}. This effect is achieved by applying random blur kernels to individual objects within the image, mimicking the blur that occurs due to motion movement.
%
The GoPro dataset~\cite{gopro} that has limited scene diversity is supplemented with OID motion-deblur dataset (details for this dataset are given in supplementary material).
%
%More details are given in supplementary material.

%DF2K\footnote{DF2K is a combination of DIV2K~\cite{div2k} and Flickr2K~\cite{flickr2k}.}
%and LSDIR~\cite{lsdir} 
%with Real-ESRGAN~\cite{realesrgan} degradation pipeline for super-resolution,
%GoPro~\cite{gopro} for motion-deblur,
%DPDD~\cite{dpdd} for defocus-deblur,
%and SIDD~\cite{sidd} for real denoise.
%
%Similar to \cite{instructgpt}, we also keep the pre-training task (text-to-image synthesis)
%with the WebLI-v3~\cite{pali} dataset.
%
%The training tasks are summarized in Tab.~\ref{tab:trainlist}.
%
When fine-tuning the LDM, we randomly sample among super-resolution, motion-deblurring, defocus-deblurring and denoising with probabilities $0.32$, $0.28$, $0.18$, $0.22$, respectively.
The random drop rates for both the image and text conditions are set to $0.1$.
We fine-tuned the model for 200K steps using JAX~\cite{jax} and $32$ TPU-v5.
The batch size and learning rate are set to $256$ and \verb|8e-5|, respectively.

% \textbf{Inference.}
%
% During training and inference, the resolution for input and output images are both $512\times 512$.
%
%Input images smaller than 512 will be resized to 512 first
%with bicubic sampling.
%
To find the optimal combination of latent predictions from the six domain experts --blind restoration (BR), super resolution (SR), motion deblur (MD), defocus deblur (DD), denoise (DN), and the fidelity-quality trade-off term (DownLQ)-- we use a grid search strategy (Sec.~\ref{sec:33}).
%
% When not specified, the combination weights are found through grid search as described in Sec.~\ref{sec:33}.
%
The search \rev{grid} $[\gamma,\delta]$ is set to $[-0.2,1.2]$ \rev{with $0.2$ interval} based on empirical evidence. We also constrain the search space to allow at most one negative weight \rev{to reduce search space size}.
% and at most one negative weight at the same time.
%
%During inference, we use different noise
%combination weights to demonstrate its
%effect.
%
%For brevity, we denote super-resolution,
%motion-deblur, defocus-deblur, denoise, and
%blind-restoration as "SR", "MD", "DD",
%"DN", and "BR", respectively.
%
%For instace, we use "SR=$1.0$" to denote the
%case where the weight for super-resolution is $1.0$,
%and $0.0$ for the rest.
%
%The semantic text caption is not used by default
%because it is not always available in the wild.
%
Following StableSR~\cite{stablesr},
we post-process our generated image with AdaIN~\cite{stylegan} for color correction.

\vspace{0.2em}
\noindent\textbf{Evaluation.}
% To validate the effectiveness of the proposed method, we quantitatively and qualitatively evaluate it on several sets of test images.
We validate the efficacy of our method through quantitative and qualitative evaluations.
%
% Following \cite{supir,stablesr},
% image quality was quantitatively evaluated using non-reference metrics, including ClipIQA~\cite{clipiqa}, MUSIQ~\cite{musiq} and ManIQA~\cite{maniqa}.
% When HQ reference images are available, the full-reference metrics such as PSNR, SSIM, LPIPS and FID are also reported.
% Image quality was assessed using both reference and non-reference metrics. 
Non-reference metrics, such as ClipIQA~\cite{clipiqa}, MUSIQ~\cite{musiq} and ManIQA~\cite{maniqa}, are utilized in the absence of high-quality (HQ) reference images. When HQ reference images are available, we additionally report full-reference metrics, including PSNR, SSIM, LPIPS, and FID. This evaluation approach aligns with~ \cite{supir,stablesr,diffbir}.
For baseline comparison, we consider the following state-of-the-art image
restoration models:
StableSR~\cite{stablesr},
DiffBIR~\cite{diffbir},
SUPIR~\cite{supir},
DACLIP-IR~\cite{daclip-uir}.
All baseline results are generated using respective official code and checkpoints.
%
% \textbf{DiversePhotos}.
%
We evaluate our method on the commonly used Real60~\cite{supir}, RealSR~\cite{realsr}, and DRealSR~\cite{drealsr}, \rev{but note that they are single (instead of complex) degradations.}
% following the protocol of \cite{stablesr}.
% The commonly used real out-of-domain degradation test
% The benchmark datasets commonly used \cite{realsr,drealsr,supir} only 
%These datasets encompass a limited range of capture devices and degredations, with low resolution as the dominating degradation.

% Preview source code for paragraph 3

\begin{table}
\resizebox{\columnwidth}{!}{%
\setlength{\tabcolsep}{9pt}%
\begin{tabular}{cccc}
\toprule 
\textbf{Model} & \textbf{ClipIQA}~\cite{clipiqa} & \textbf{MUSIQ}~\cite{musiq} & \textbf{ManIQA}~\cite{maniqa}\tabularnewline

\midrule 
\rowcolor{gblue!10}\multicolumn{4}{c}{DiversePhotos$\times$1 ($160$ images, size $512\times 512$)}\tabularnewline
\midrule

StableSR~\cite{stablesr} & \cellcolor{gblue!10}0.6227 & \cellcolor{gblue!20}61.39 & \cellcolor{gblue!10}0.3992\tabularnewline
DiffBIR~\cite{diffbir} & \cellcolor{gblue!20}0.6453 & \cellcolor{gblue!10}59.97 & \cellcolor{gblue!20}0.4922\tabularnewline
SUPIR~\cite{supir} & 0.5060 & 51.68 & 0.3745\tabularnewline
DACLIP-IR~\cite{daclip-uir} & 0.3497 & 46.16 & 0.2567\tabularnewline
%\midrule 
%SR-Only &  &  & \tabularnewline
UniRes & \cellcolor{gblue!40}0.6519 & \cellcolor{gblue!40}68.22 & \cellcolor{gblue!40}0.5021\tabularnewline

\midrule
\rowcolor{ggreen!10}\multicolumn{4}{c}{Real60~\cite{supir} ($60$ images, size $512\times 512$)}\tabularnewline
\midrule

StableSR~\cite{stablesr} & 0.7593 & \cellcolor{ggreen!20}72.06 & 0.4997\tabularnewline
DiffBIR~\cite{diffbir} & \cellcolor{ggreen!10}0.7851 & 70.10 & \cellcolor{ggreen!10}0.5772\tabularnewline
SUPIR~\cite{supir} & \cellcolor{ggreen!40}0.8217 & \cellcolor{ggreen!10}71.61 & \cellcolor{ggreen!40}0.6716\tabularnewline
DACLIP-IR~\cite{daclip-uir} & 0.3802 & 54.57 & 0.2765\tabularnewline
%\midrule 
%SR-Only & 0.7223 & 72.22 & 0.5138\tabularnewline
UniRes & \cellcolor{ggreen!20}0.7894 & \cellcolor{ggreen!40}75.15 & \cellcolor{ggreen!20}0.6080\tabularnewline

\bottomrule
\end{tabular}}
\vspace{-0.7em}
\caption{Real-world image restoration with $\times 1$ upscaling. The output size
is identical to the input image size.
The top-$3$ results are highlighted in different color transparency,
where the top-$1$ result is marked by the darkest color.
Our method manifests robustness on real-world complex degradations
reflected by DiversePhotos$\times$1.
Visualizations can be found in Fig.~\ref{fig:wild1x}.
Even on test images where low-resolution is the dominating degradation (\emph{i.e.}, Real60), our model still achieves competitive results. }
\vspace{-1.5em}

\label{tab:wild1x}
\end{table}

\noindent\textbf{DiversePhotos.}
\rev{Existing image restoration benchmarks~\cite{div2k,gopro,supir,dpdd,sidd} often focus on isolated degradation types, failing to capture the complexity of real-world scenarios where multiple degradations frequently co-occur.
To bridge this gap, we constructed ``DiversePhotos,'' a new dataset for evaluating performance under complex degradation.
It is compiled from SPAQ~\cite{spaq}, KONIQ~\cite{koniq}, and LIVE~\cite{live}, encompassing a wide variety of degradation types.
Each image in DiversePhotos is characterized by the presence of at least two distinct real-world degradations.
%Each image in DiversePhotos contains at least two distinct real-world degradations.
We created two versions: ``DiversePhotos$\times 1$,'' consisting of $160$ images at $512\times 512$ resolution, divided into four dominating degradation categories of $40$ images each (low resolution, motion blur, defocus blur, and real noise), and ``DiversePhotos$\times 4$,'' which utilizes $128\times 128$ center crops of the same images.
As of now, no publicly available benchmark specifically reflects the challenge of complex degradations. Further details are provided in the supplementary material.}

\subsection{Restoration \rev{of Complex Degradations}}

% Preview source code for paragraph 2

\begin{table}
\resizebox{\columnwidth}{!}{%
\setlength{\tabcolsep}{4pt}%
\begin{tabular}{ccccccc}
\toprule 
\textbf{Model} & \textbf{PSNR} & \textbf{SSIM} & \textbf{LPIPS$\downarrow$} & \textbf{ClipIQA} & \textbf{MUSIQ} & \textbf{ManIQA}\tabularnewline
\midrule
\rowcolor{gorange!10}\multicolumn{7}{c}{ DiversePhotos$\times$4 (160 images, size $128\times128$)}\tabularnewline
\midrule
StableSR~\cite{stablesr} & - & - & - & 0.5177 & \cellcolor{gorange!10}41.53 & 0.2983\tabularnewline
DiffBIR~\cite{diffbir} & - & - & - & \cellcolor{gorange!40}0.6190 & \cellcolor{gorange!20}54.09 & \cellcolor{gorange!20}0.4551\tabularnewline
SUPIR~\cite{supir} & - & - & - & \cellcolor{gorange!10}0.5308 & 39.11 & \cellcolor{gorange!10}0.3403\tabularnewline
DACLIP-IR~\cite{daclip-uir} & - & - & - & 0.2924 & 32.99 & 0.2339\tabularnewline
%\midrule 
%SR-Only & - & - & - &  &  & \tabularnewline
UniRes & - & - & - & \cellcolor{gorange!20}0.6050 & \cellcolor{gorange!40}62.40 & \cellcolor{gorange!40}0.4656\tabularnewline

\midrule 
\rowcolor{gyellow!10}\multicolumn{7}{c}{RealSR$\times4$~\cite{realsr} (100 images,
size $128\times128$)}\tabularnewline
\midrule
StableSR~\cite{stablesr} & 23.32 & \cellcolor{gyellow!20}0.6799 & \cellcolor{gyellow!40}0.3002 & \cellcolor{gyellow!10}0.6234 & \cellcolor{gyellow!20}65.88 & 0.4275\tabularnewline
DiffBIR~\cite{diffbir} & \cellcolor{gyellow!10}23.51 & 0.6180 & 0.3650 & \cellcolor{gyellow!40}0.7053 & \cellcolor{gyellow!40}69.28 & \cellcolor{gyellow!40}0.5582\tabularnewline
SUPIR~\cite{supir} & 22.31 & 0.6275 & 0.3554 & \cellcolor{gyellow!20}0.6658 & 62.55 & \cellcolor{gyellow!20}0.5095\tabularnewline
DACLIP-IR~\cite{daclip-uir} & \cellcolor{gyellow!40}25.11 & \cellcolor{gyellow!40}0.7129 & \cellcolor{gyellow!20}0.3158 & 0.2828 & 46.66 & 0.2621\tabularnewline
%\midrule 
%SR-Only & 24.30 & 0.6868 & 0.2909 & 0.5495 & 62.46 & 0.4007\tabularnewline
%UniRes & 21.96 & 0.5218 & 0.4448 & 0.6693 & 70.56 & 0.5510\tabularnewline

% unires-std config
UniRes & \cellcolor{gyellow!20}24.34 & \cellcolor{gyellow!10}0.6493 & \cellcolor{gyellow!10}0.3282 & 0.5710 & \cellcolor{gyellow!10}65.46 & \cellcolor{gyellow!10}0.4347\\

\midrule 
\rowcolor{gred!10}\multicolumn{7}{c}{DRealSR$\times4$~\cite{drealsr} (93 images, size
$128\times128$)}\tabularnewline
\midrule
StableSR~\cite{stablesr} & \cellcolor{gred!40}26.71 & \cellcolor{gred!40}0.7224 & \cellcolor{gred!40}0.3284 & \cellcolor{gred!10}0.6356 & 58.51 & 0.3874\tabularnewline
DiffBIR~\cite{diffbir} & 24.58 & 0.5830 & 0.4670 & \cellcolor{gred!40}0.7068 & \cellcolor{gred!40}66.14 & \cellcolor{gred!40}0.5543\tabularnewline
SUPIR~\cite{supir} & 23.72 & 0.6016 & 0.4348 & \cellcolor{gred!20}0.6852 & \cellcolor{gred!10}59.83 & \cellcolor{gred!20}0.4974\tabularnewline
DACLIP-IR~\cite{daclip-uir} & \cellcolor{gred!20}26.45 & \cellcolor{gred!20}0.6693 & \cellcolor{gred!10}0.4179 & 0.3173 & 41.03 & 0.2568\tabularnewline
%\midrule 
%SR-Only & 25.82 & 0.6925 & 0.3488 & 0.5821 & 59.73 & 0.3927\tabularnewline
%UniRes & 23.70 & 0.5355 & 0.5056 & 0.6872 & 68.14 & 0.5366\tabularnewline

% unires-std config
UniRes & \cellcolor{gred!10}26.25 & \cellcolor{gred!10}0.6611 & \cellcolor{gred!20}0.3927 & 0.6055 & \cellcolor{gred!20}62.08 & \cellcolor{gred!10}0.4247\\

\bottomrule
\end{tabular}}
\vspace{-0.7em}
\caption{Real-world image restoration with $\times 4$ upscaling. The output size
is $512\times512$ for all test sets. While the dominating degradation becomes
low resolution in this case, it is still accompanied with other
degradations in the DiversePhotos$\times$4 test set.}
\vspace{-1.5em}

\label{tab:wild4x}
\end{table}

% !! summary

% As the emphasis of this paper is the out-of-domain
% (in-the-wild) image restoration performance, we
% first report the results on the real degradation test sets.
% The quantitative results are shown in Tab.~\ref{tab:wild1x} and Tab.~\ref{tab:wild4x}
% for image restoration with $1\times$ and
% $4\times$ upscaling, respectively.
This paper focuses on \rev{complex degradations.} %image restoration in out-of-domain (in-the-wild) scenarios.
We begin with presenting the quantitative results for image restoration on real degradation test sets. Tab.~\ref{tab:wild1x} and Tab.~\ref{tab:wild4x} show the results of $\times 1$ and $\times 4$ upscaling, respectively.
%
% The above observations on DiversePhotox$\times$1 are consistent with the quantitative results in Tab.~\ref{tab:wild1x}.
%
Our method produces superior results when addressing complex, real-world degradations. As a result, it outperforms all other methods on every metric on DiversePhotos$\times$1 (Tab.~\ref{tab:wild1x} top)
%
% For example, our method achieves a MUSIQ score of $68.22$, which is higher than the second place, \emph{i.e.}, StableSR ($61.39$) by a large margin.
--our method's MUSIQ score of 68.22 significantly surpasses that of the second-place method, StableSR, which scored 61.39.
%
% On the other hand, 
% On Real60$\times$1 (Tab.~\ref{tab:wild1x} bottom),
% our model still achieve a competitive performance, where the dominating degradation, down-sampling, is used as the main training tasks for our competitors.
Meanwhile, our model still achieves a competitive performance \rev{when the degradation is less complex, as suggested by the result on the real-world super-resolution benchmark, \emph{i.e.}, } Real60~\cite{supir} (Tab.~\ref{tab:wild1x} bottom). %, which is a commonly used benchmark for super-resolution that favors our competitors (SUPIR proposed this set of test images).

\begin{figure*}[t]
    %\centering
    \resizebox{\linewidth}{!}{%
    \setlength{\tabcolsep}{1pt}%
    \renewcommand{\arraystretch}{0.5}%
    \large%
    \begin{tabular}{ccccccccccc}
    
%        & LQ & StableSR~\cite{stablesr} & DiffBIR~\cite{diffbir} &
%        SUPIR~\cite{supir} & DACLIP-IR~\cite{daclip-uir} & Ours
%        \\
        
        \rotatebox{90}{\qquad\quad LQ} &
        \includegraphics[width=0.18\linewidth]{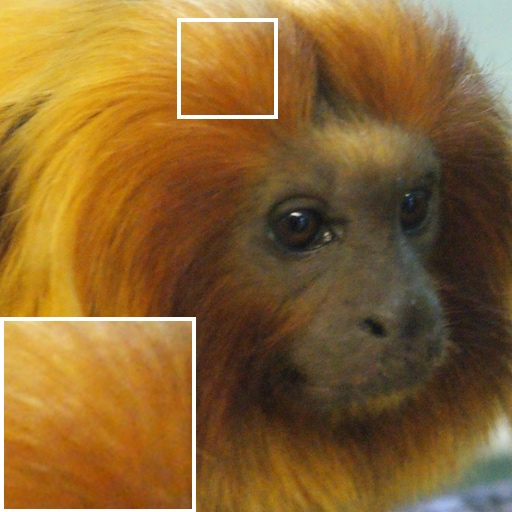} &
        \includegraphics[width=0.18\linewidth]{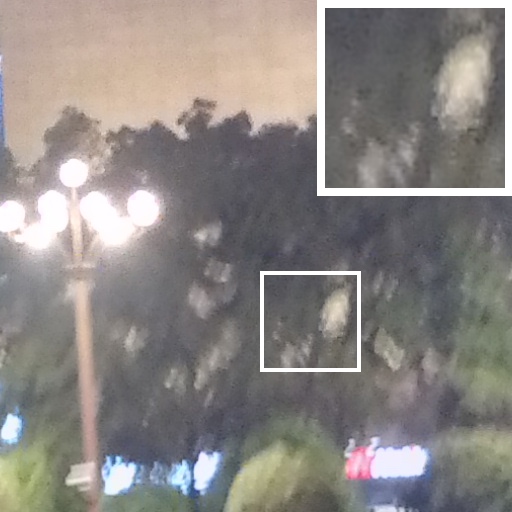} &
        \includegraphics[width=0.18\linewidth]{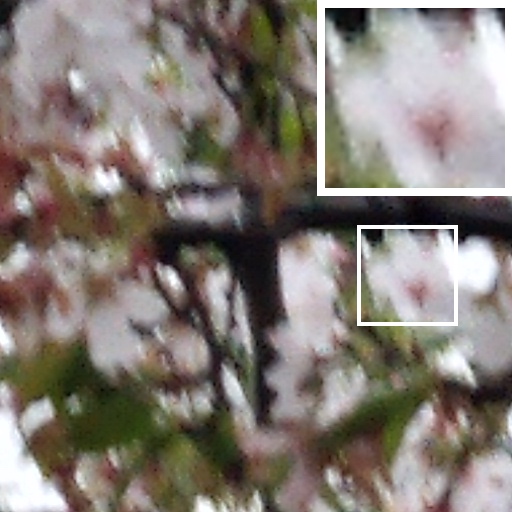} &
        \includegraphics[width=0.18\linewidth]{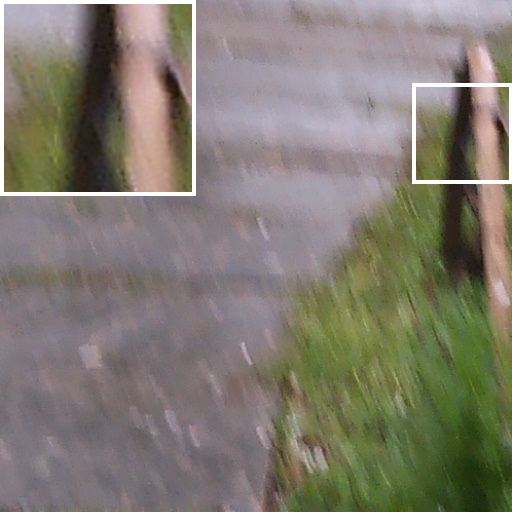} &
        \includegraphics[width=0.18\linewidth]{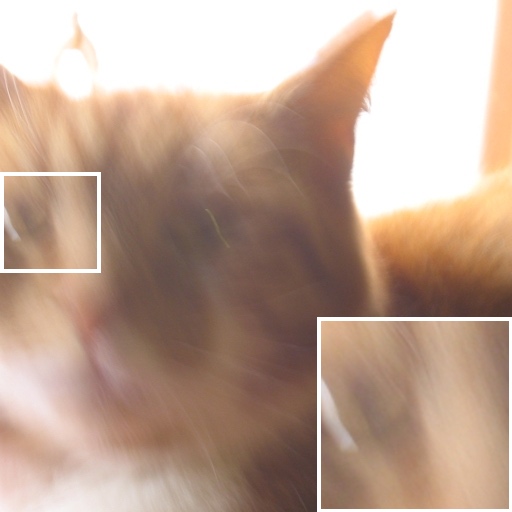} &
        \includegraphics[width=0.18\linewidth]{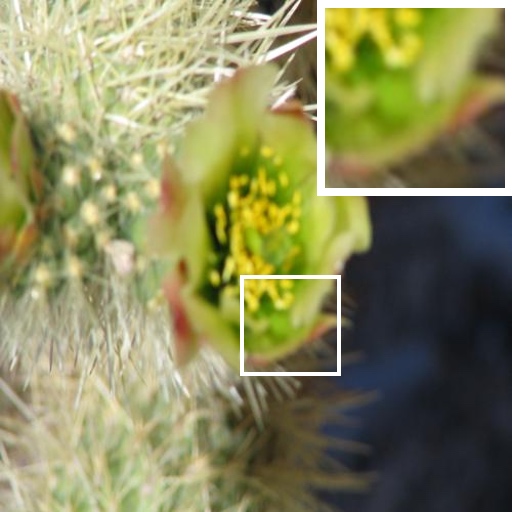} &
        \includegraphics[width=0.18\linewidth]{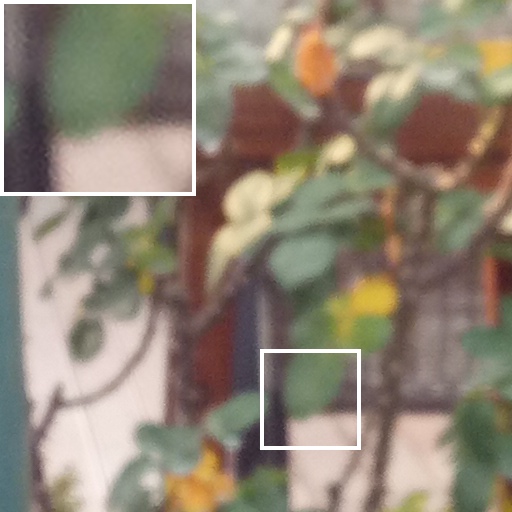} &
        \includegraphics[width=0.18\linewidth]{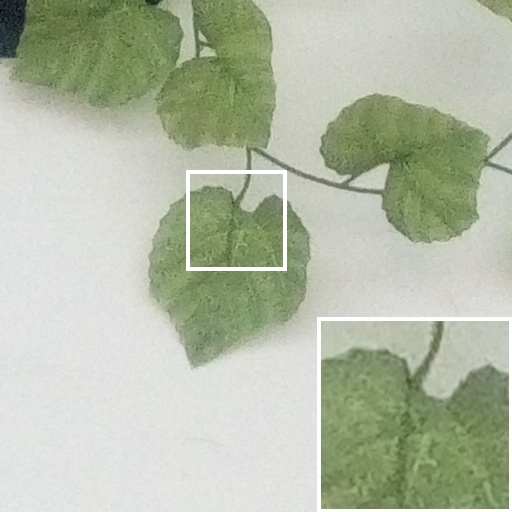} &
        \includegraphics[width=0.18\linewidth]{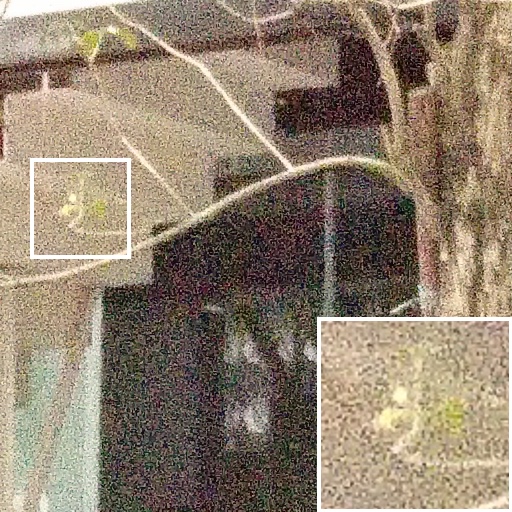} &
        \\
        \rotatebox{90}{\quad StableSR~\cite{stablesr}} &
        \includegraphics[width=0.18\linewidth]{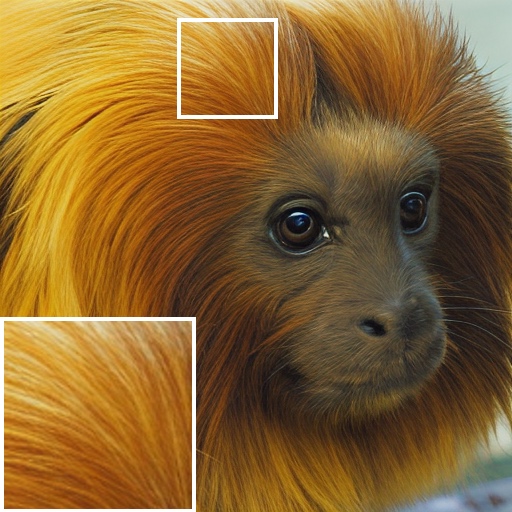} &
        \includegraphics[width=0.18\linewidth]{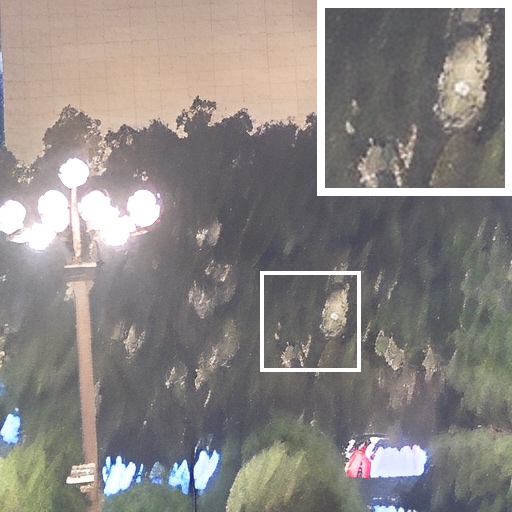} &
        \includegraphics[width=0.18\linewidth]{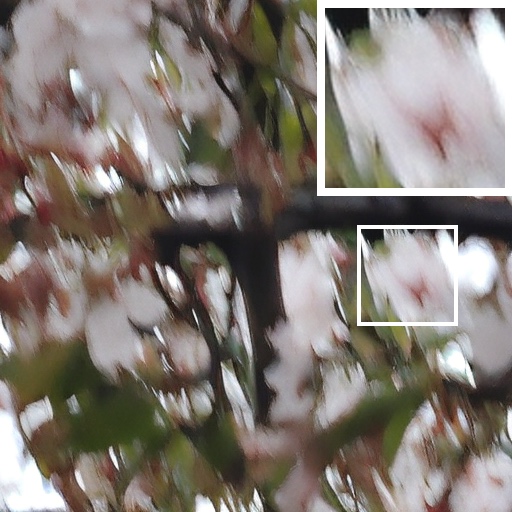} &
        \includegraphics[width=0.18\linewidth]{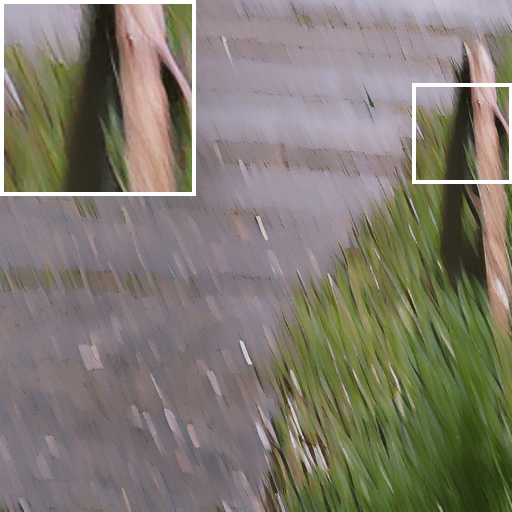} &
        \includegraphics[width=0.18\linewidth]{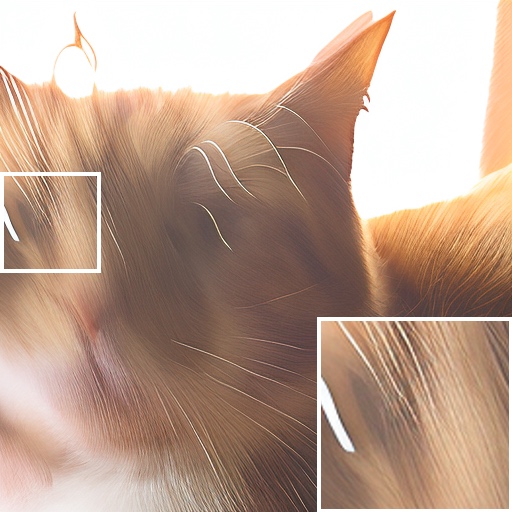} &
        \includegraphics[width=0.18\linewidth]{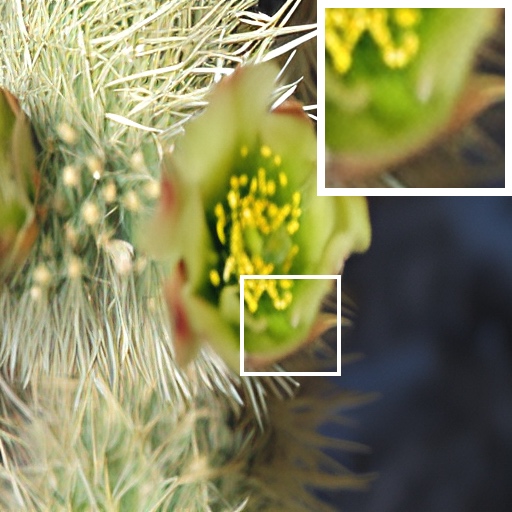} &
        \includegraphics[width=0.18\linewidth]{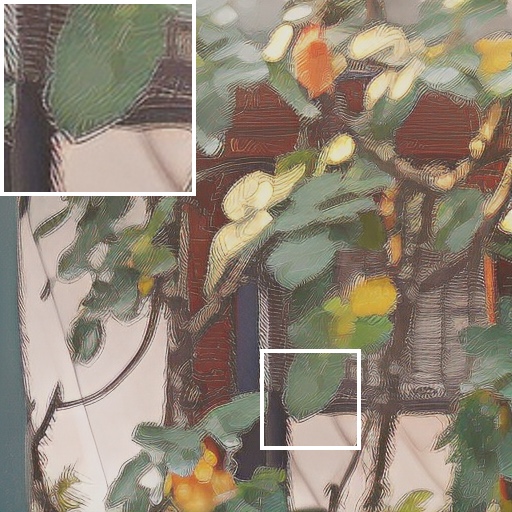} &
        \includegraphics[width=0.18\linewidth]{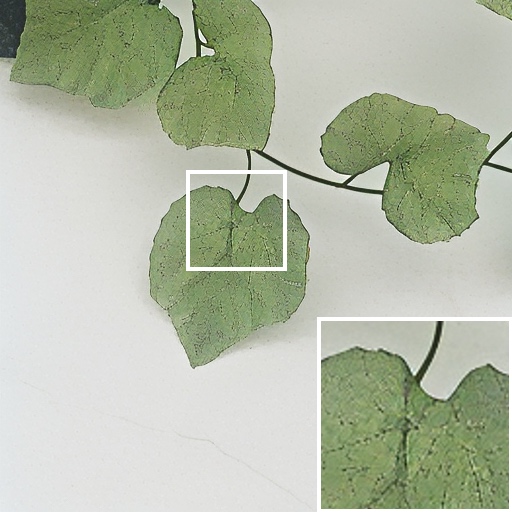} &
        \includegraphics[width=0.18\linewidth]{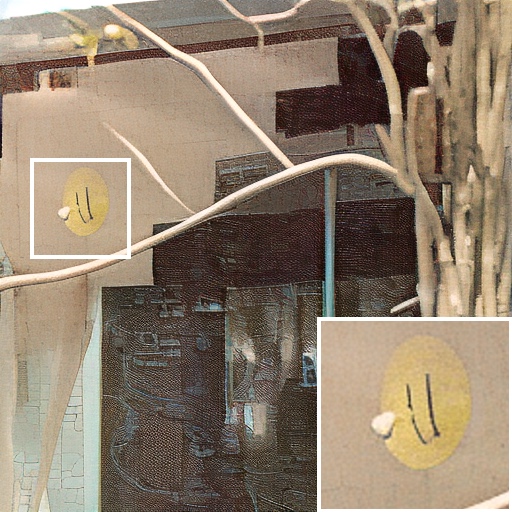} &
        \\
        \rotatebox{90}{\quad DiffBIR~\cite{diffbir}} &
        \includegraphics[width=0.18\linewidth]{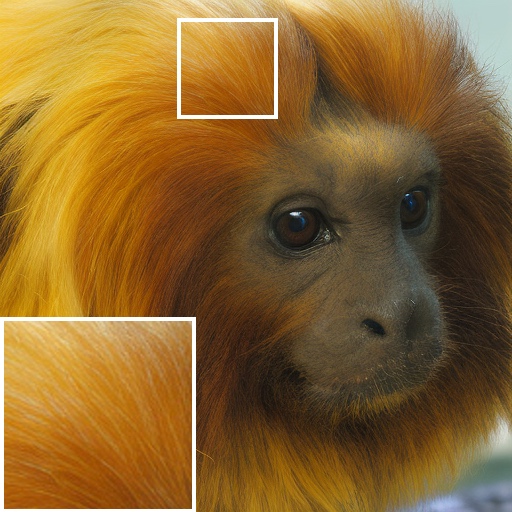} &
        \includegraphics[width=0.18\linewidth]{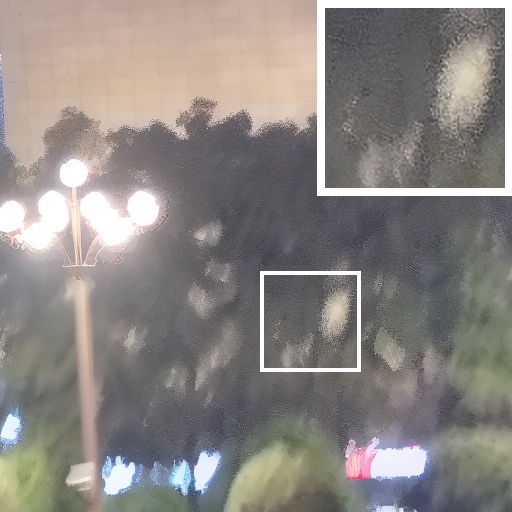} &
        \includegraphics[width=0.18\linewidth]{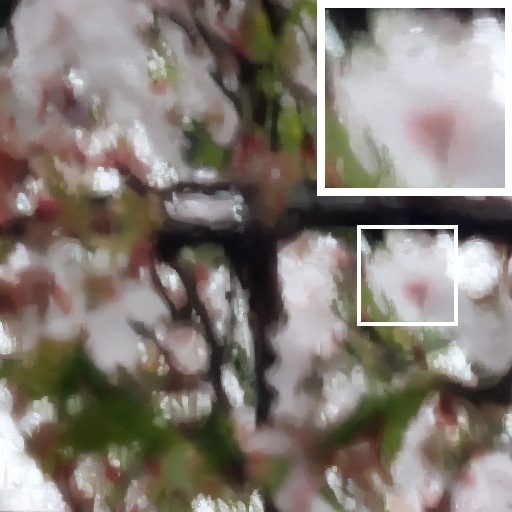} &
        \includegraphics[width=0.18\linewidth]{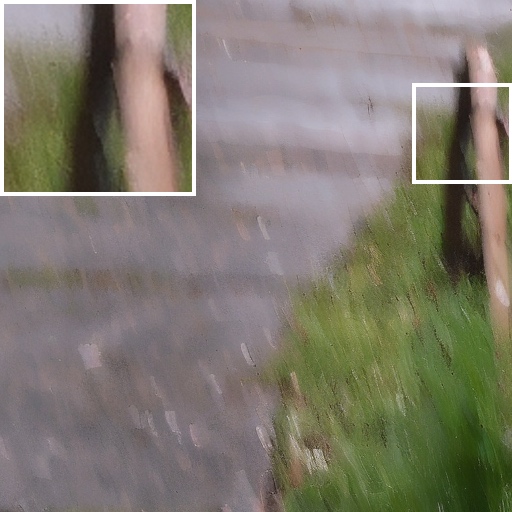} &
        \includegraphics[width=0.18\linewidth]{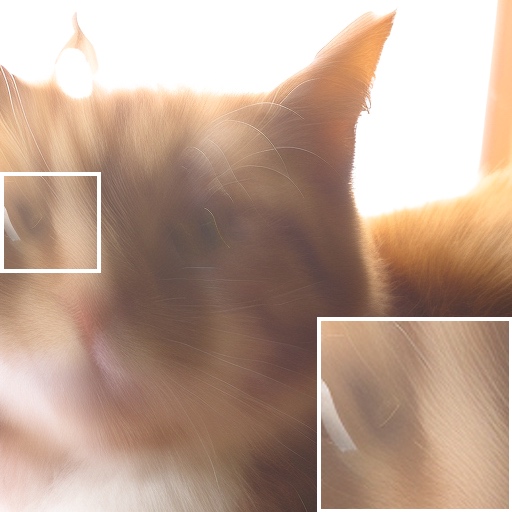} &
        \includegraphics[width=0.18\linewidth]{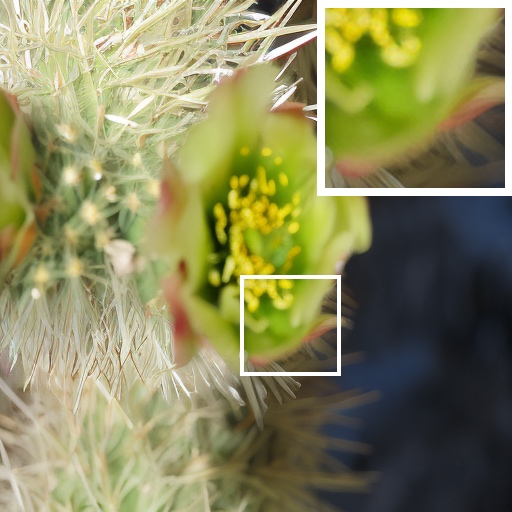} &
        \includegraphics[width=0.18\linewidth]{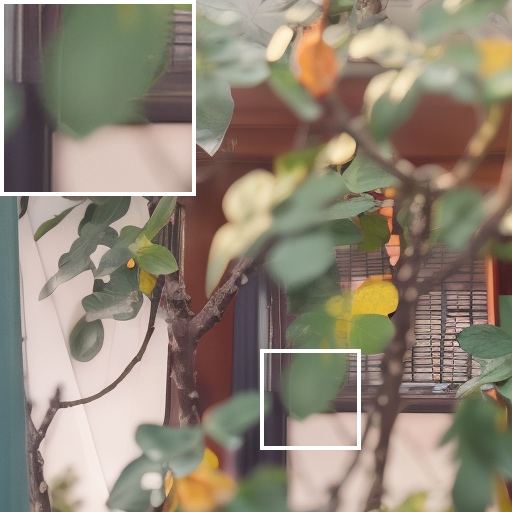} &
        \includegraphics[width=0.18\linewidth]{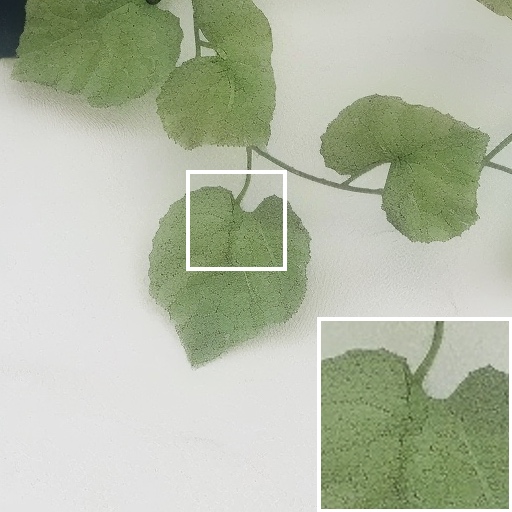} &
        \includegraphics[width=0.18\linewidth]{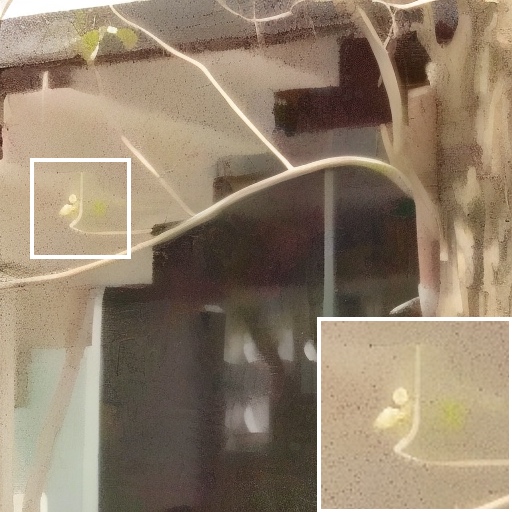} &
        \\
        \rotatebox{90}{\quad SUPIR~\cite{supir}} &
        \includegraphics[width=0.18\linewidth]{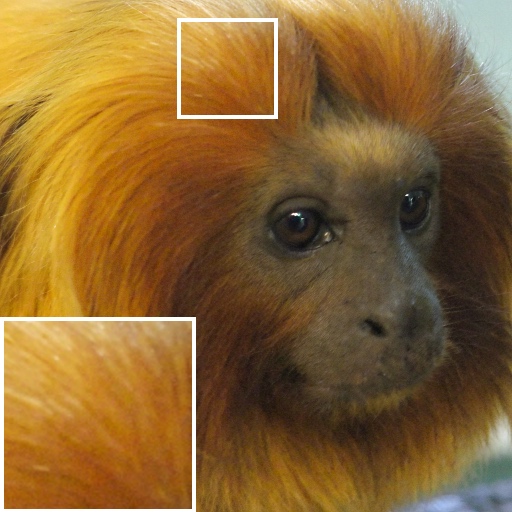} &
        \includegraphics[width=0.18\linewidth]{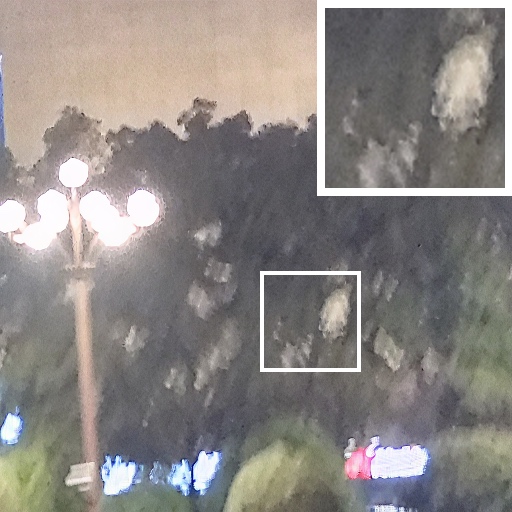} &
        \includegraphics[width=0.18\linewidth]{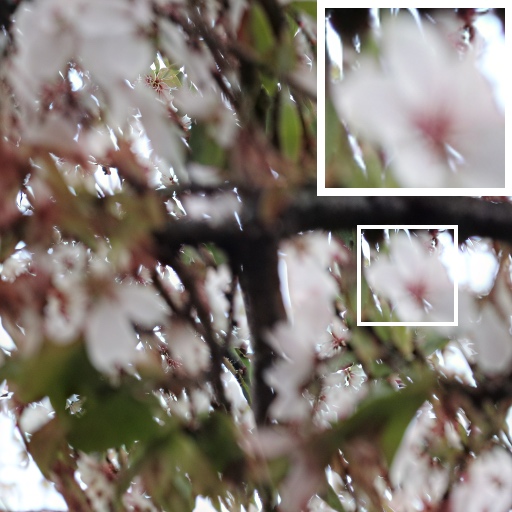} &
        \includegraphics[width=0.18\linewidth]{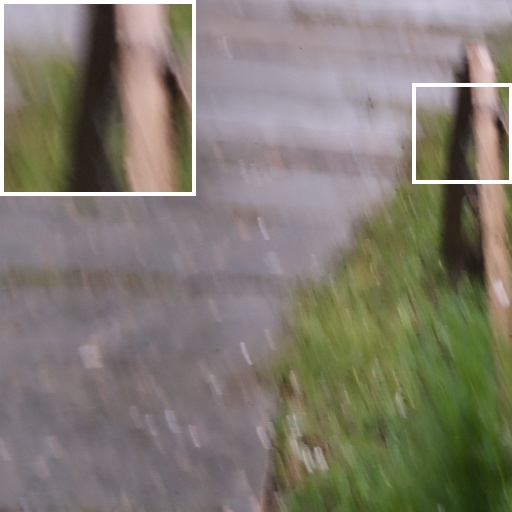} &
        \includegraphics[width=0.18\linewidth]{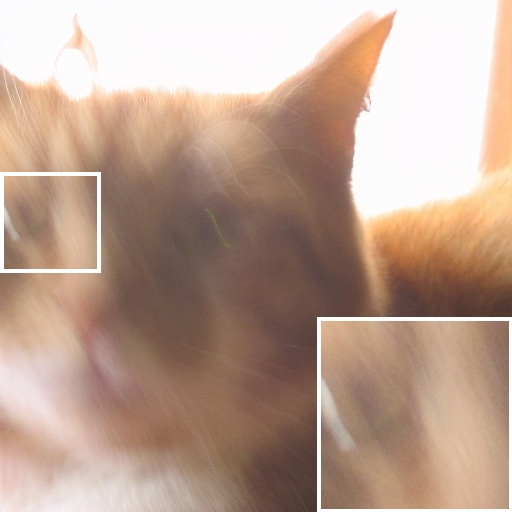} &
        \includegraphics[width=0.18\linewidth]{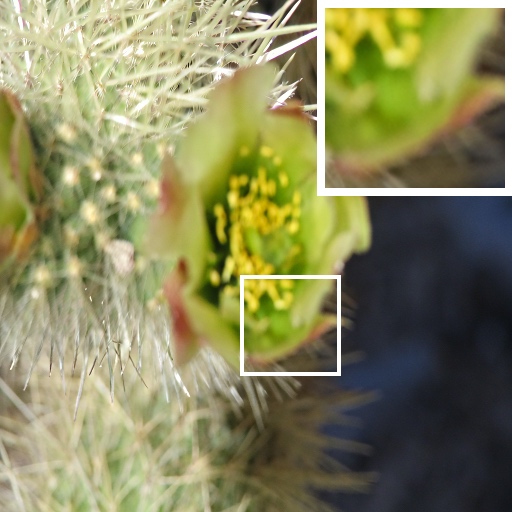} &
        \includegraphics[width=0.18\linewidth]{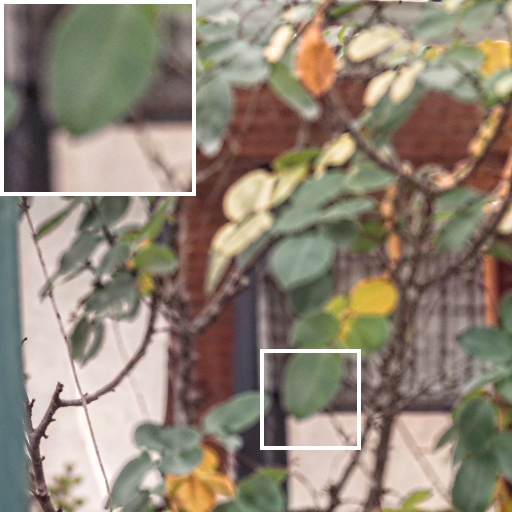} &
        \includegraphics[width=0.18\linewidth]{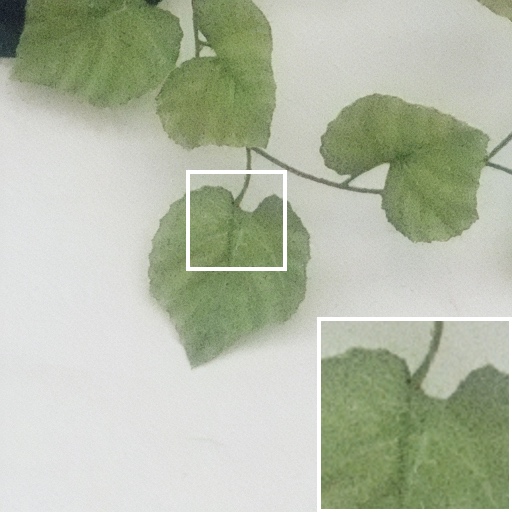} &
        \includegraphics[width=0.18\linewidth]{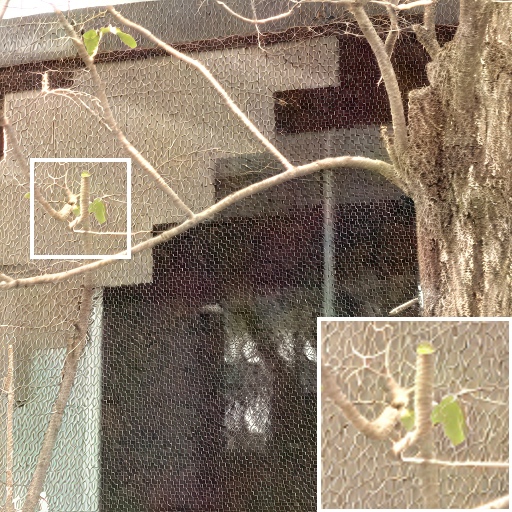} &
        \\
        \rotatebox{90}{\, DACLIP-IR~\cite{daclip-uir}} &
        \includegraphics[width=0.18\linewidth]{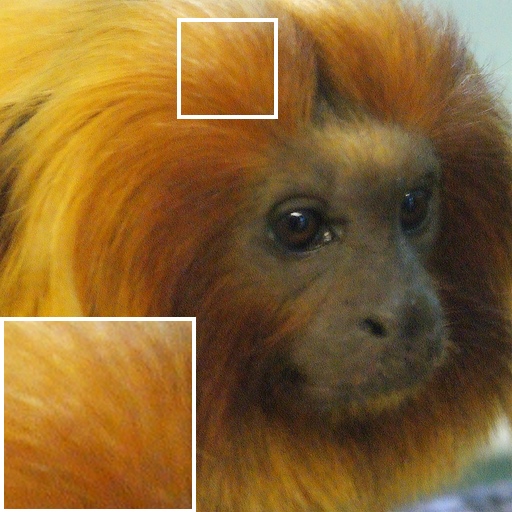} &
        \includegraphics[width=0.18\linewidth]{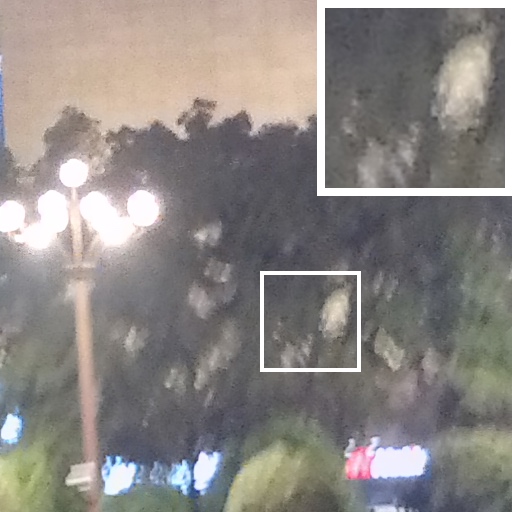} &
        \includegraphics[width=0.18\linewidth]{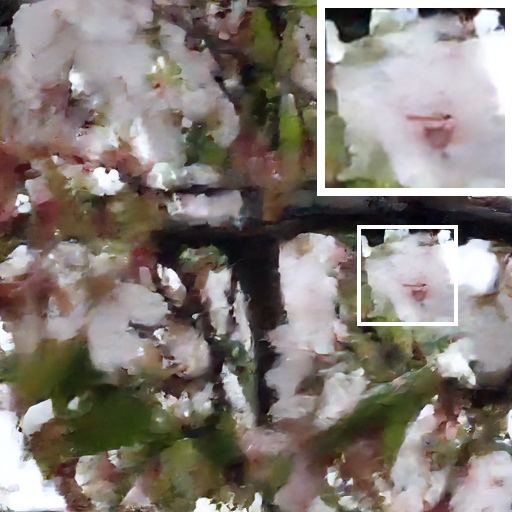} &
        \includegraphics[width=0.18\linewidth]{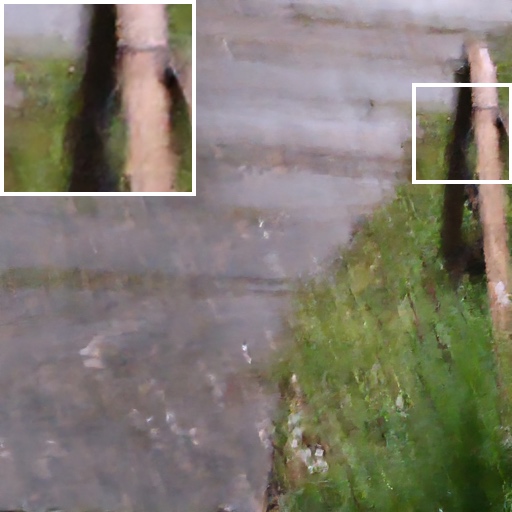} &
        \includegraphics[width=0.18\linewidth]{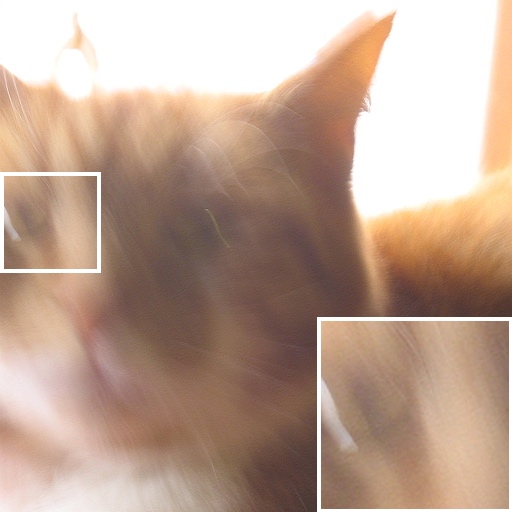} &
        \includegraphics[width=0.18\linewidth]{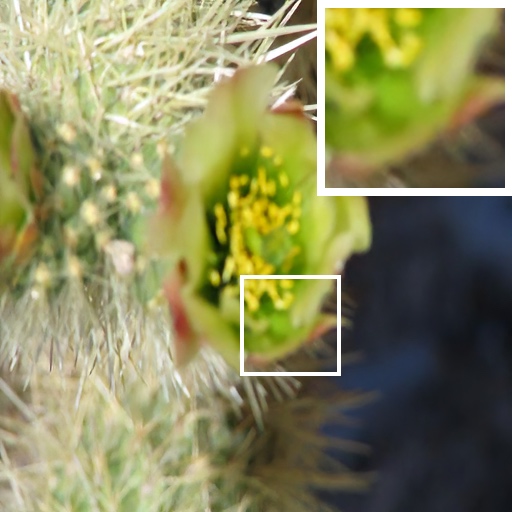} &
        \includegraphics[width=0.18\linewidth]{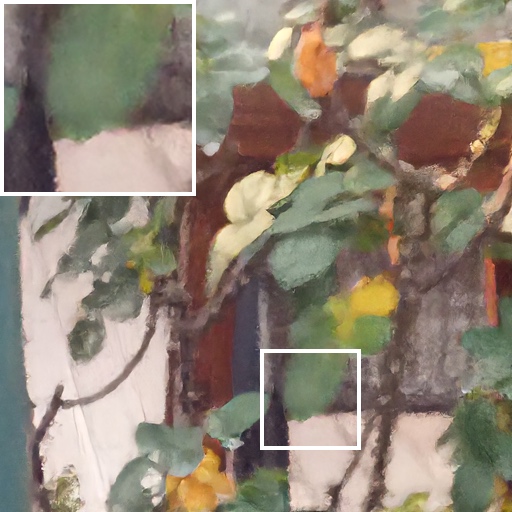} &
        \includegraphics[width=0.18\linewidth]{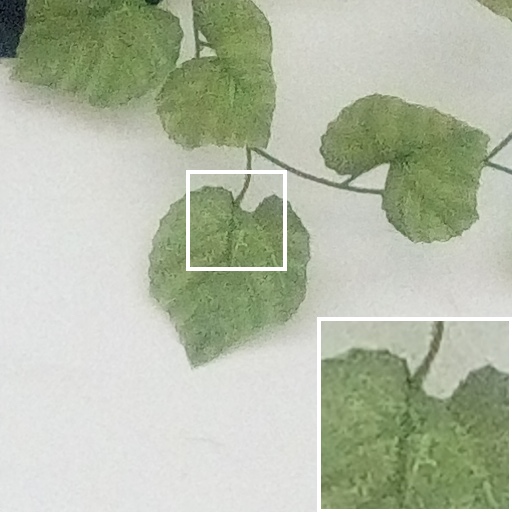} &
        \includegraphics[width=0.18\linewidth]{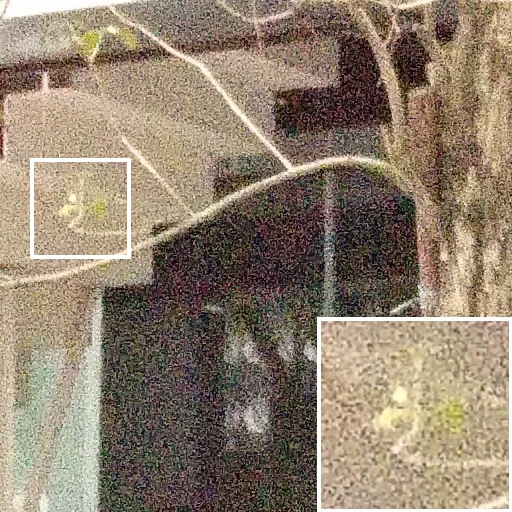} &
        \\
        \rotatebox{90}{\qquad \quad Ours} &
        \includegraphics[width=0.18\linewidth]{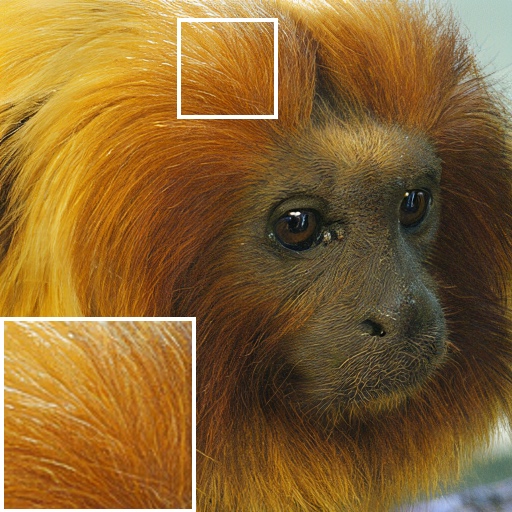} &
        \includegraphics[width=0.18\linewidth]{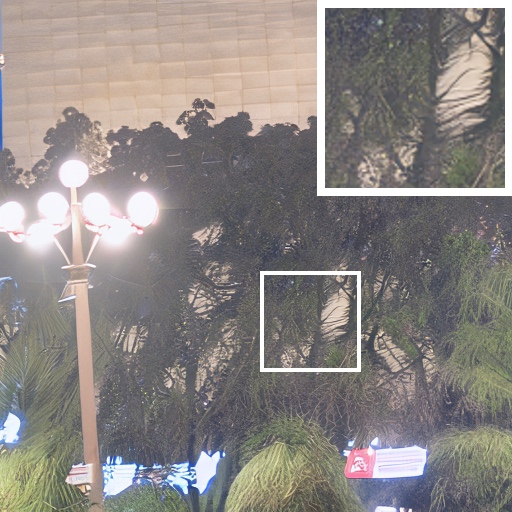} &
        \includegraphics[width=0.18\linewidth]{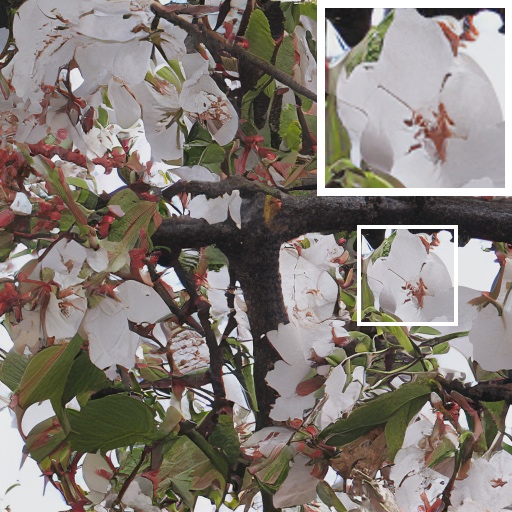} &
        \includegraphics[width=0.18\linewidth]{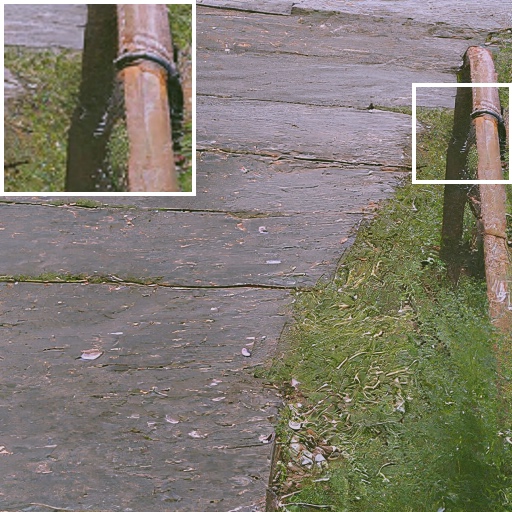} &
        \includegraphics[width=0.18\linewidth]{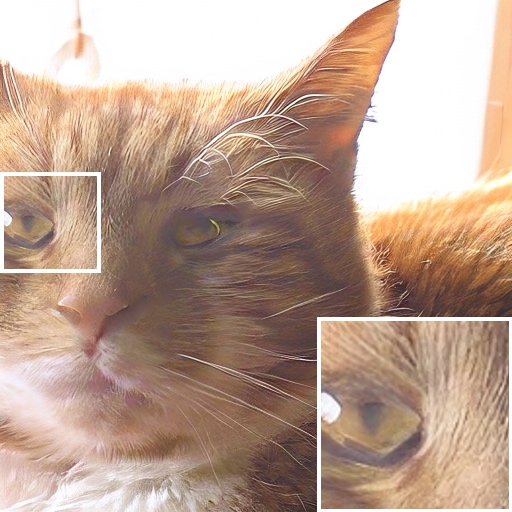} &
        \includegraphics[width=0.18\linewidth]{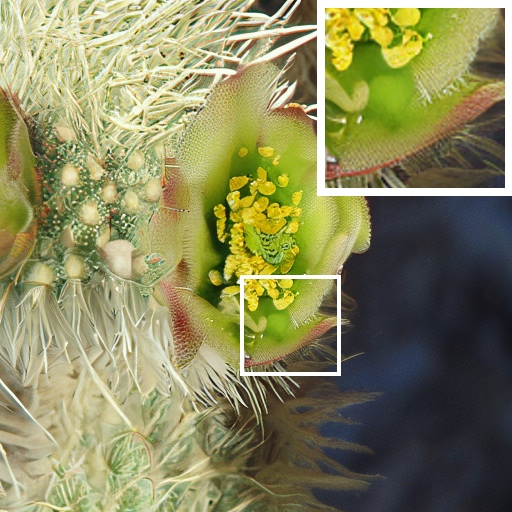} &
        \includegraphics[width=0.18\linewidth]{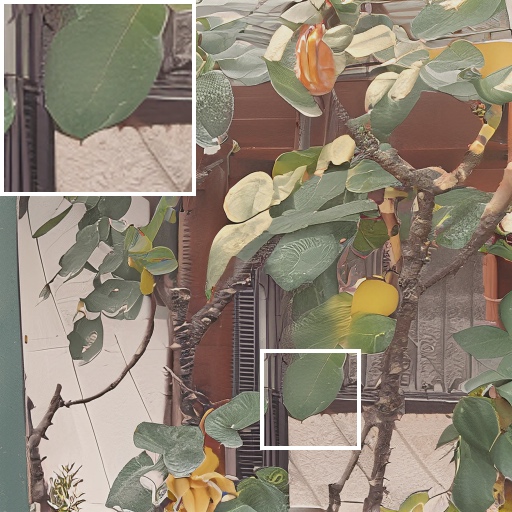} &
        \includegraphics[width=0.18\linewidth]{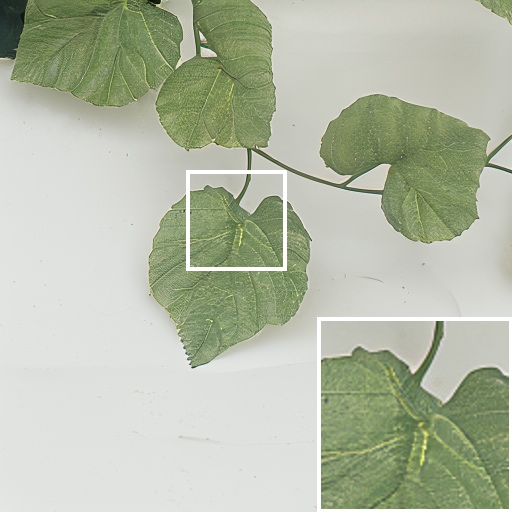} &
        \includegraphics[width=0.18\linewidth]{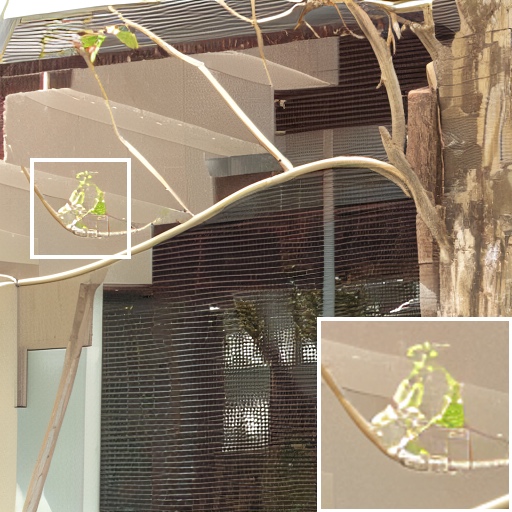} &
        
    \end{tabular}}
    \vspace{-1em}
    \caption{Real-world image restoration on DiversePhotos$\times$1. This figure corresponds to
    Tab.~\ref{tab:wild1x}. The LQ images involve diverse real-world complex degradations. Our model is more robust against those degradations than other models. Zoom in for details.}
    \label{fig:wild1x}
\end{figure*}

\begin{figure}[t]
    %\centering
    \resizebox{\columnwidth}{!}{%
    \setlength{\tabcolsep}{1pt}%
    \renewcommand{\arraystretch}{0.1}%
    \large%
    \begin{tabular}{cccccc}
        \rotatebox{90}{\qquad \quad LQ} &
        \includegraphics[width=0.4\linewidth]{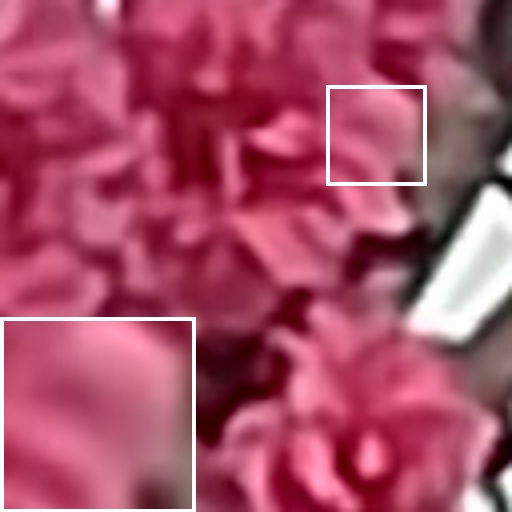} &
        \includegraphics[width=0.4\linewidth]{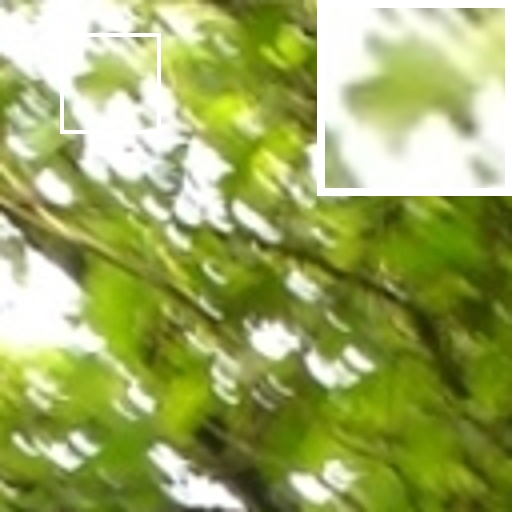} &
        \includegraphics[width=0.4\linewidth]{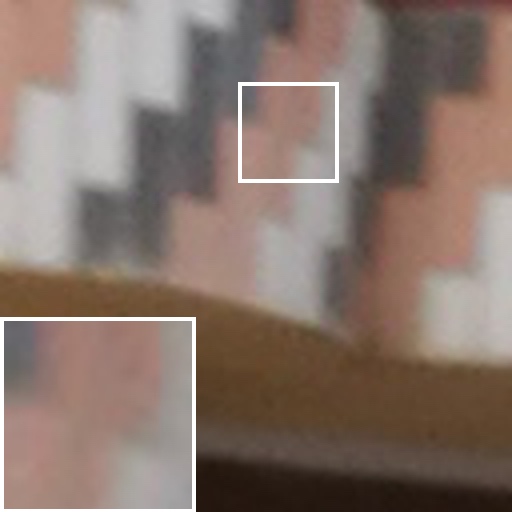} &
        \includegraphics[width=0.4\linewidth]{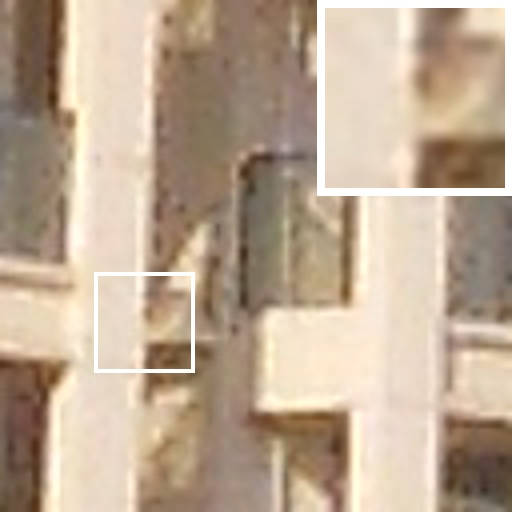} &
        \includegraphics[width=0.4\linewidth]{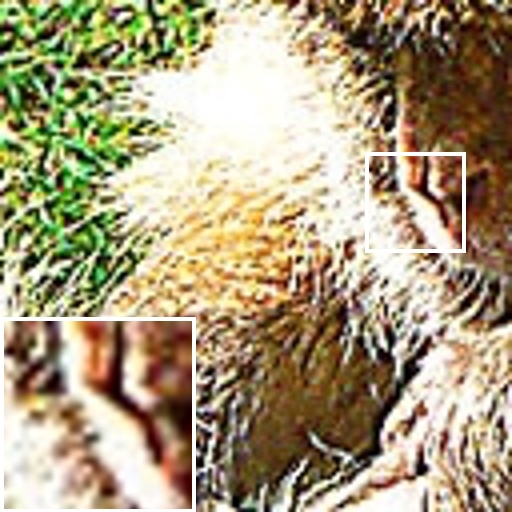}
        \\
        
        \rotatebox{90}{\quad StableSR~\cite{stablesr}} &
        \includegraphics[width=0.4\linewidth]{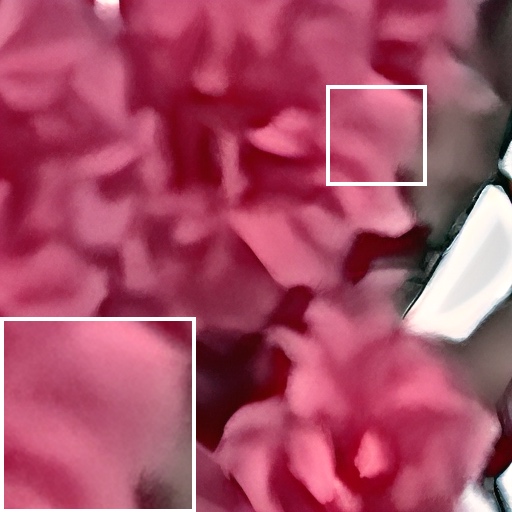} &
        \includegraphics[width=0.4\linewidth]{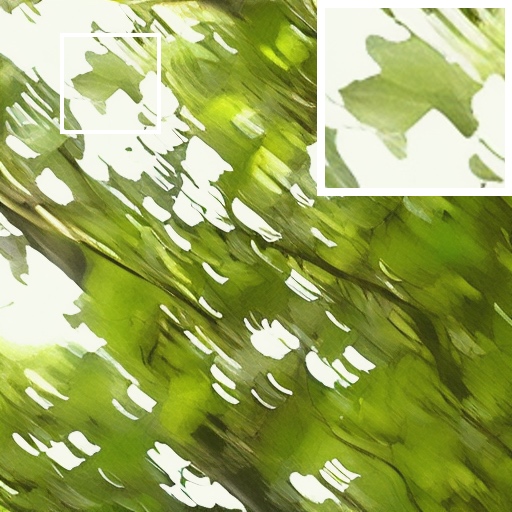} &
        \includegraphics[width=0.4\linewidth]{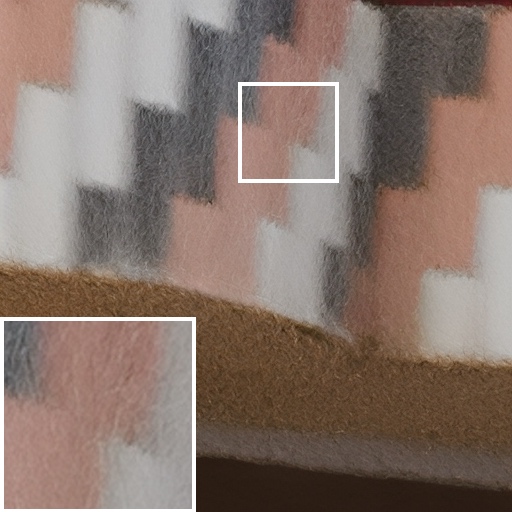} &
        \includegraphics[width=0.4\linewidth]{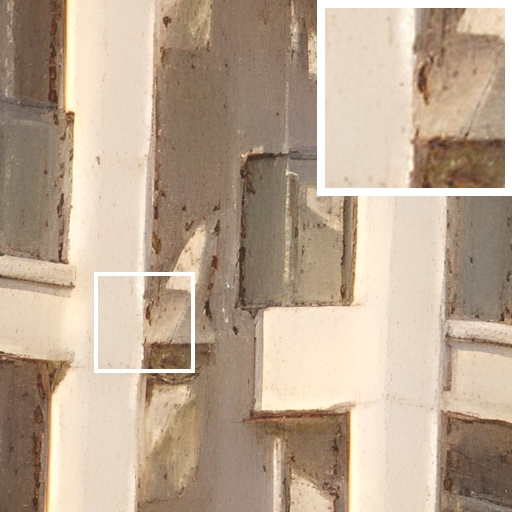} &
        \includegraphics[width=0.4\linewidth]{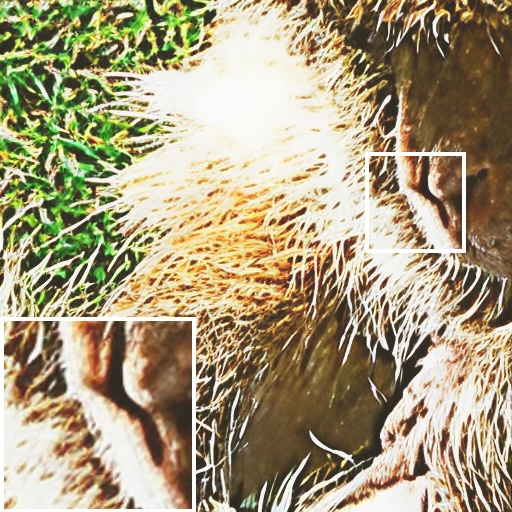}
        \\
        
        \rotatebox{90}{\quad DiffBIR~\cite{diffbir}} &
        \includegraphics[width=0.4\linewidth]{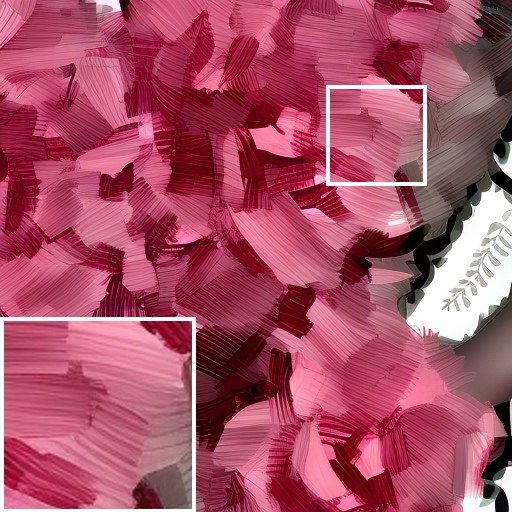} &
        \includegraphics[width=0.4\linewidth]{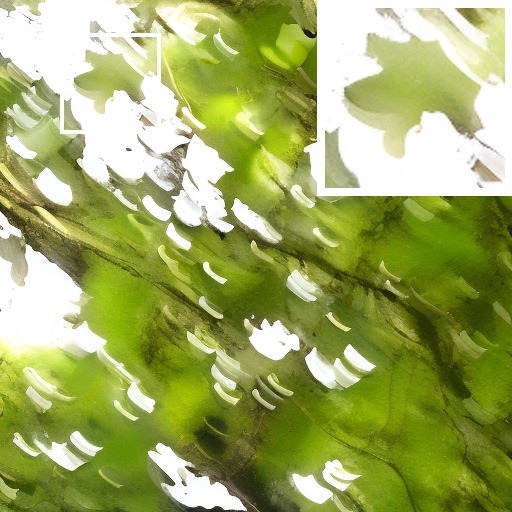} &
        \includegraphics[width=0.4\linewidth]{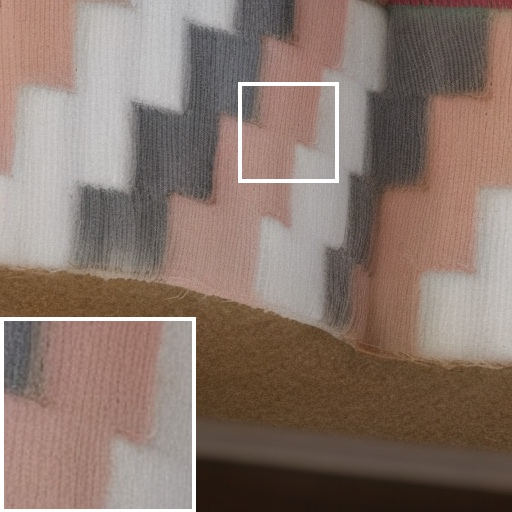} &
        \includegraphics[width=0.4\linewidth]{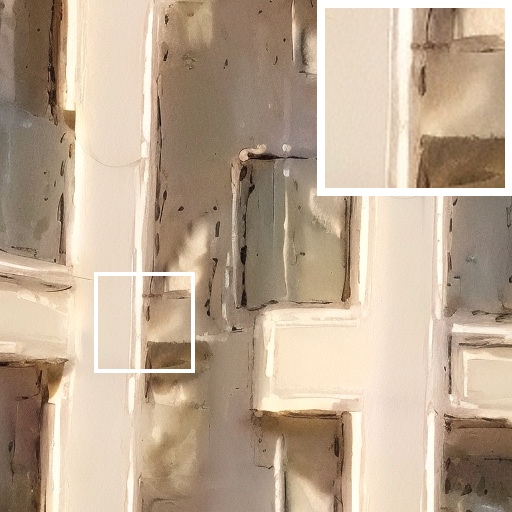} &
        \includegraphics[width=0.4\linewidth]{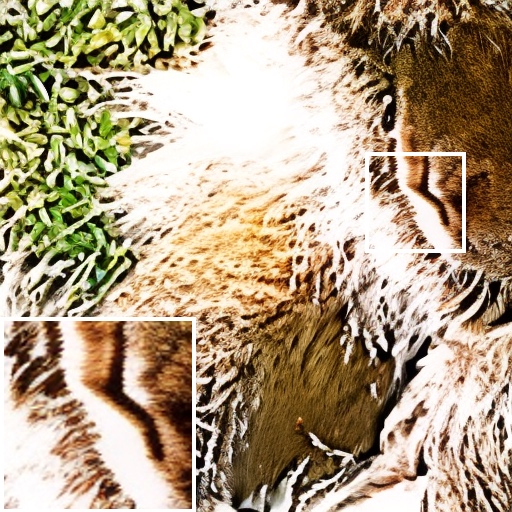}
        \\
        
        \rotatebox{90}{\quad\quad SUPIR~\cite{supir}} &
        \includegraphics[width=0.4\linewidth]{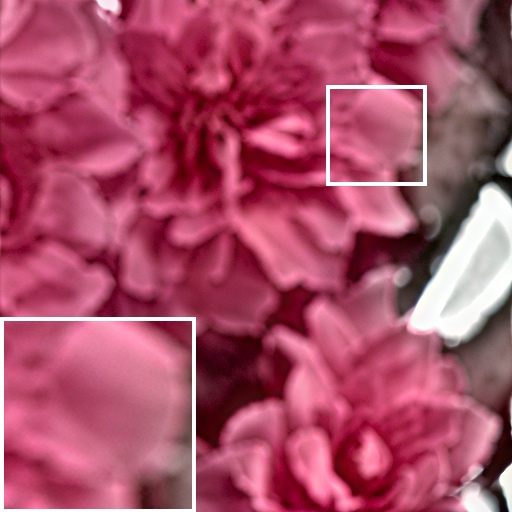} &
        \includegraphics[width=0.4\linewidth]{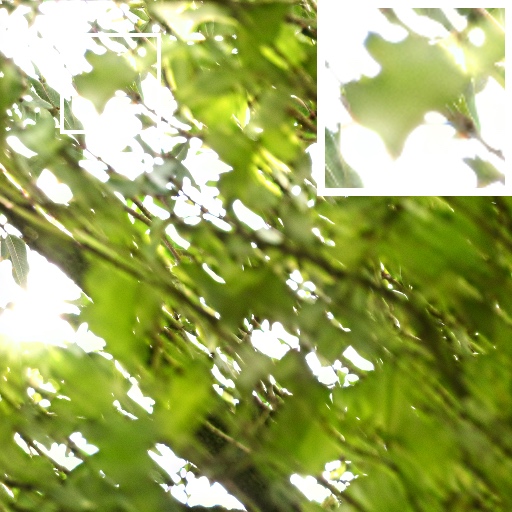} &
        \includegraphics[width=0.4\linewidth]{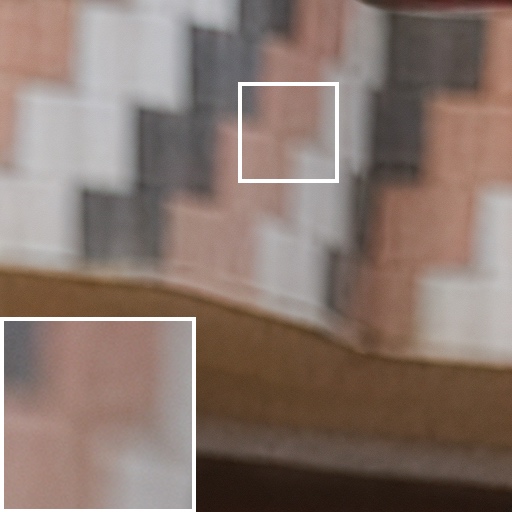} &
        \includegraphics[width=0.4\linewidth]{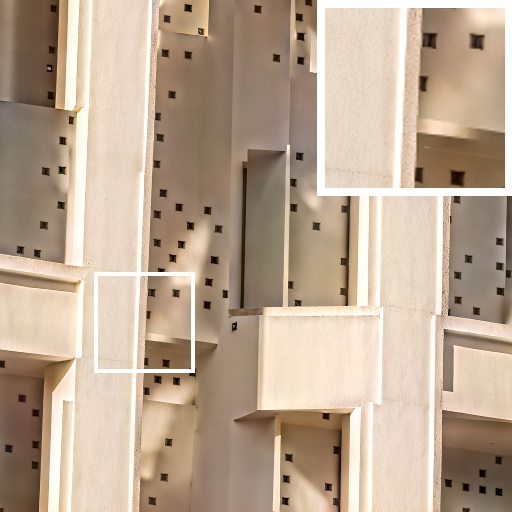} &
        \includegraphics[width=0.4\linewidth]{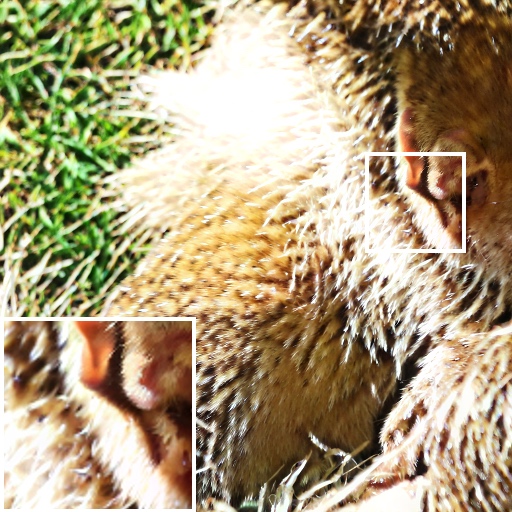}
        \\
        
        \rotatebox{90}{\,\, DACLIP-IR~\cite{daclip-uir}} &
        \includegraphics[width=0.4\linewidth]{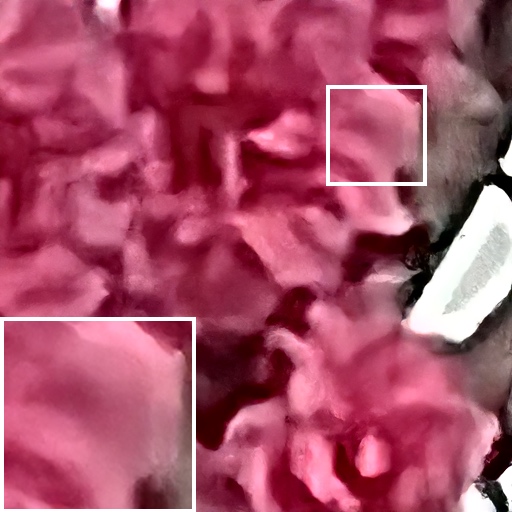} &
        \includegraphics[width=0.4\linewidth]{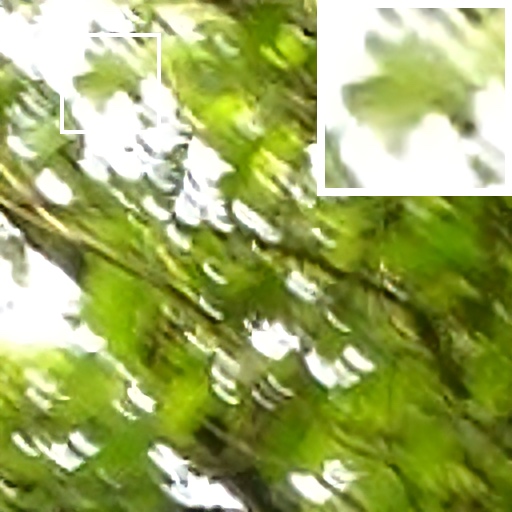} &
        \includegraphics[width=0.4\linewidth]{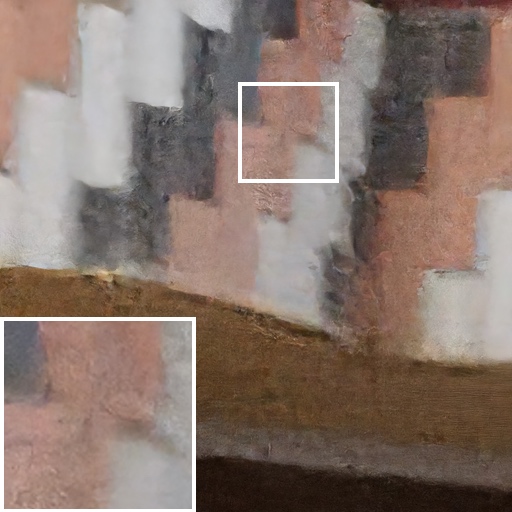} &
        \includegraphics[width=0.4\linewidth]{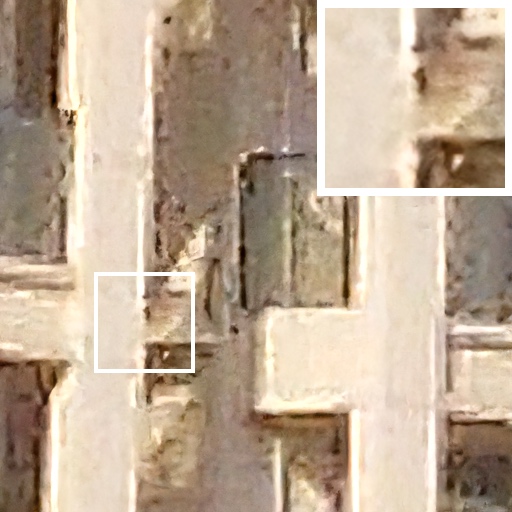} &
        \includegraphics[width=0.4\linewidth]{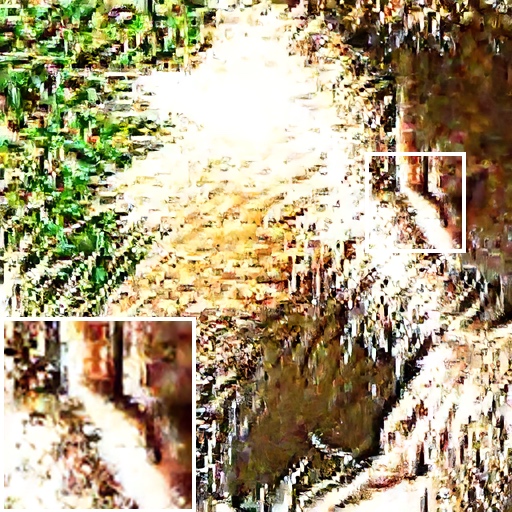}
        \\
        
        \rotatebox{90}{\qquad \quad Ours} &
        \includegraphics[width=0.4\linewidth]{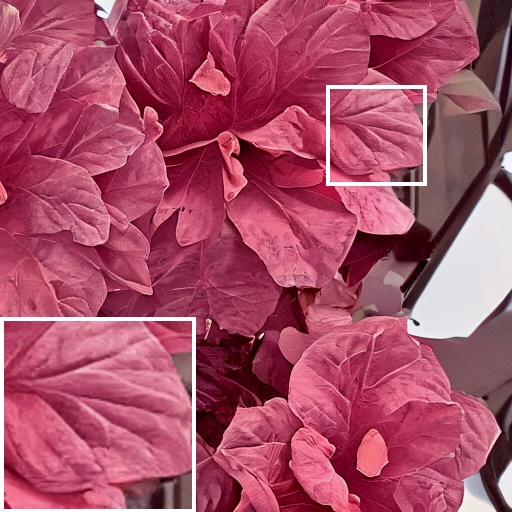} &
        \includegraphics[width=0.4\linewidth]{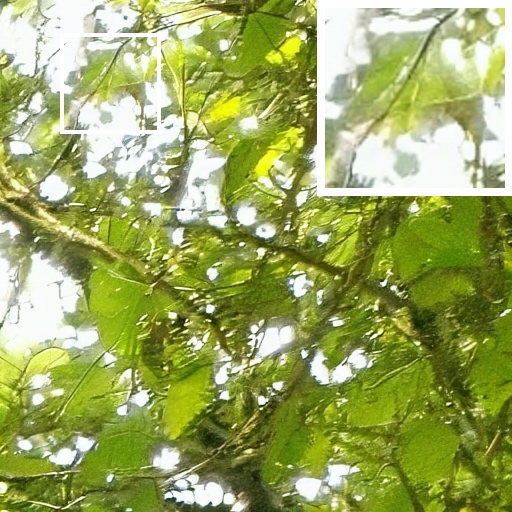} &
        \includegraphics[width=0.4\linewidth]{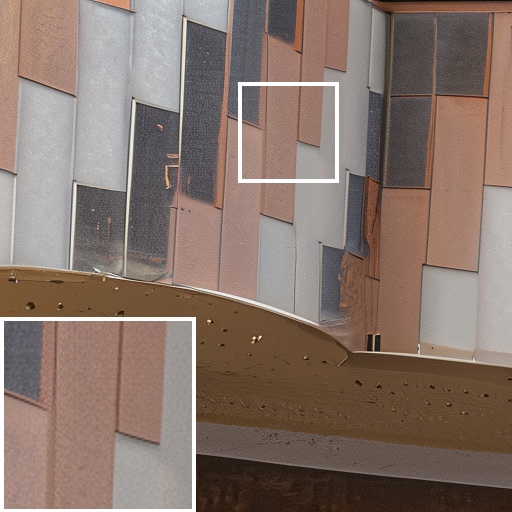} &
        \includegraphics[width=0.4\linewidth]{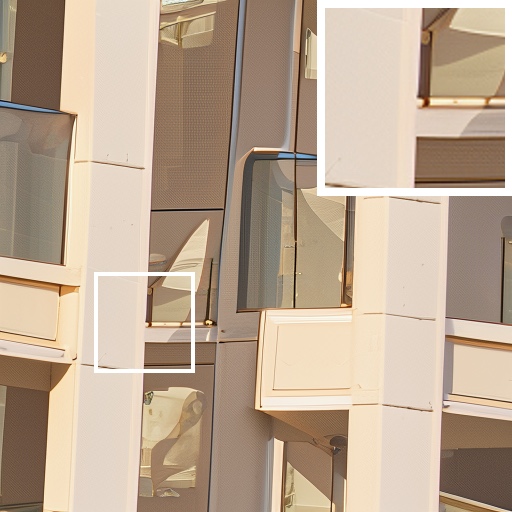} &
        \includegraphics[width=0.4\linewidth]{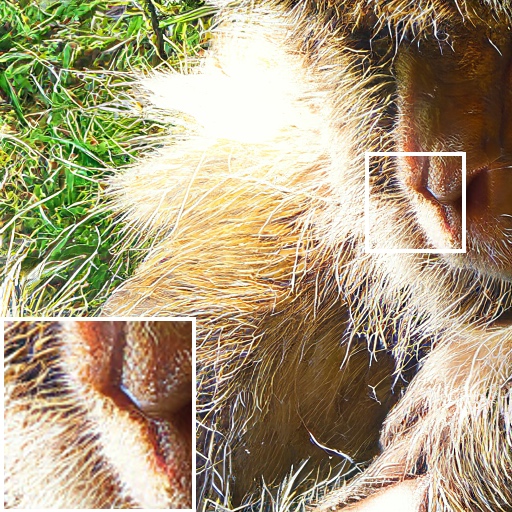} 
        %\\
        
    \end{tabular}}
    \vspace{-1em}
    \caption{Real-world image restoration on DiversePhotos$\times$4. This figure corresponds to Tab.~\ref{tab:wild4x}.
    In this case, the dominating degradation is low-resolution, but it can be accompanied with other degradations. Besides first column reflecting low resolution, second through the fifth columns contain additional motion blur,
    defocus blur, noise, and unseen degradation, respectively. Our model is robust against complex degradations. Zoom in for image details.}
    \vspace{-1em}
    \label{fig:wild4x}
\end{figure}

The visualizations of the upscaling results for both $\times 1$ and $\times 4$ are shown in Fig. \ref{fig:wild1x} and Fig. \ref{fig:wild4x}, respectively.
% !! our result and observation
% According to the visualization of DiversePhotos$\times$1 results (Fig.~\ref{fig:wild1x}), 
Fig.~\ref{fig:wild1x} shows that our proposed method is robust %against the
%complex real degradations, 
even if the test image has multiple types of degradations.
%
%It is important to point out
Note that when an image contains multiple
degradations, our model handles them simultaneously in an end-to-end
manner, without the need to iteratively restore each single degradation~\cite{airnet,restoreagent}.

% !! general observation, including the behavior of sota works.

%Among the state-of-the-art methods involved in our
%comparison,
\rev{According to the visualizations,}
SUPIR can generate impressive details. % when the dominating degradation is low resolution.
%probably due to
%their private high-quality training data and
%its much larger SDXL~\cite{sdxl} backbone which has 2.6B parameters.
%
%Nevertheless, we also observe that SUPIR always
%generates excessive wrinkles when restoring human faces,
%which might be caused by age bias in its training dataset.
%
However, when the degradation on test image is more than low resolution,
SUPIR \rev{frequently fails by making} 
objects out-of-focus (\emph{e.g.}, the 7-th
column in Fig.~\ref{fig:wild1x}).
%
%We speculate such behavior stems from a potential
%training dataset bias caused by a preference
%over images with shallow depth of field that makes some objects 
%out-of-focus. %(due to large-aperture lens).
%A shallow depth of field can make the main object stand out, but meanwhile makes the objects in the background out-of-focus with defocus blur.
%When trained with such ground-truth, the restoration model
%may inherit such bias and lead to the observed behavior.
%For example, the leaves as shown by the 7-th column in Fig.~\ref{fig:wild1x} is kept out-of-focus.
%
Besides, StableSR and DiffBIR can occasionally remove
motion blur or defocus blur, but \rev{the effect is not consistent}. %they are not as robust as our
%method for complex degradations.
%
While DACLIP-IR is an all-in-one model (including motion deblur)
particularly designed for image restoration in the wild,
it is still not robust enough against real degradations.

%
%
%Furthermore, StableSR, DiffBIR, and SUPIR, may sometimes excessively hallucinate and cause obvious inconsistency with the input low-quality image.
%
%In particular, StableSR and DiffBIR may even change
%the text even if they are readable from the low-quality
%input.
%
%This phenomenon is also reported by \cite{pasd} for
%ControlNet-based resotration methods.

When restoring images with $\times 4$ upscaling, the dominating
degradation will spontaneously become low resolution.
Nevertheless, it may accompany with other degradations, as shown in Fig.~\ref{fig:wild4x} for DiversePhotos$\times$4 test images.
We note that low-resolution images with other types of degradations can still cause failure for other methods, \rev{as shown in the first column in Fig.~\ref{fig:wild4x}}.
Conversely, our method is more robust to those complex degradations.
This is also reflected by the quantitative results in Tab.~\ref{tab:wild4x}.
The RealSR and DRealSR datasets are specifically
created with \rev{real-world low resolution degradation instead of complex degradation.}
According to Tab.~\ref{tab:wild4x}, our model achieves a balanced quality (reflected by non-reference metrics) and fidelity (reflected by full-reference metrics) \rev{for such specialized degradations.}

The above qualitative and quantitative experimental results, collectively suggests that our proposed method is effective and robust against the real and complex degradations present in the DiversePhoto test images.
Even on real test datasets specifically tailored towards low resolution, our model can still achieve a competitive performance.

\subsection{Ablation Studies \rev{and Discussions}}
\label{sec:42}

We conduct ablation studies to identify
the impact of our model components.
% and expand the discussion on several key aspects.
%
The results are presented in Tab.~\ref{tab:ablation}.

\noindent \emph{Multi-task Training.}
%
% Since every related work use different training datasets, 
To create a controlled experiment, we train a model solely for super-resolution (``SR-Only'').
% The only difference between the SR-Only model and UniRes lies in the training tasks.
%
The findings of the ``SR-Only'' group indicate that a model trained solely on the super-resolution task is less robust enough to handle complex degradations \rev{compared to UniRes}. %As a result, these models are outperformed by our UniRes model.

%(?) DownLQ Scaling Factor.
%The study from the SR-Only group also supports the conclusion that the "DownLQ" method strikes an effective balance between fidelity and quality in the generation of detail. Our analysis indicates that a $4\times$ upscaling factor is the sweet spot for achieving this balance. Furthermore, the application of "DownLQ" resulted in a notable improvement of 8.7 points in the MUSIQ score compared to SR-Only (47.76).

%\vspace{.25em}
\noindent \emph{Single-Task Inference.}
As shown by the ``Single-Task'' group in Tab.~\ref{tab:ablation},
while inference on a single task (by setting one-hot weights) can handle the corresponding degradation, one task is insufficient to cover the wide range
of complex degradations in real world, as any single task
would lead to much lower metrics compared to our default setting of UniRes. %suggested by
%the lower metrics comparing our ``SR=1'' case (second row of this group) with the default.

%\vspace{.25em}
%\vspace{-10em}
\noindent \emph{Strike-out Some Tasks.}
As suggested by the ``Strike-Out'' group,
all tasks have their own contributions in UniRes's performance.
Note, since blind restoration learns all the tasks and could behave like the
other tasks to be strike-out, we strike out blind restoration first for this ablation group (\emph{i.e.}, its corresponding
weight is fixed to $0$ during grid search),
as shown in the first row.
The DownLQ is similar to SR, so we disable it as well, as shown in the second
row in this group.
Next, we further strike out each of the four tasks involved in the training process, as shown in the rest rows.
Once any of the four task is disabled , UniRes \rev{shows an impacted restoration performance}. %not able to achieve
%the best restoration performance.
\rev{All tasks involved are needed.}

%\vspace{.25em}
\noindent \emph{DownLQ Scaling Factor.}
As discussed in Sec.~\ref{sec:32},
different down-sampling factors (See Fig.~\ref{fig:downlq}) lead to different effect
of detail generation.
This is also reflected by the results in the ``DownLQ'' group,
where a lower factor leads to less image detail generation,
and hence lower metrics.
%
%The effect of fidelity-quality trade-off
%mechanism on the metrics is consistent with related works~\cite{stablesr,diffbir,supir}.
%
Higher factors than $\times 4$ may potentially break the
balance between fidelity and quality \rev{according to observation}.
So we empirically choose $\times 4$ for a balance between fidelity and quality.
%Besides, multi-scale DownLQ could be a potential direction for
%further improvement, but in this paper we empirically choose
%$\times 4$ for a balance between fidelity and quality.

% Preview source code for paragraph 8

\begin{table}
\resizebox{\columnwidth}{!}{%
\setlength{\tabcolsep}{4pt}%
\begin{tabular}{ccccc}
\toprule 
\textbf{Ablation Group} & \textbf{Settings} & \textbf{ClipIQA} & \textbf{MUSIQ} & \textbf{ManIQA}\tabularnewline
\midrule
%\midrule 
\rowcolor{ggreen!10}\multicolumn{5}{c}{DiversePhotos$\times$1 (160 images, size $512\times512$)}\tabularnewline
\midrule 
%LQ & N/A & 0.3141 & 35.93 & 0.1909\tabularnewline
UniRes & Default (see Sec.~\ref{sec:4}) & 0.6519 & 68.22 & 0.5021 \\
\midrule 
\multirow{2}{*}{SR-Only} & SR=1 (\emph{i.e.}, super resolution) & 0.4173 & 47.76 & 0.2921\tabularnewline
 & DownLQ($\times 4$)=1 & 0.5008 & 56.48 & 0.3559\tabularnewline
\midrule
\multirow{6}{*}{Single-Task} & BR=1 (\emph{i.e.}, blind restoration)& 0.4589 & 53.49 & 0.3333\tabularnewline
 & SR=1 (\emph{i.e.}, super resolution)& 0.4640 & 53.54 & 0.3423\tabularnewline
 & MD=1 (\emph{i.e.}, motion deblur)& 0.4467 & 52.71 & 0.3196\tabularnewline
 & DD=1 (\emph{i.e.}, defocus deblur)& 0.4705 & 52.61 & 0.3538\tabularnewline
 & DN=1 (\emph{i.e.}, denoise)& 0.3744 & 39.21 & 0.2202\tabularnewline
 & DownLQ($\times 4$)=1 (see Sec.~\ref{sec:31})& 0.5610 & 61.38 & 0.4013\tabularnewline
 
\midrule
\multirow{6}{*}{Strike-Out} & Grid search w/ BR=0     & 0.6520 & 68.19 & 0.5019
\tabularnewline
 & Grid search w/ BR=0, DownLQ=0           & 0.5664 & 62.81 & 0.4302 
\tabularnewline
 & Grid search w/ BR=0, DownLQ=0, SR=0     & 0.5366 & 59.70 & 0.3959 
\tabularnewline
 & Grid search w/ BR=0, DownLQ=0, MD=0     & 0.5595 & 61.05 & 0.4273
\tabularnewline
 & Grid search w/ BR=0, DownLQ=0, DD=0     & 0.5441 & 60.63 & 0.4075 
\tabularnewline
 & Grid search w/ BR=0, DownLQ=0, DN=0     & 0.5405 & 61.41 & 0.4076
\tabularnewline

\midrule
DownLQ & DownLQ($\times 2$) & 0.4883 & 55.38 & 0.3480\tabularnewline
% & DownLQ($8\times$) & 0.5797 & 63.48 & 0.4105\tabularnewline
% & DownLQ($16\times$) & 0.6050 & 65.59 & 0.4350\tabularnewline

\midrule
\multirow{3}{*}{Search Grid} & $[\gamma,\delta]$ changed to $[0,1]$ & 0.5667 & 63.24 & 0.4154 \tabularnewline
& Most frequent $8$ sets of weights & 0.6613 & 68.02 & 0.5101 \tabularnewline
& Random Forest (skip search) & 0.5873 & 61.91 & 0.4257 \tabularnewline
 
\bottomrule
\end{tabular}}
\vspace{-1em}
\caption{Ablation studies on DiversePhotos$\times$1.
%The output size is identical to the input image size.
%For instace, we use "SR=$1.0$" to denote the
%case where the weight for super-resolution is $1.0$,
%and $0.0$ for the rest.
These ablation studies are categorized into several groups,
which are discussed one-by-one in Sec.~\ref{sec:42}.
The ``UniRes'' group based on the default setting is provided
for convenience of comparison.}
\vspace{-1em}

\label{tab:ablation}
\end{table}

\vspace{.2em}
\noindent \emph{Search Grid $[\gamma,\delta]$.}
Our default search range (see Sec.~\ref{sec:4}) effectively
enables classifier-free guidance~\cite{cfg}, which pulls
the latent prediction closer to the desired direction while
pushing away from undesired direction.
As suggested by the ``Search Grid" group in Tab.~\ref{tab:ablation}, a smaller search range $[\gamma,\delta]=[0,1]$, leads to lower performance.
We refrain from using larger search ranges because they occasionaly produce undesired artifacts according to our observation.

\vspace{0.2em}
\noindent\emph{Search Complexity.}
\rev{Grid search has a high complexity and
the search space size is $1512$ under the default settings.
Detailed discussion on this can be found in supplementary material.
The complexity can be mitigated by reducing the search space.
More specifically, using the $8$ most frequent sets of weights observed on $120$ extra images collected similar to DiversePhotos, %leads to nearly similar visual quality as compared to the grid search.
as shown in the second row of ``Search Grid'' group in Tab.~\ref{tab:ablation}, our method
can still maintain its performance in MUSIQ, while ClipIQA and ManIQA slightly fluctuate since they are not our optimization objective.
Also, worth highlighting that replacing grid search with direct prediction of the weights using a Random Forest Regressor~\cite{scikit-learn} based on the MT-A~\cite{spaq} image feature leads to a competitive performance (see the third row in the ``Search Grid'' group).
Better degradation-aware image features for weight prediction is a different topic,
and hence left for future study.}

%\vspace{-1em}
%\noindent \textbf{Supplementary Material}
In the supplementary material, we provide additional detailed discussions, visualizations, and details for reproducing the DiversePhotos
and OID-Motion datasets.
\rev{It also includes quantitative comparisons with additional image restoration methods such as \cite{autodir} due to the space limit of the manuscript and a large performance gap compared to UniRes.} % since their performance fall behind the other methods in Tab.~\ref{tab:wild1x}.}
\rev{Notably, while UniRes only focus on camera-based degradations instead of all types of degradations like all-in-one methods~\cite{autodir}, our method is still more effective on the DiversePhotos benchmark
than the state-of-the-art methods.}

\section{Conclusion}
In this paper, we introduce UniRes, a flexible diffusion-based image restoration framework for \rev{complex degradations}. We demonstrate that real-world image with complex degradations can be effectively addressed in an end-to-end manner by flexibly combining the knowledge for several well-isolated image restoration tasks. %degradation types (low resolution, motion blur, defocus blur, and noise). % By training a diffusion model on corresponding restoration tasks, in-domain knowledge is transferred to handle out-of-domain degradations through a flexible combination of weighted latent diffusion predictions.   
%
%The UniRes framework is flexible and readily extended to incorporate new image restoration tasks.
By adjusting combination weights, the model adapts to arbitrary complex degradation composed of various degradation types, leading to improved restoration across diverse scenarios.  Extensive experimental results, including evaluations on the newly curated DiversePhotos dataset for properly reflecting the complex degradation challenge, show the effectiveness of our method in handling complex real-world degradations.    

%\noindent \textbf{Limitations.} Optimizing combination weights via grid search has an exponential cost. While UniRes is robust against complex real-world degradations, occasional failures occur, as illustrated in the supplementary material. Future work could explore machine learning models to predict combination weights based on degradation-aware image features.

\vspace{1em}
\noindent\textbf{Acknowledgement} We thank Mojtaba Ardakani for insightful discussions on diffusion models.

% \noindent \textbf{Limitations.}
% %
% The cost for combination weight optimization via
% grid search grows exponentially.
% %
% One potential solution to this issue is to engineer a simple
% machine learning model that predicts the grid search result
% based on degradation-aware image features such as \cite{daclip,spaq}.
% %DA-CLIP~\cite{daclip} embedding
% %and MT-A~\cite{spaq} attributes.
% %
% We leave this for future work.
% %
% Besides, while our method is more robust than the other methods
% against complex real-world degradations, it still occasionally fail to restore the image. Some failure cases are shown in the supplementary material.

% \section{Conclusion}

% Real-world image degradations can be approximately decomposed into down-sampling, motion blur, defocus blur, and noise.
% After training a diffusion model on their corresponding restoration tasks,
% the in-domain knowledge can be transferred to out-of-domain real-world image
% restoration through a flexible combination of weighted latent diffusion predictions.
% Experimental results manifest our model's
% effectiveness against complex real-world degradations.

%%%%%%%%%%%%%%%%%%%%%%%%%%%%%%%%%%%%%%%%%%%%%%%%%%%%%%%%%%%%%%%%%%%%
{
    \small
    \bibliographystyle{ieeenat_fullname}
    \bibliography{arxiv}
}

\appendix

\section{More Experiments and Discussions}

\subsection{Why Specifically Four Degradation Types?}

In this paper, we particularly focus on complex degradation, an arbitrary
mixture of four fundamental degradation types: low resolution, motion blur, defocus blur, and real noise.
Those degradations stem from capture condition, capture device and post-processing pipelines.
This background is clarified in the manuscript, including the abstract and the first paragraph
of the introduction section.

Apart from the four types of degradations, in the low-level vision literature,
there are other types of degradations such as rain, haze, fog, and snow.
These degradations are not caused by capture device or post-processing pipelines, and hence
are not included in the scope of this paper.
The effectiveness of our method on these types of degradations is left for future study.

\subsection{Time Complexity and Search Space}

As mentioned in the "Implementation Details" in the paper, the default search space parameters for
UniRes are $(\gamma,\delta)=(-0.2,1.2)$, with an interval of $0.2$.
Namely, each weight $w_i$ has $n=7$ possible values (\emph{i.e.}, $-0.2, 0.0, 0.2, 0.4, 0.6, 0.8, 1.0, 1.2$).
Additionally, the weights should sum to one, \emph{i.e.}, $\sum_{i=1}^K w_i = 1.0$, and only one
negative value is allowed among $w_i$, $i=1,\ldots,K$.
While the complexity of the grid search algorithm is $O(n^K)$, the concrete size of the
search space is not $n^K$ due to the two constraints.
The search space size in the default settings is $1512$.
We provide a Python snippet below for the search space and contraints.

\begin{lstlisting}
from typing import *
import numpy as np
import itertools as it

def search_grid(vmin: float = -0.2,
                vmax: float =  1.2,
                nvars: int = 6,
                interval: float = 0.2,
                ) -> List[List[float]]:
    """
    Find all valid possible combination weights.
    """
    values = np.arange(vmin, vmax + 1e-3, interval)
    allcombs = it.product(*([values] * nvars))
    allcombs = [np.array(x) for x in allcombs]
    # figure out valid combinations
    validcombs = []
    for x in allcombs:
        if not np.abs(1 - np.sum(x)) < 1e-5:
            # they must sum to one
            continue
        elif not np.count_nonzero(x < -1e-5) <= 1:
            # no more than one negative value
            continue
        else:
            # this one is valid
            validcombs.append(x.tolist())
    print('Valid Combinations:', len(validcombs))
    return validcombs

if __name__ == '__main__':
    validcombs = search_grid(-0.2, 1.2)
\end{lstlisting}

\subsection{Detailed Results on DiversePhotos$\times 1$}

\begin{table*}
\resizebox{\linewidth}{!}{%
\begin{tabular}{cc|cc|ccc}
\toprule
\textbf{Method} & \textbf{Combination Weights} & \textbf{Platform} & \textbf{Inference Time per image (seconds)} & \textbf{ClipIQA} & \textbf{MUSIQ} & \textbf{ManIQA} \\
\midrule
SwinIR~\cite{swinir} & N/A & PyTorch/Nvidia A100 & $0.374 \pm 0.063$ & 0.3727 & 49.26 & 0.3008 \\
Restormer~\cite{restormer} & N/A & PyTorch/Nvidia A100 & $0.132\pm 0.035$ & 0.3407 & 41.80 & 0.2243 \\
PromptIR~\cite{promptir} & N/A & PyTorch/Nvidia A100 & $0.136\pm 0.032$ & 0.3069 & 36.20 & 0.1950 \\
AirNet~\cite{airnet} & N/A & PyTorch/Nvidia A100 & $0.074\pm 0.033$ & 0.3031 & 35.93 & 0.1893 \\
AutoDIR~\cite{autodir} & N/A & PyTorch/Nvidia A100 & $ 10.633\pm 15.909$ & 0.3260 & 40.32 & 0.2147 \\
NAFNet~\cite{nafnet} & N/A & PyTorch/Nvidia A100 & $0.023\pm 0.007$ & 0.3372 & 43.62 & 0.2323 \\
StableSR~\cite{stablesr} & N/A & PyTorch/Nvidia A100 & $11.002 \pm 0.171$ & 0.6277 & 61.39 & 0.3992\\
DiffBIR~\cite{diffbir} & N/A & PyTorch/Nvidia A100 & $6.522 \pm 0.034$ & 0.6453 & 59.97 & 0.4922\\
SUPIR~\cite{supir} & N/A & PyTorch/Nvidia A100 & $15.601 \pm 0.629$ & 0.5060 & 51.68 & 0.3745\\
DACLIP-IR~\cite{daclip-uir} & N/A & PyTorch/Nvidia A100 & $4.940\pm 0.064$& 0.3497 & 46.16 & 0.2567\\
\midrule
UniRes & Grid search & JAX/TPUv5 & $(2.332 + 0.1)\times1512\approx 3677$ & 0.6519 & 68.22 & 0.5021\\
UniRes & Most frequent 8 sets of combination weights & JAX/TPUv5 & $(2.332+0.1)\times 8= 19.456$ & 0.6613 & 68.02 & 0.5101 \\
UniRes & Most frequent 6 sets of combination weights & JAX/TPUv5 & $(2.332+0.1)\times 6= 14.592$ & 0.6633 & 67.92 & 0.5096 \\
UniRes & Most frequent 4 sets of combination weights & JAX/TPUv5 & $(2.332+0.1)\times 4= 9.728$ & 0.6655 & 67.68 & 0.5095 \\
UniRes & Most frequent 2 sets of combination weights & JAX/TPUv5 & $(2.332+0.1)\times 2= 4.864$ & 0.6581 & 66.89 & 0.5052 \\
UniRes & Most frequent 1 set of combination weights & JAX/TPUv5 & $(2.332+0.1)\times 1= 2.432$ & 0.6590 & 66.44 & 0.5042 \\
UniRes & Average optimal combination weights & JAX/TPUv5 & $(2.332+0.1)\times 1= 2.432$ & 0.5941 & 62.10 & 0.4266 \\
UniRes & Random Forest (skip search) & JAX/TPUv5 & $0.035 + 2.332 = 2.367$ & 0.5873 & 61.91 & 0.4257\\
\bottomrule
\end{tabular}}
\caption{Full Quantitative Experimental Details on DiversePhotos$\times 1$.}
\label{tab:fulldpx1}
\end{table*}

The grid search algorithm for optimization has an exponential complexity.
The search space size given the default setting of UniRes is $1512$.
The inference time of UniRes per image for a given set of combination weights is $2.332{\pm}0.005s$ on JAX/TPUv5.
MUSIQ takes $0.1s$ per image on CPU.
The full experimental details on DiversePhotos$\times 1$, including the total inference time
per image is shown in Tab.~\ref{tab:fulldpx1}.
Two potential speed-up methods are discussed in the manuscript, and they are also included in
this table.
Some other related works, including AutoDIR~\cite{autodir} are
not compared in the paper due to space limit, and their performance lagging behind the
other methods such as DiffBIR~\cite{diffbir} and SUPIR~\cite{supir} by a margin.
The PromptIR~\cite{promptir} is advertized as ``all-in-one'' image resotration,
but the official model only support denoise, derain, and dehaze.

Potential future directions for accelerating the proposed method includes, but are not
limited to (1) distillation for single-step inference,
(2) caching mechanisms,
(3) better degradation-aware image features and combination weight prediction.
They are beyond the scope of this paper, so we leave them for future explorations.

\subsection{More Visualizations and Failure Cases}

\begin{figure*}[t]
\setlength{\tabcolsep}{1pt}
\renewcommand{\arraystretch}{0.2} % Adjust the value as needed
\centering
    \begin{tabular}{c|ccccc}
    LQ & StableSR & DiffBIR & SUPIR & DACLIP-IR & Ours\\
    \midrule
    
    \includegraphics[width=0.16\linewidth]{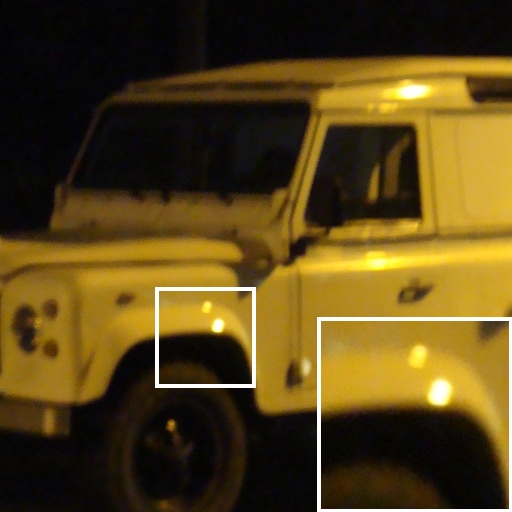} &
    \includegraphics[width=0.16\linewidth]{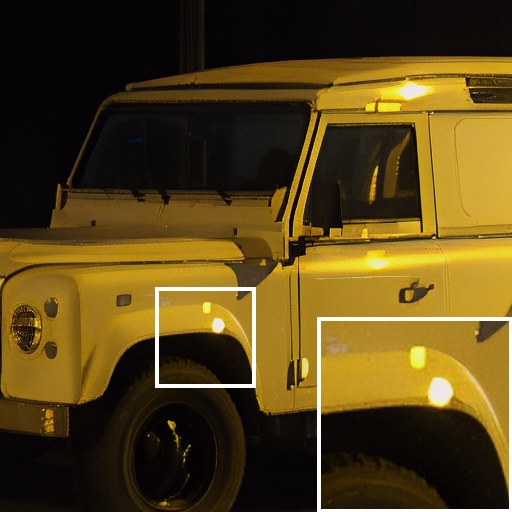} &
    \includegraphics[width=0.16\linewidth]{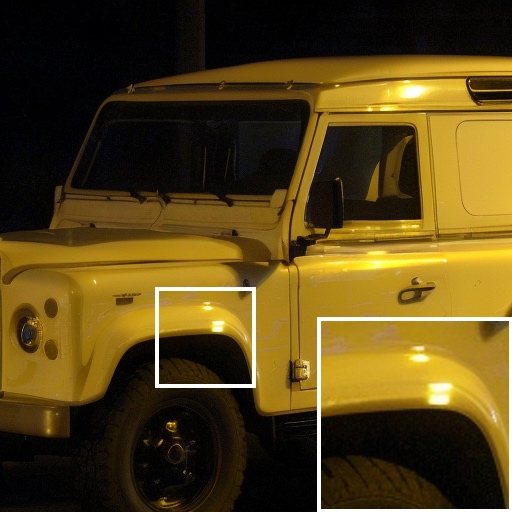} &
    \includegraphics[width=0.16\linewidth]{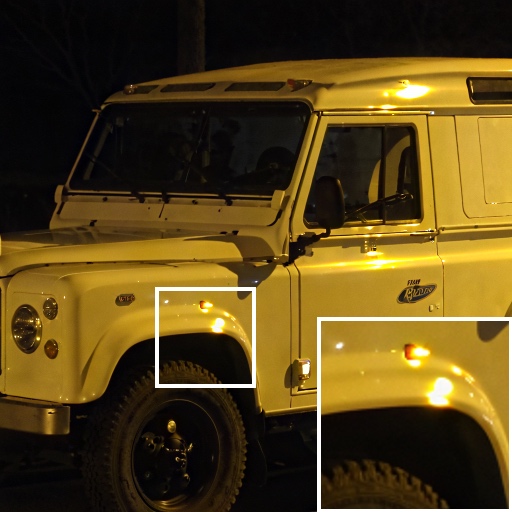} &
    \includegraphics[width=0.16\linewidth]{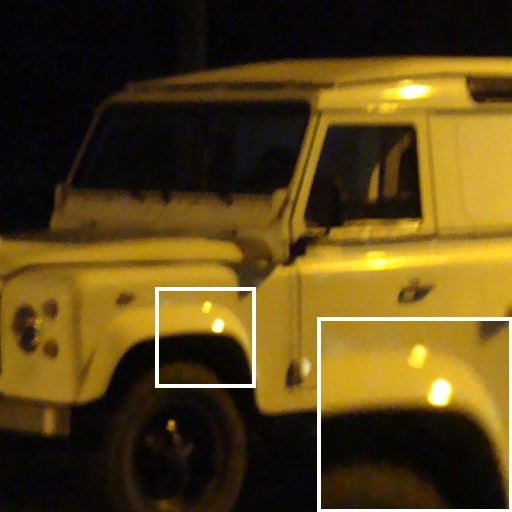} &
    \includegraphics[width=0.16\linewidth]{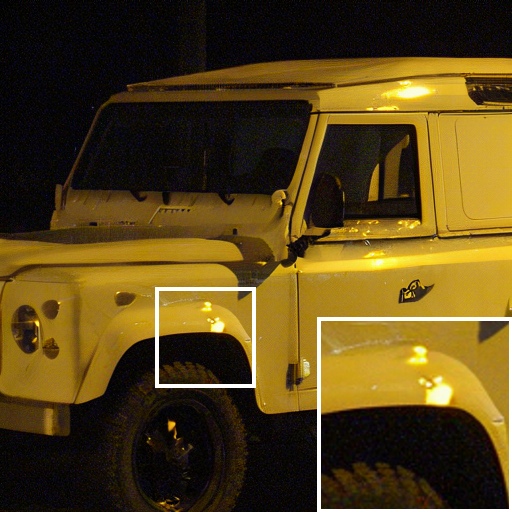} \\

    \includegraphics[width=0.16\linewidth]{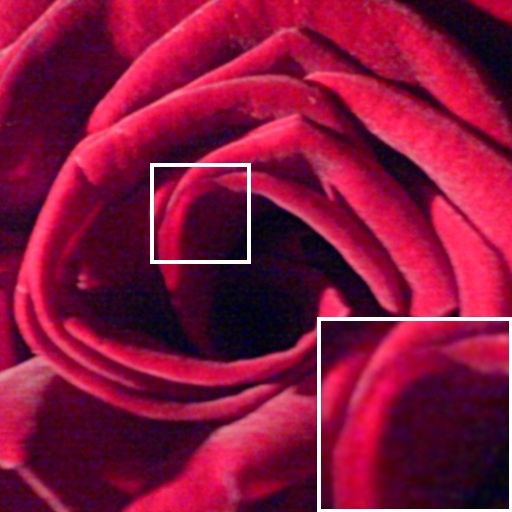} &
    \includegraphics[width=0.16\linewidth]{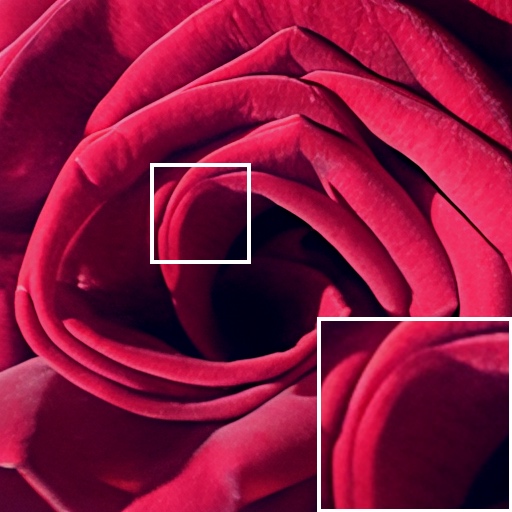} &
    \includegraphics[width=0.16\linewidth]{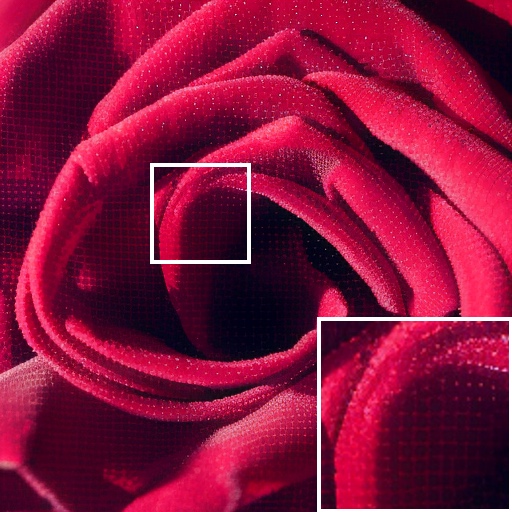} &
    \includegraphics[width=0.16\linewidth]{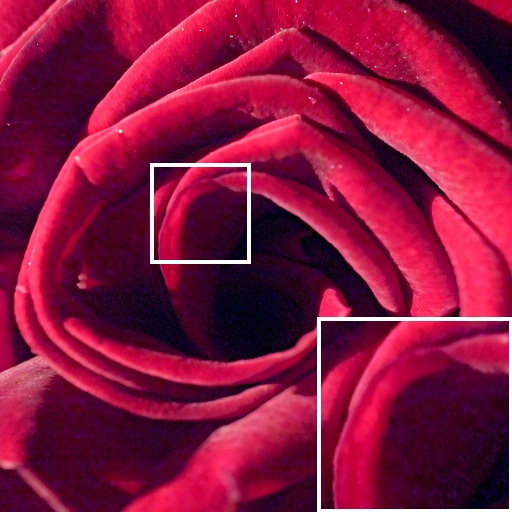} &
    \includegraphics[width=0.16\linewidth]{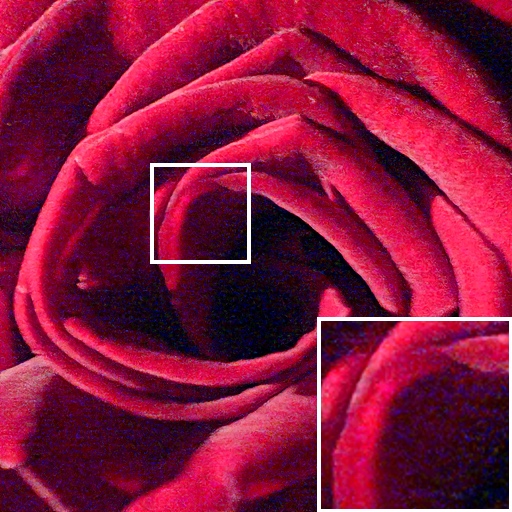} &
    \includegraphics[width=0.16\linewidth]{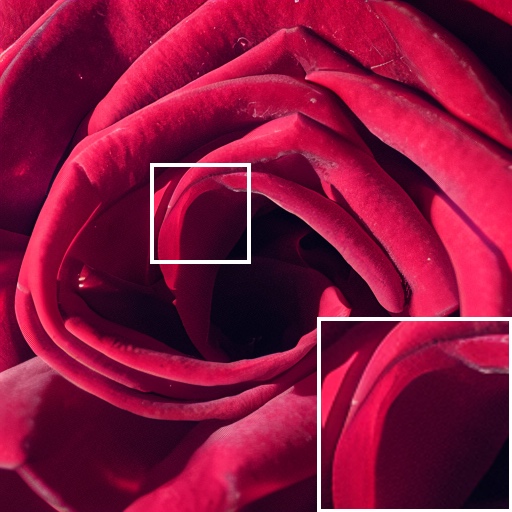} \\
    
    \includegraphics[width=0.16\linewidth]{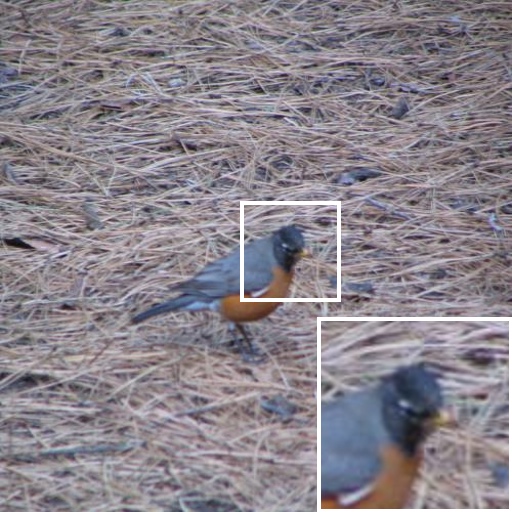} &
    \includegraphics[width=0.16\linewidth]{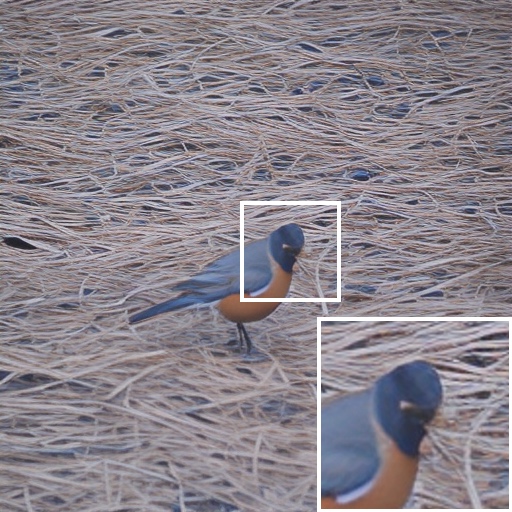} &
    \includegraphics[width=0.16\linewidth]{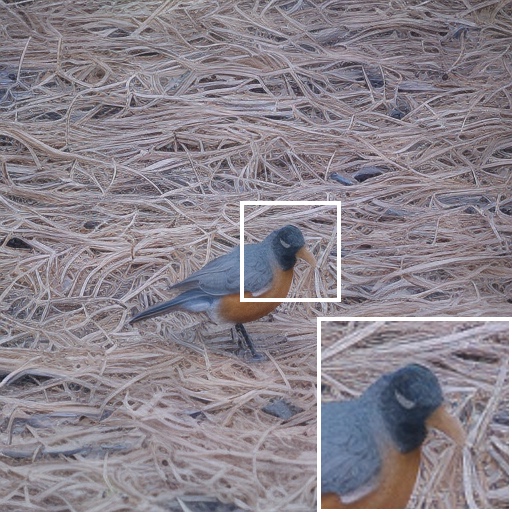} &
    \includegraphics[width=0.16\linewidth]{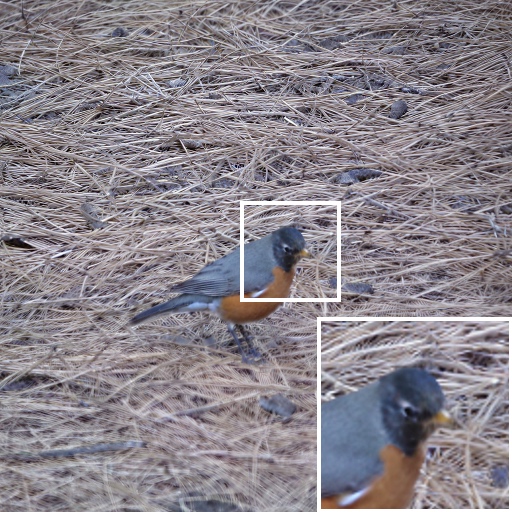} &
    \includegraphics[width=0.16\linewidth]{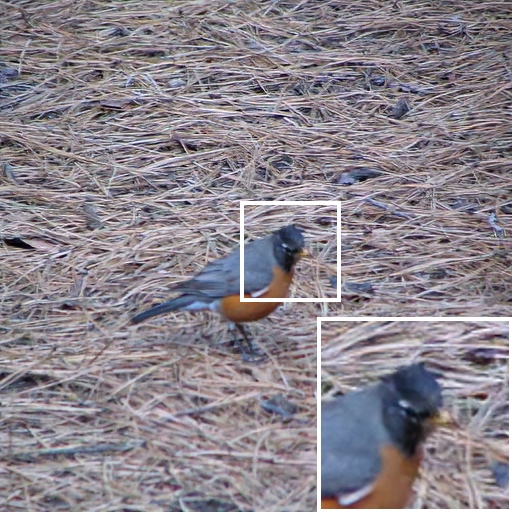} &
    \includegraphics[width=0.16\linewidth]{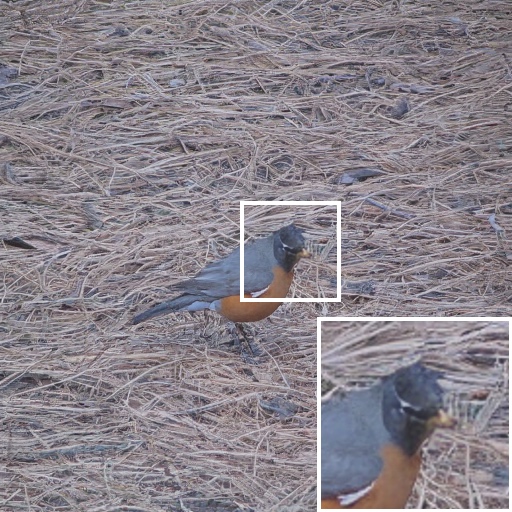} \\
    
    \includegraphics[width=0.16\linewidth]{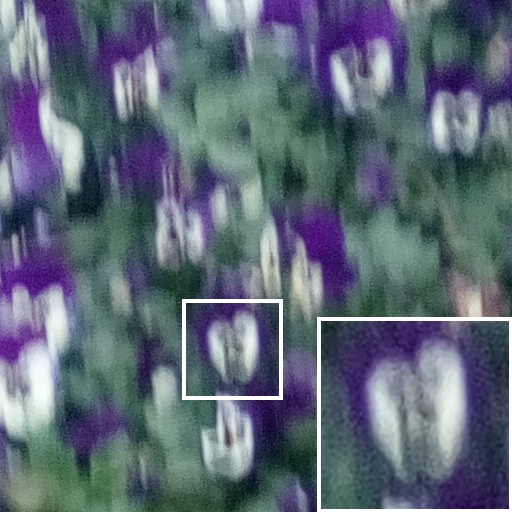} &
    \includegraphics[width=0.16\linewidth]{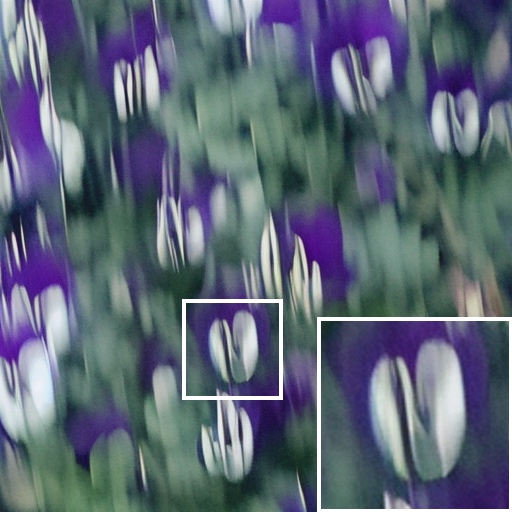} &
    \includegraphics[width=0.16\linewidth]{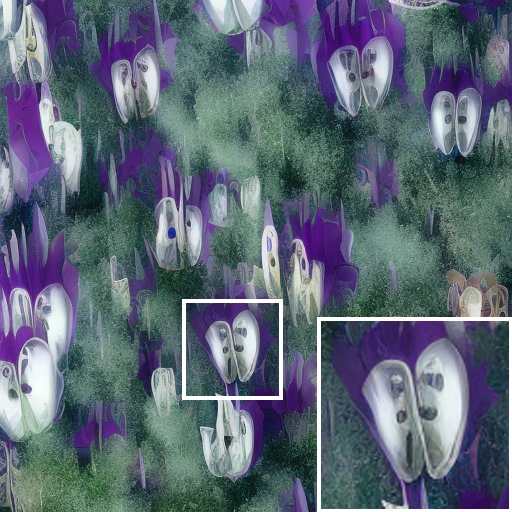} &
    \includegraphics[width=0.16\linewidth]{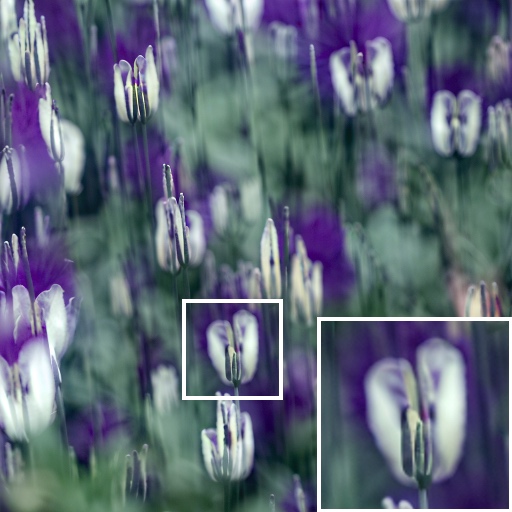} &
    \includegraphics[width=0.16\linewidth]{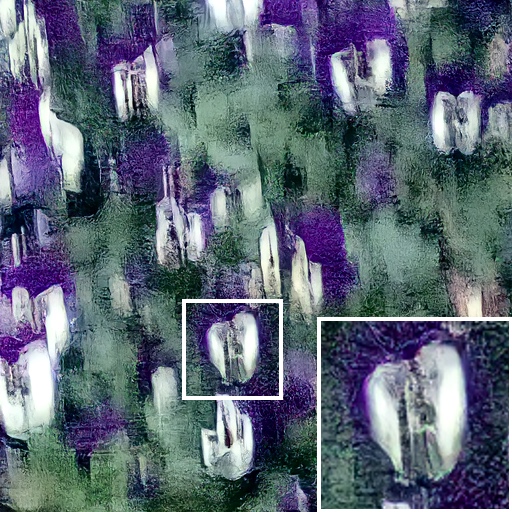} &
    \includegraphics[width=0.16\linewidth]{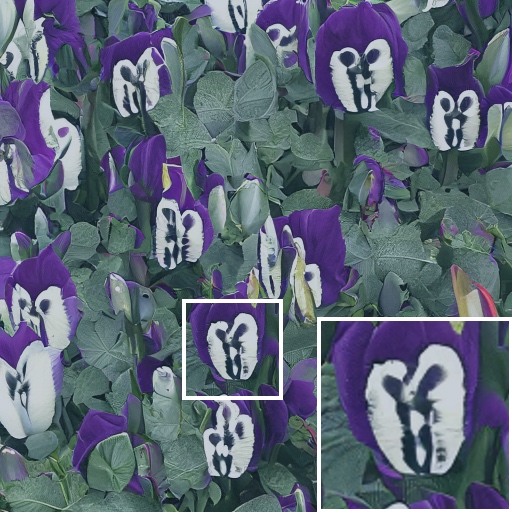} \\

    \includegraphics[width=0.16\linewidth]{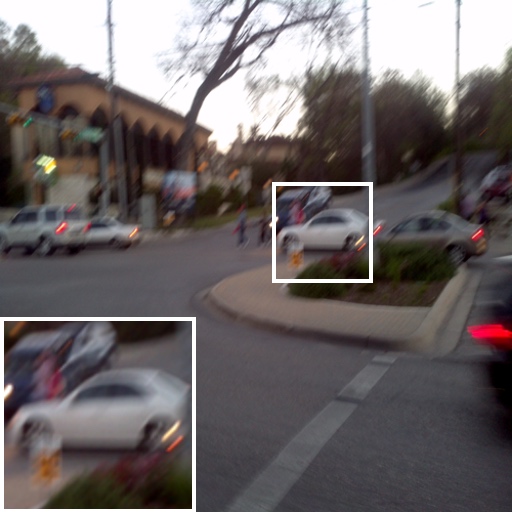} &
    \includegraphics[width=0.16\linewidth]{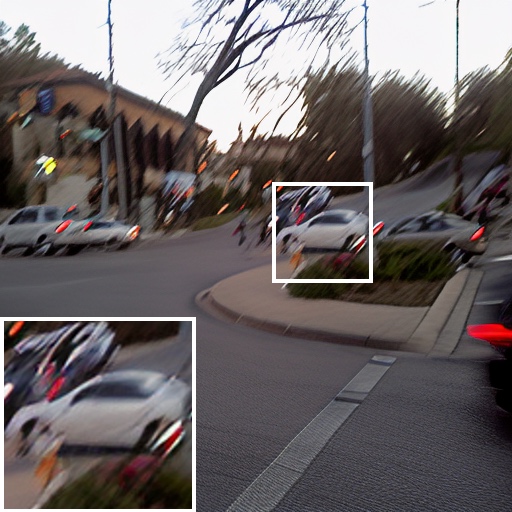} &
    \includegraphics[width=0.16\linewidth]{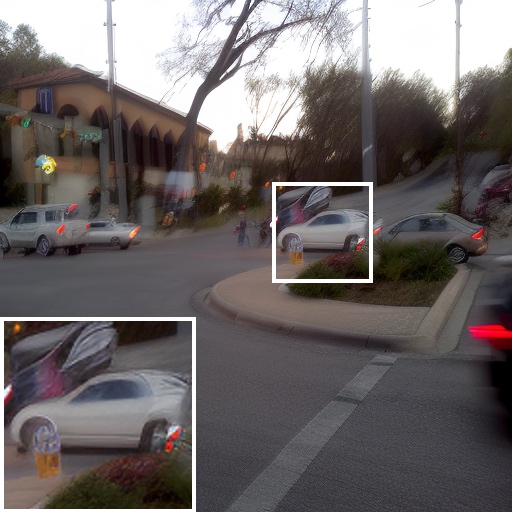} &
    \includegraphics[width=0.16\linewidth]{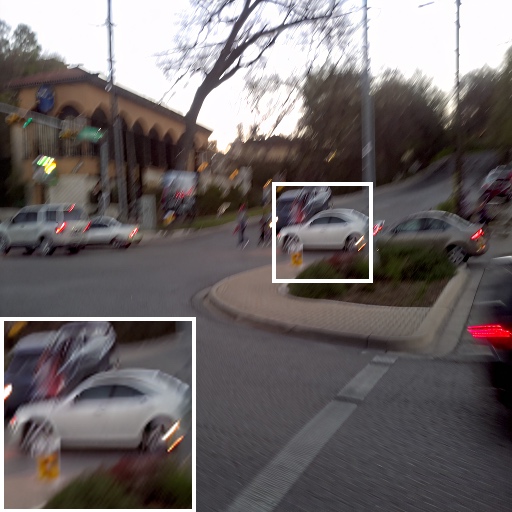} &
    \includegraphics[width=0.16\linewidth]{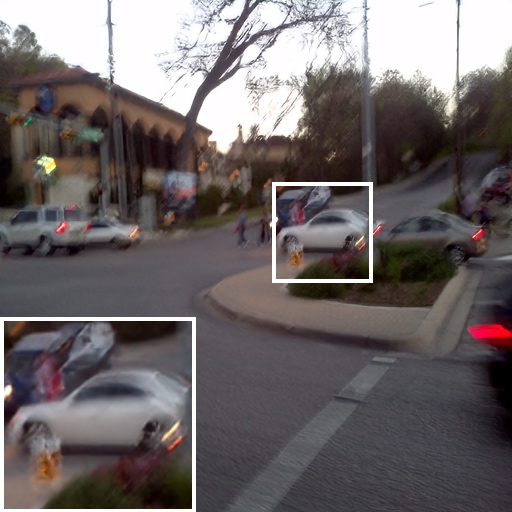} &
    \includegraphics[width=0.16\linewidth]{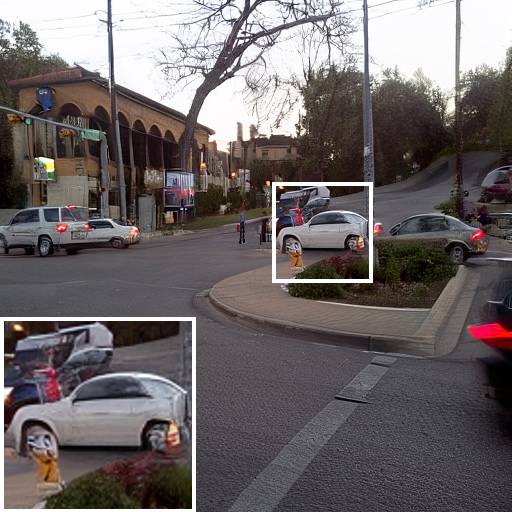} \\
    
    \includegraphics[width=0.16\linewidth]{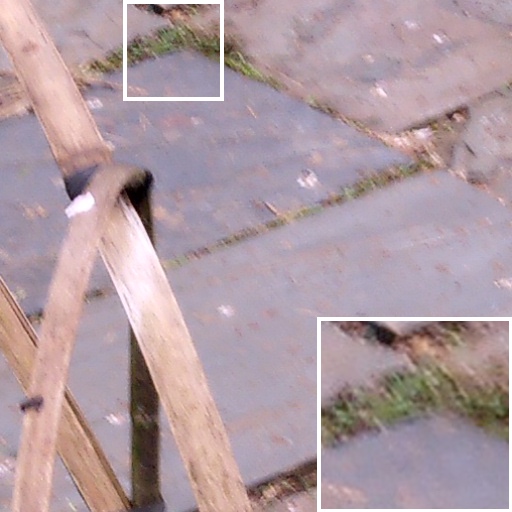} &
    \includegraphics[width=0.16\linewidth]{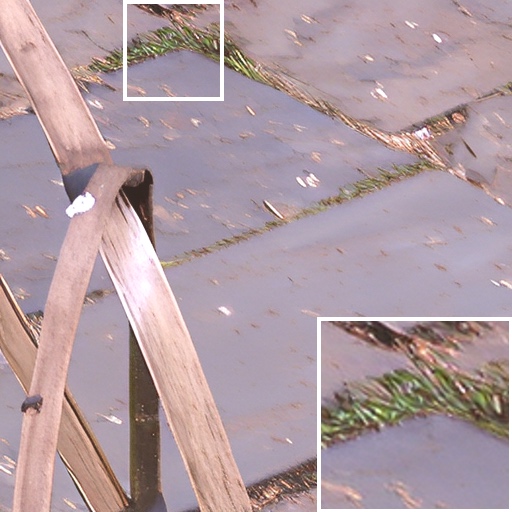} &
    \includegraphics[width=0.16\linewidth]{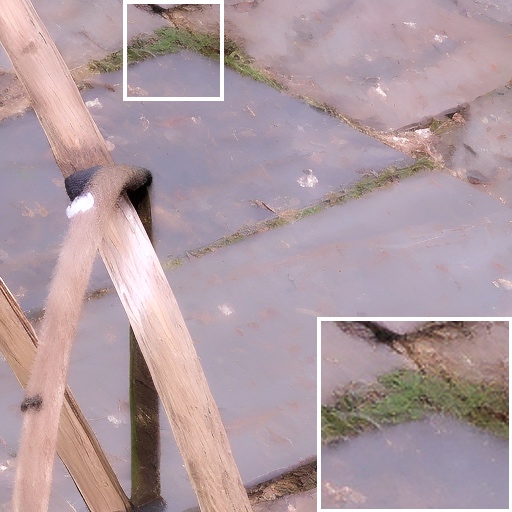} &
    \includegraphics[width=0.16\linewidth]{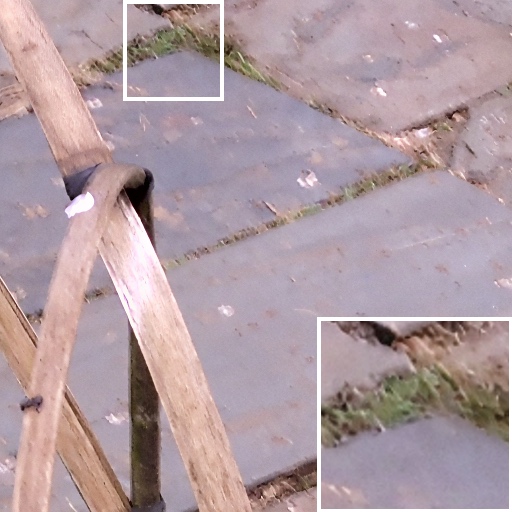} &
    \includegraphics[width=0.16\linewidth]{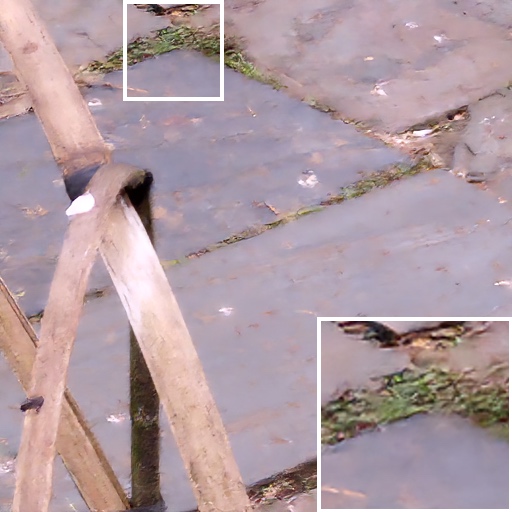} &
    \includegraphics[width=0.16\linewidth]{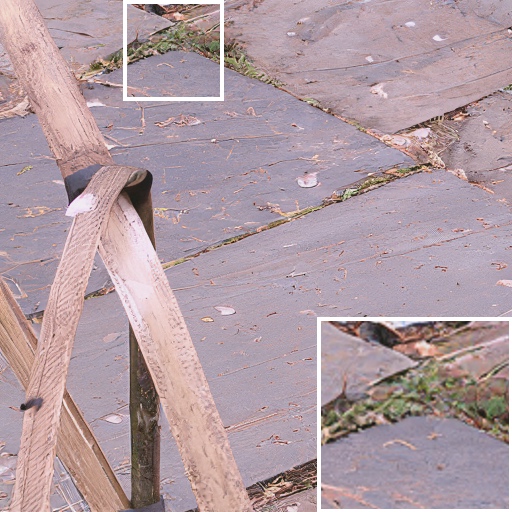} \\

    \end{tabular}
    \caption{More visualizations about real-world image restoration on the DiversePhotos$\times 1$ dataset.}
    \label{fig:moredpx1}
\end{figure*}

\begin{figure*}[t]
\setlength{\tabcolsep}{1pt}
\renewcommand{\arraystretch}{0.2} % Adjust the value as needed
\centering
    \begin{tabular}{c|ccccc}
    LQ & StableSR & DiffBIR & SUPIR & DACLIP-IR & Ours\\
    \midrule
    
    \includegraphics[width=0.16\linewidth]{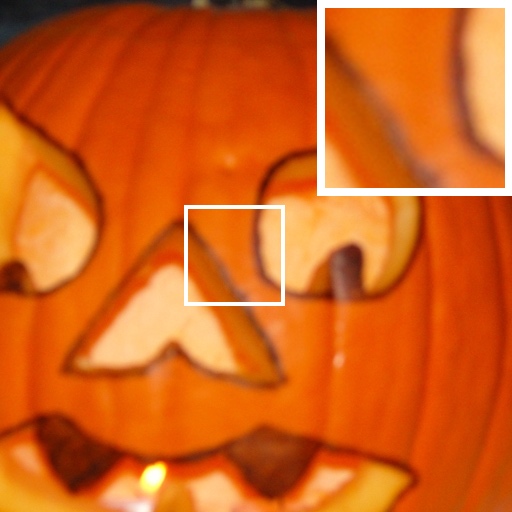} &
    \includegraphics[width=0.16\linewidth]{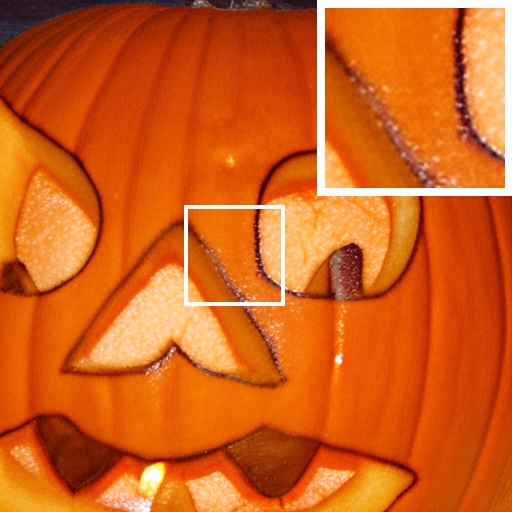} &
    \includegraphics[width=0.16\linewidth]{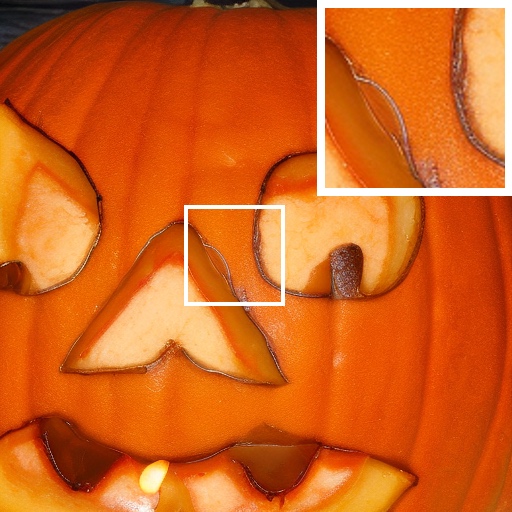} &
    \includegraphics[width=0.16\linewidth]{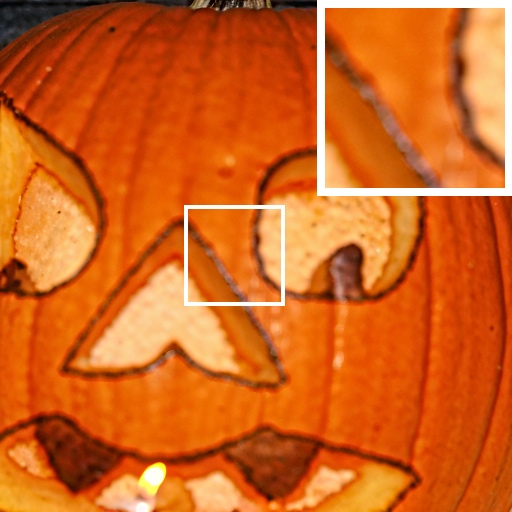} &
    \includegraphics[width=0.16\linewidth]{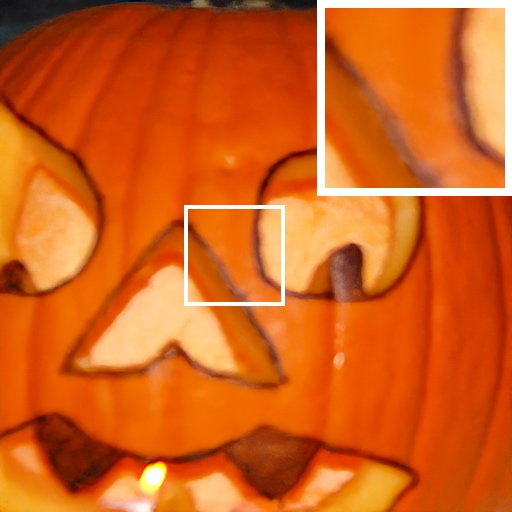} &
    \includegraphics[width=0.16\linewidth]{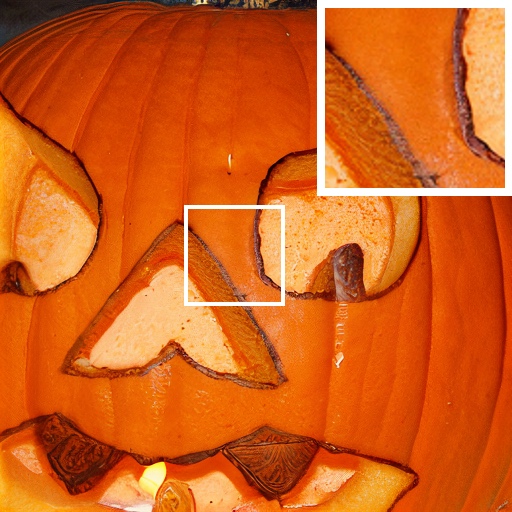} \\

    \includegraphics[width=0.16\linewidth]{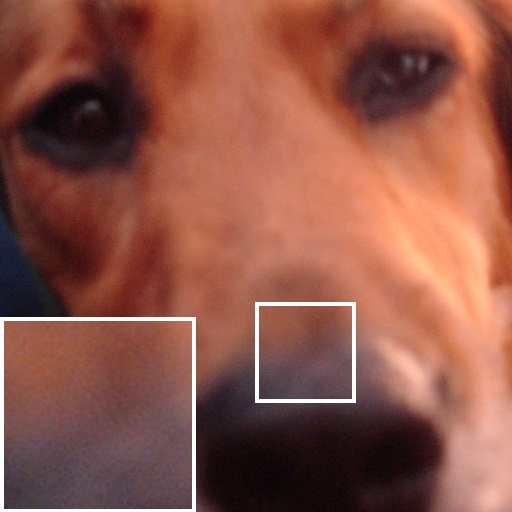} &
    \includegraphics[width=0.16\linewidth]{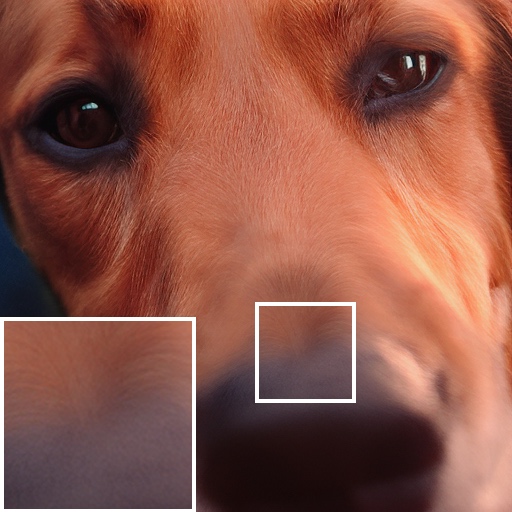} &
    \includegraphics[width=0.16\linewidth]{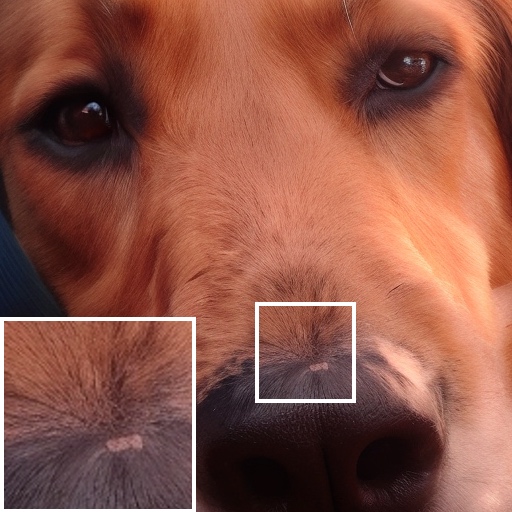} &
    \includegraphics[width=0.16\linewidth]{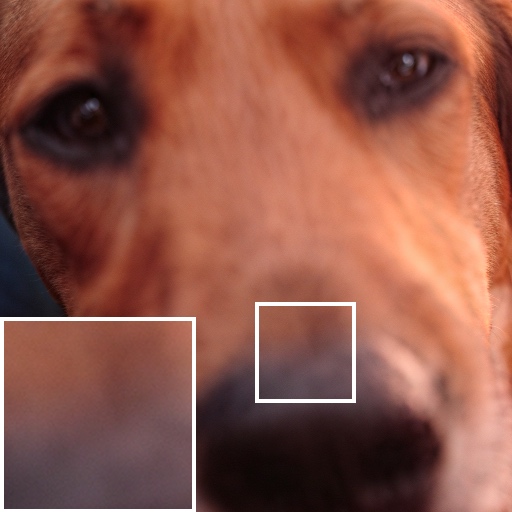} &
    \includegraphics[width=0.16\linewidth]{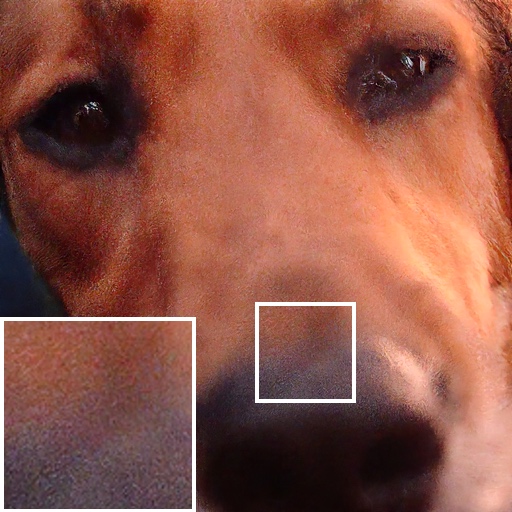} &
    \includegraphics[width=0.16\linewidth]{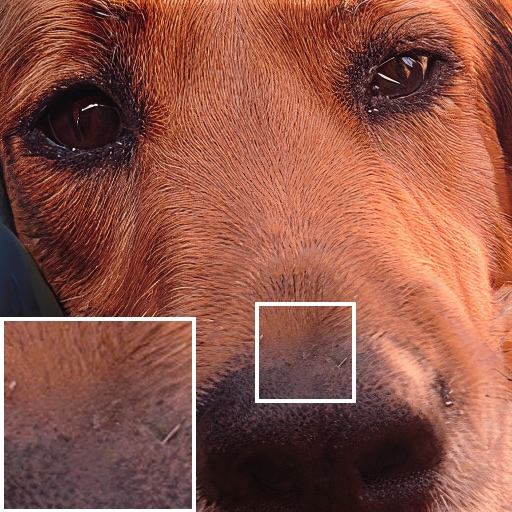} \\

    \includegraphics[width=0.16\linewidth]{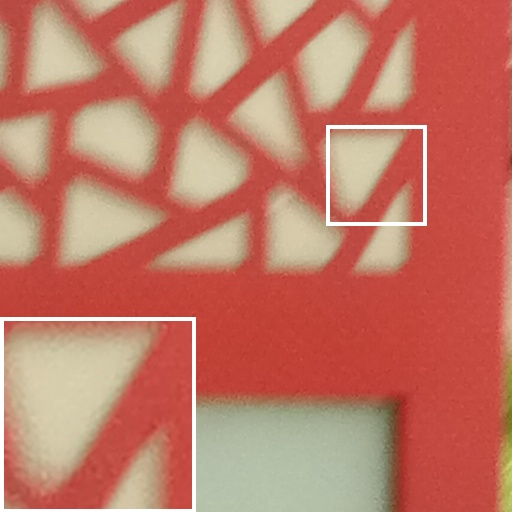} &
    \includegraphics[width=0.16\linewidth]{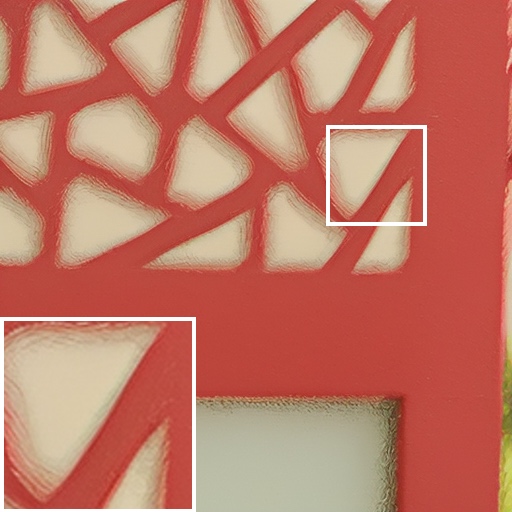} &
    \includegraphics[width=0.16\linewidth]{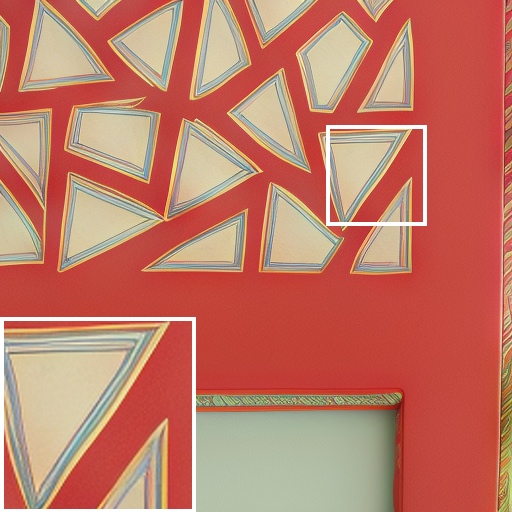} &
    \includegraphics[width=0.16\linewidth]{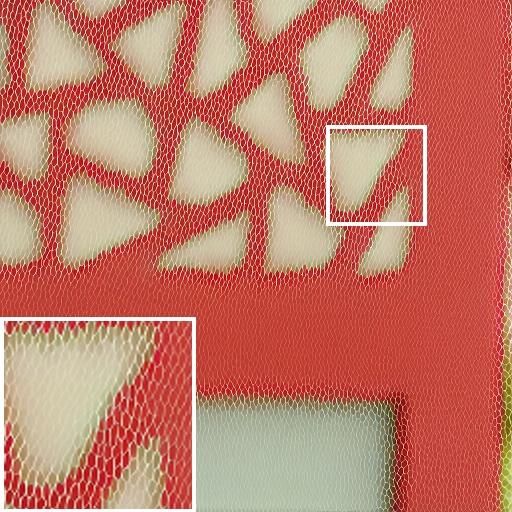} &
    \includegraphics[width=0.16\linewidth]{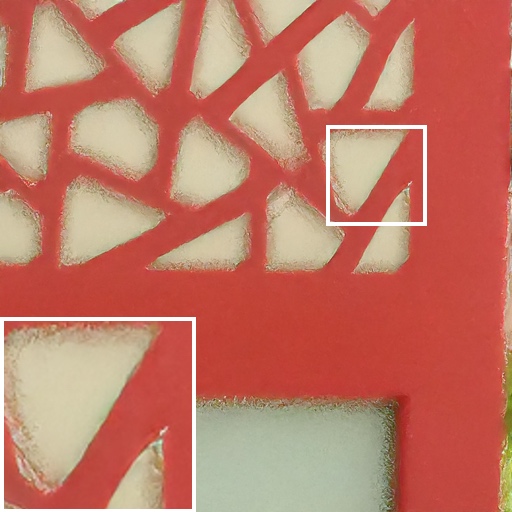} &
    \includegraphics[width=0.16\linewidth]{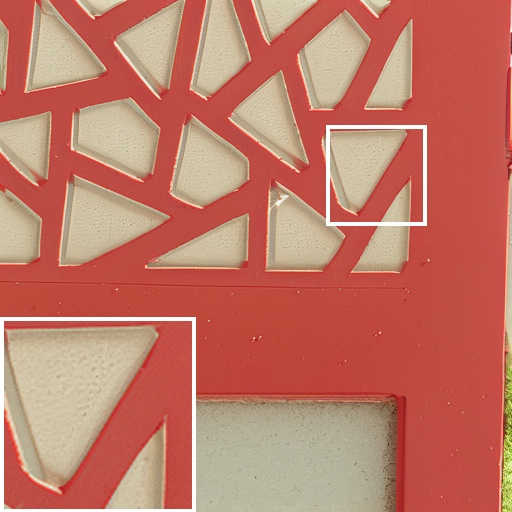} \\

    \includegraphics[width=0.16\linewidth]{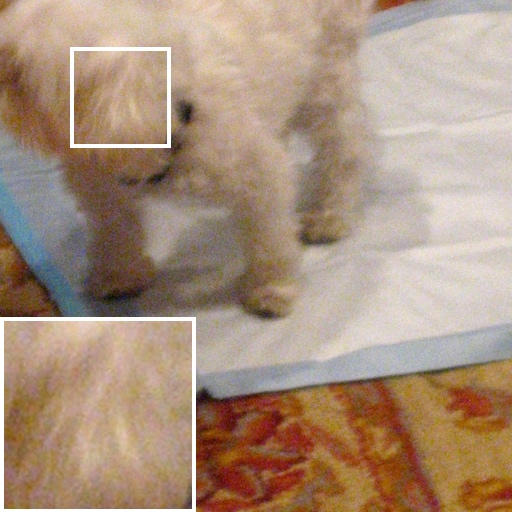} &
    \includegraphics[width=0.16\linewidth]{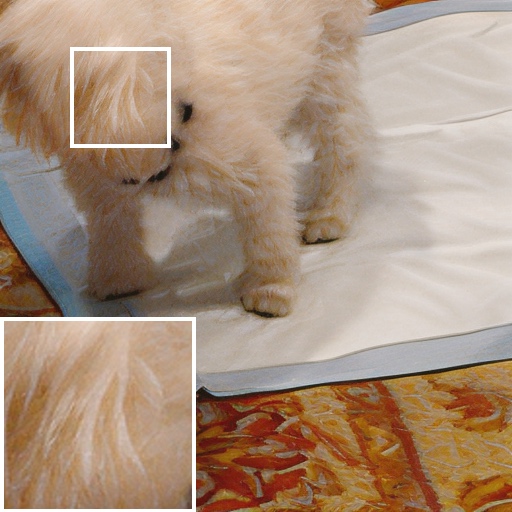} &
    \includegraphics[width=0.16\linewidth]{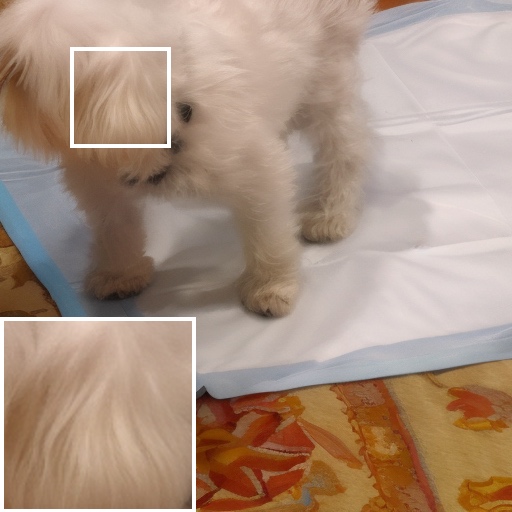} &
    \includegraphics[width=0.16\linewidth]{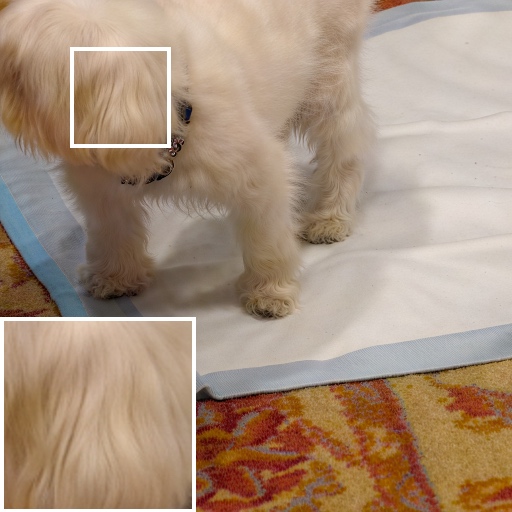} &
    \includegraphics[width=0.16\linewidth]{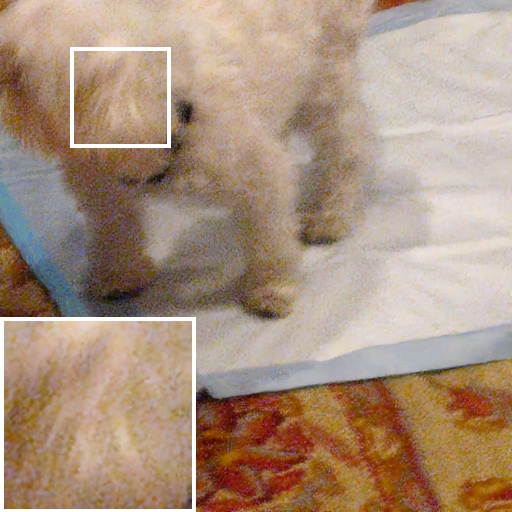} &
    \includegraphics[width=0.16\linewidth]{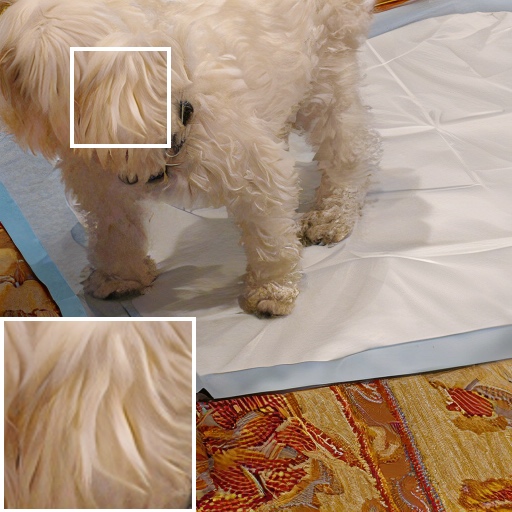} \\

    \includegraphics[width=0.16\linewidth]{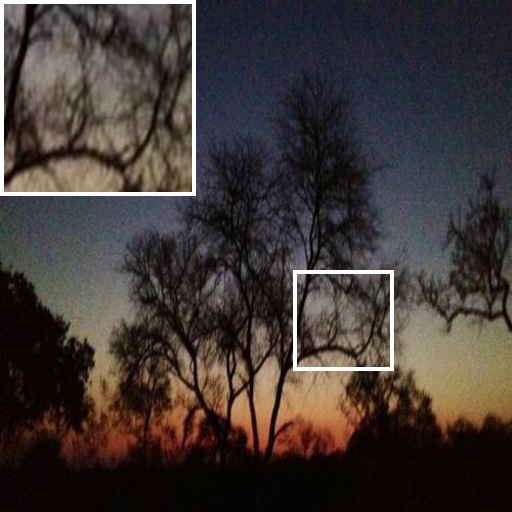} &
    \includegraphics[width=0.16\linewidth]{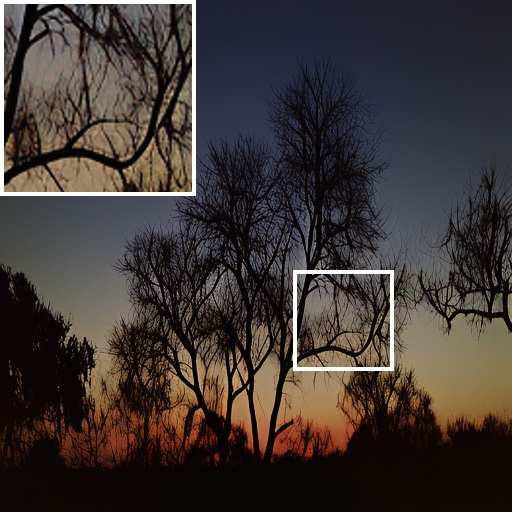} &
    \includegraphics[width=0.16\linewidth]{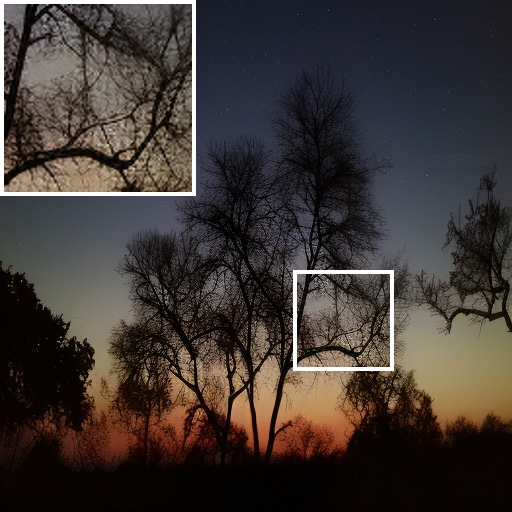} &
    \includegraphics[width=0.16\linewidth]{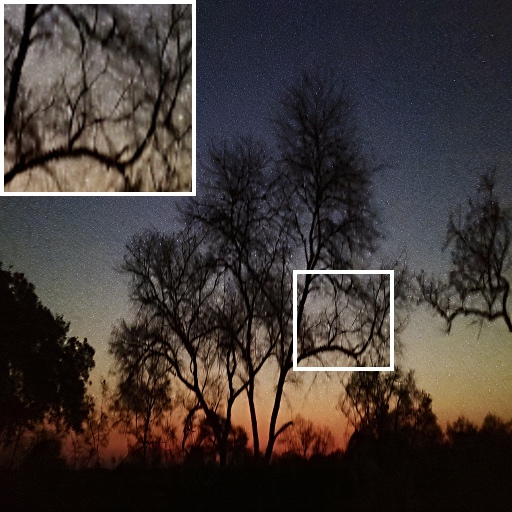} &
    \includegraphics[width=0.16\linewidth]{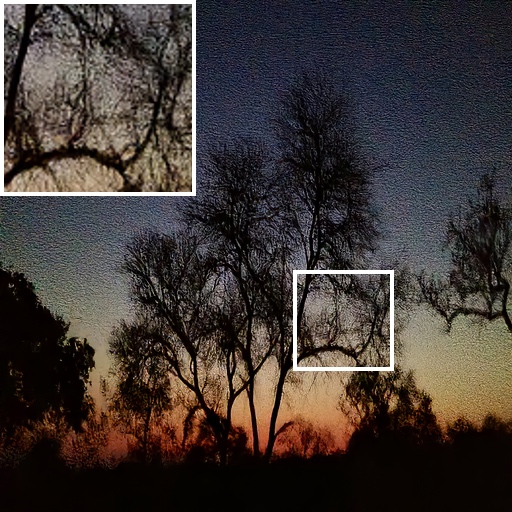} &
    \includegraphics[width=0.16\linewidth]{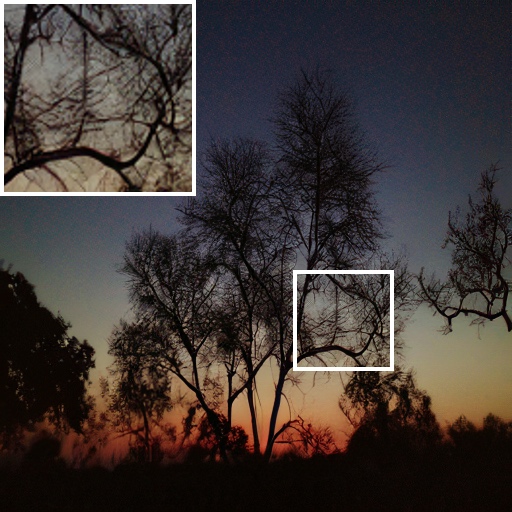} \\

    \includegraphics[width=0.16\linewidth]{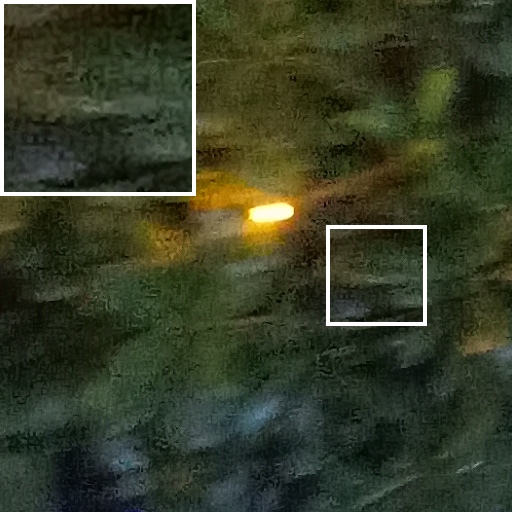} &
    \includegraphics[width=0.16\linewidth]{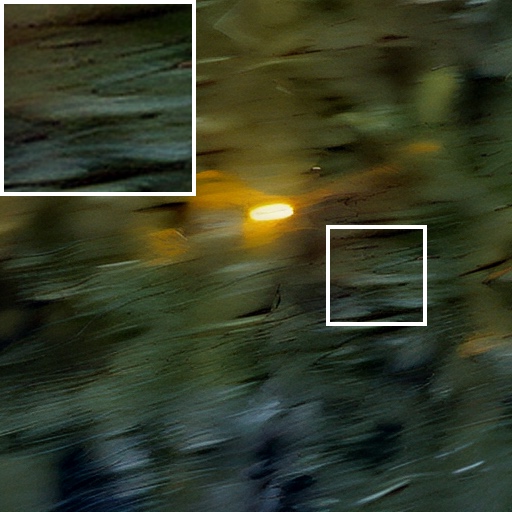} &
    \includegraphics[width=0.16\linewidth]{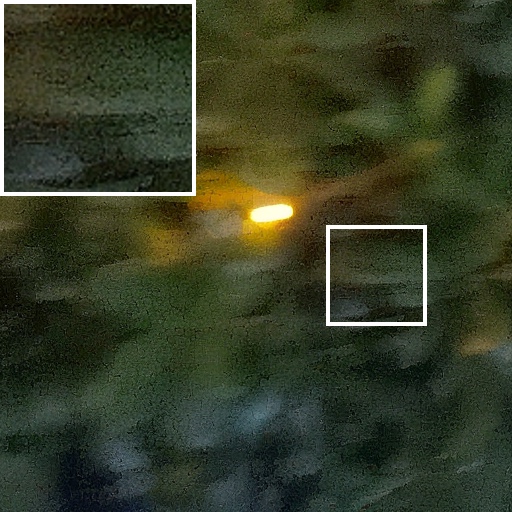} &
    \includegraphics[width=0.16\linewidth]{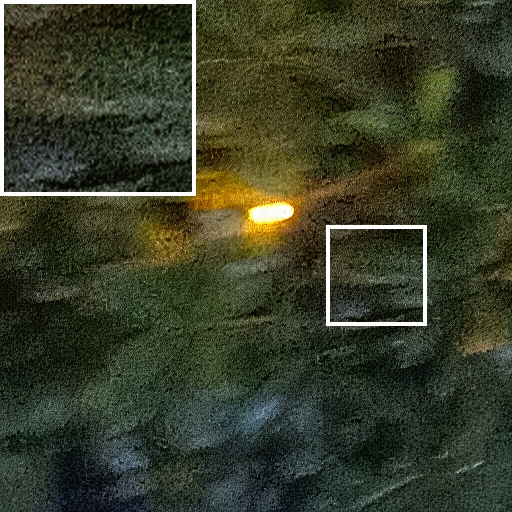} &
    \includegraphics[width=0.16\linewidth]{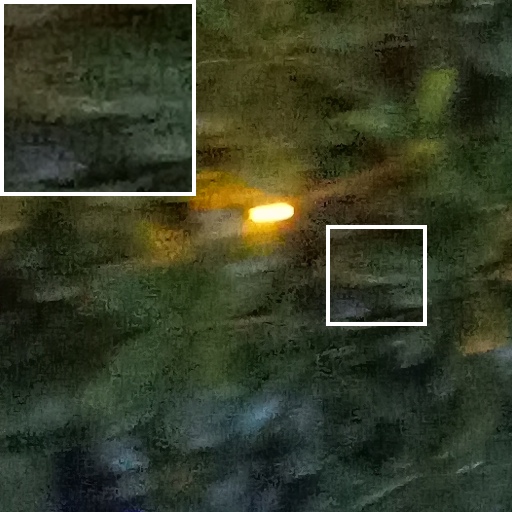} &
    \includegraphics[width=0.16\linewidth]{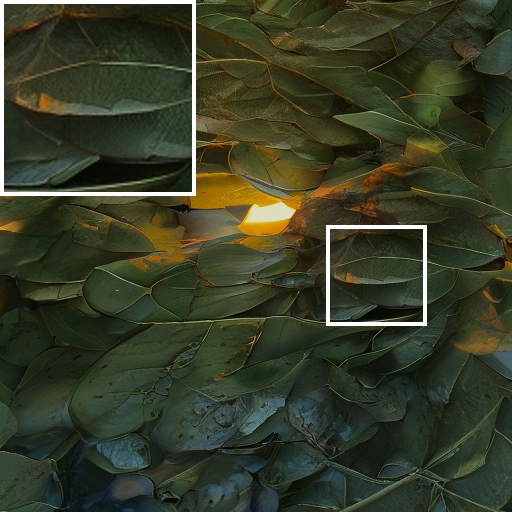} \\

    \end{tabular}
    \caption{More visualizations about real-world image restoration on the DiversePhotos$\times 1$ dataset.}
    \label{fig:evenmoredpx1}
\end{figure*}

\begin{figure*}[t]
\setlength{\tabcolsep}{1pt}
\renewcommand{\arraystretch}{0.2} % Adjust the value as needed
\centering
    \begin{tabular}{c|ccccc}
    LQ & StableSR & DiffBIR & SUPIR & DACLIP-IR & Ours\\
    \midrule
    
    \includegraphics[width=0.16\linewidth]{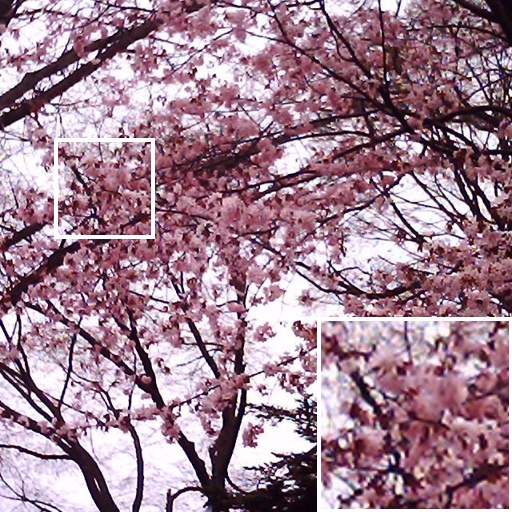} &
    \includegraphics[width=0.16\linewidth]{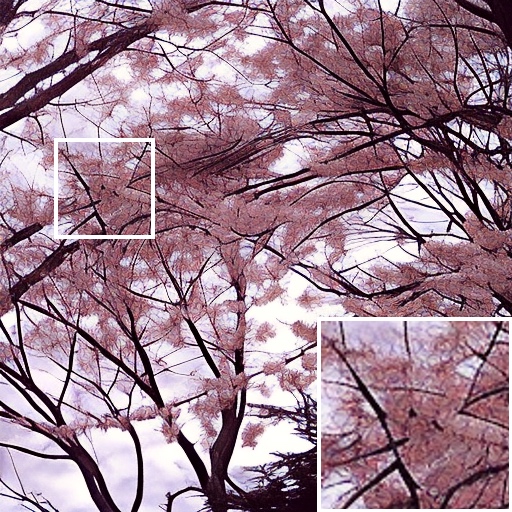} &
    \includegraphics[width=0.16\linewidth]{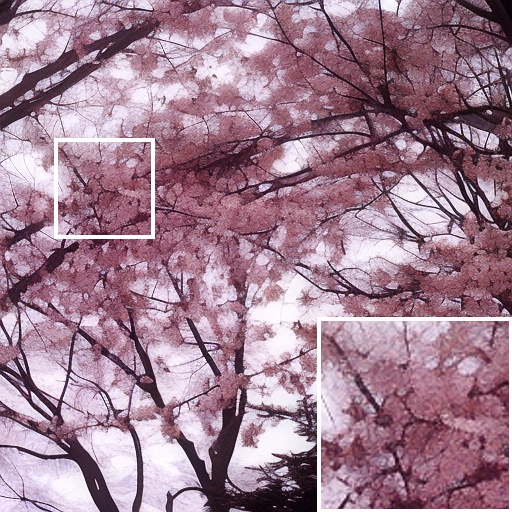} &
    \includegraphics[width=0.16\linewidth]{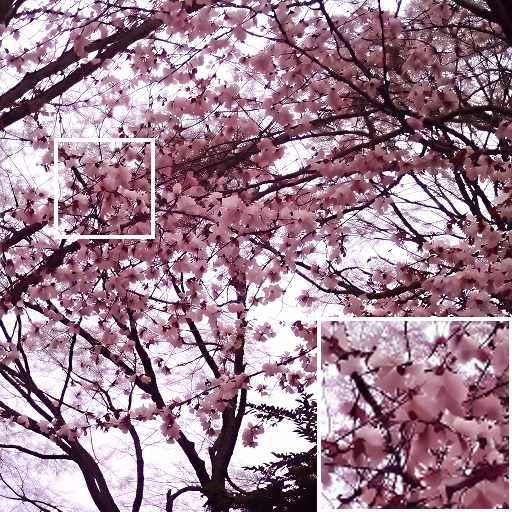} &
    \includegraphics[width=0.16\linewidth]{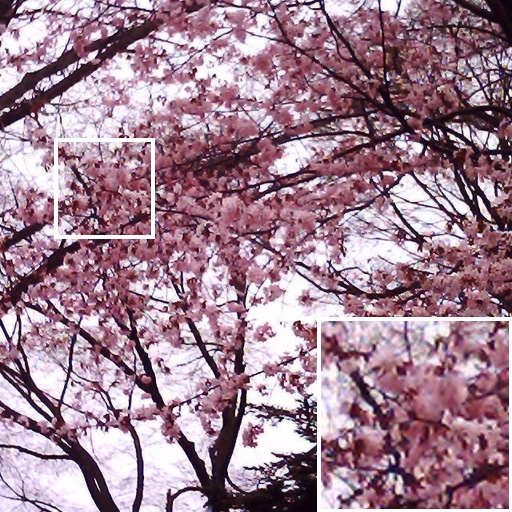} &
    \includegraphics[width=0.16\linewidth]{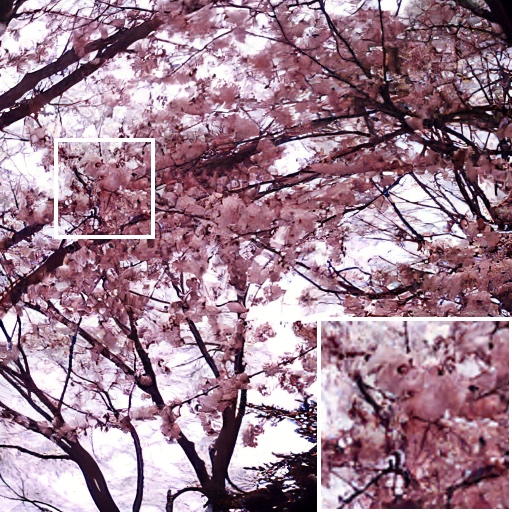} \\

    \includegraphics[width=0.16\linewidth]{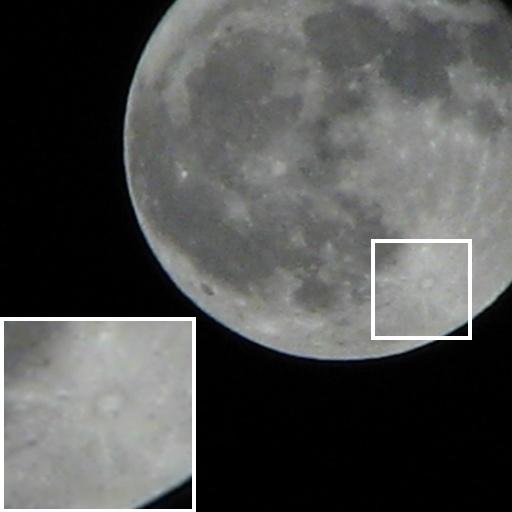} &
    \includegraphics[width=0.16\linewidth]{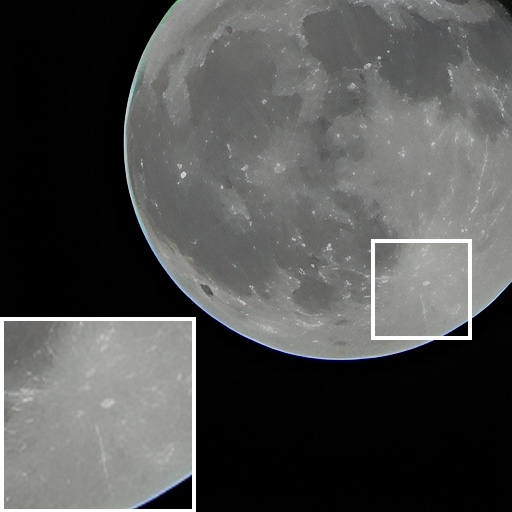} &
    \includegraphics[width=0.16\linewidth]{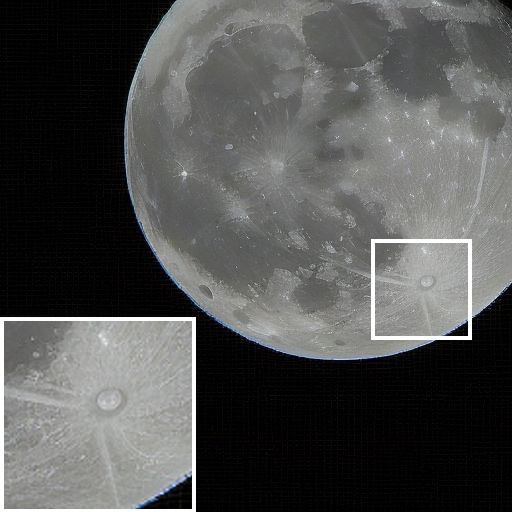} &
    \includegraphics[width=0.16\linewidth]{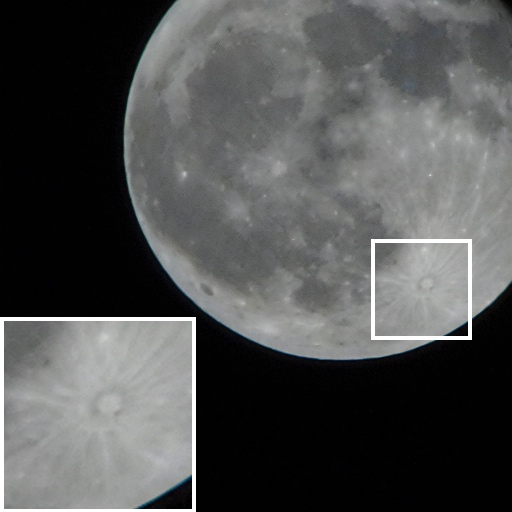} &
    \includegraphics[width=0.16\linewidth]{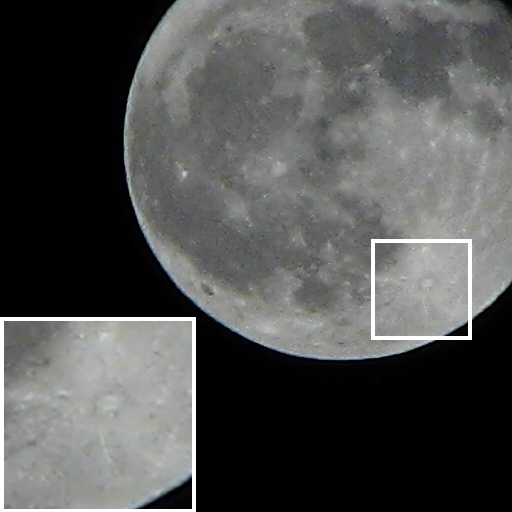} &
    \includegraphics[width=0.16\linewidth]{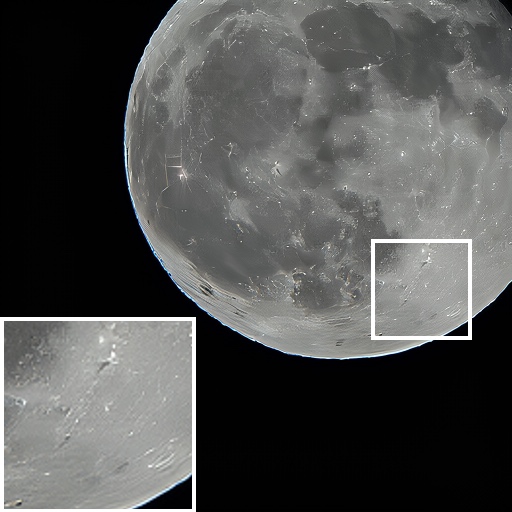} \\

    \includegraphics[width=0.16\linewidth]{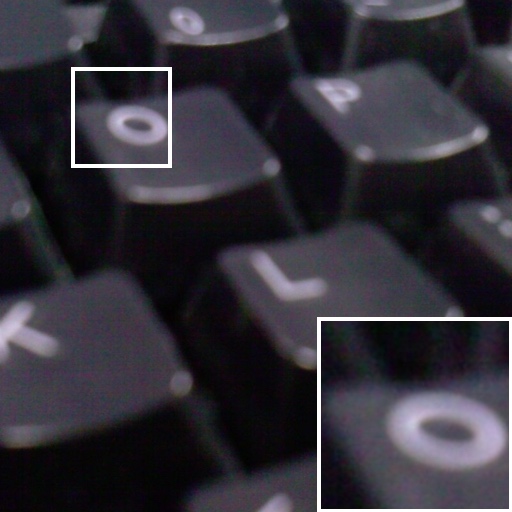} &
    \includegraphics[width=0.16\linewidth]{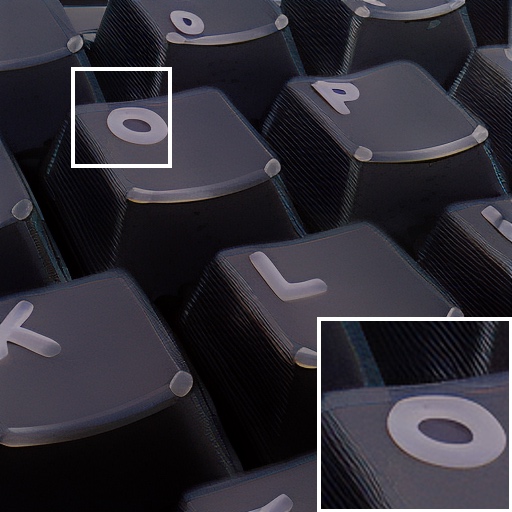} &
    \includegraphics[width=0.16\linewidth]{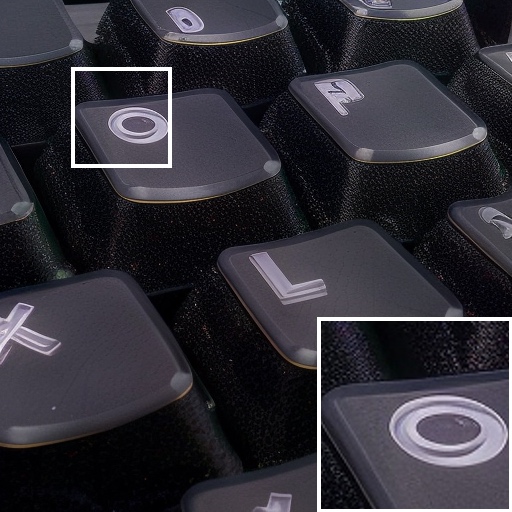} &
    \includegraphics[width=0.16\linewidth]{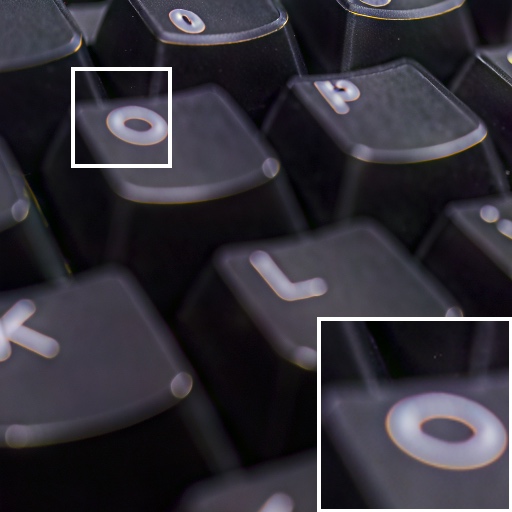} &
    \includegraphics[width=0.16\linewidth]{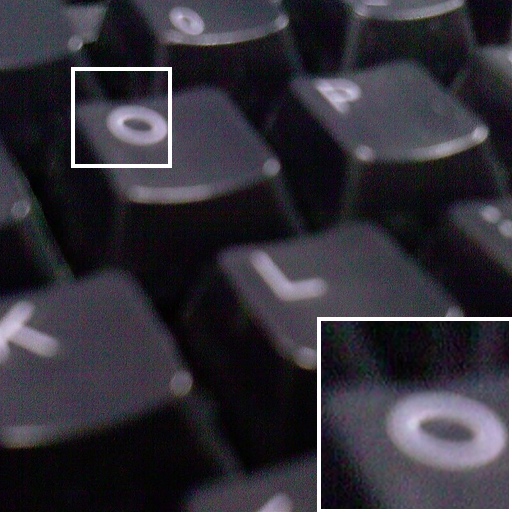} &
    \includegraphics[width=0.16\linewidth]{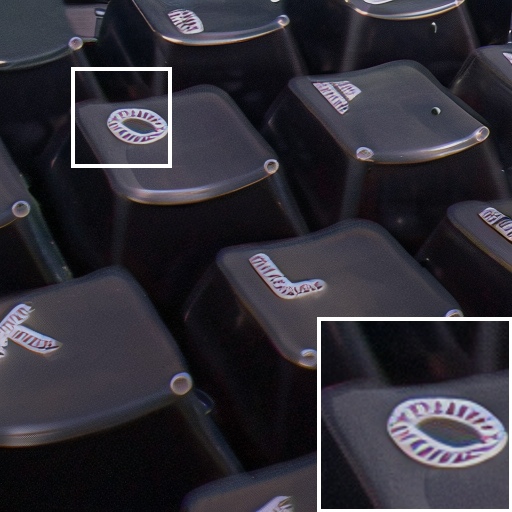} \\

    \includegraphics[width=0.16\linewidth]{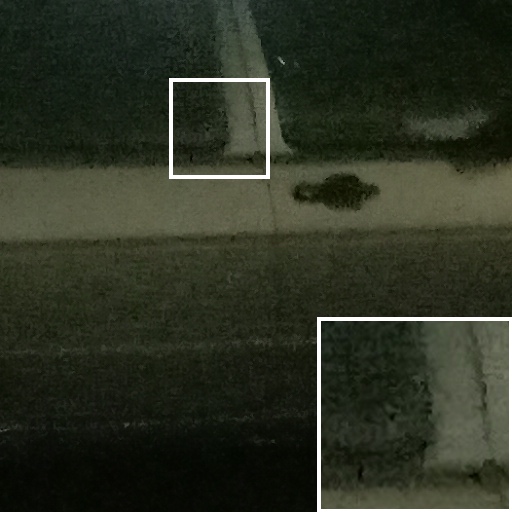} &
    \includegraphics[width=0.16\linewidth]{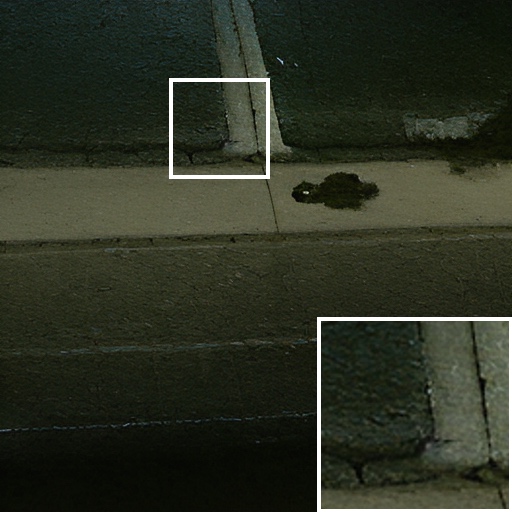} &
    \includegraphics[width=0.16\linewidth]{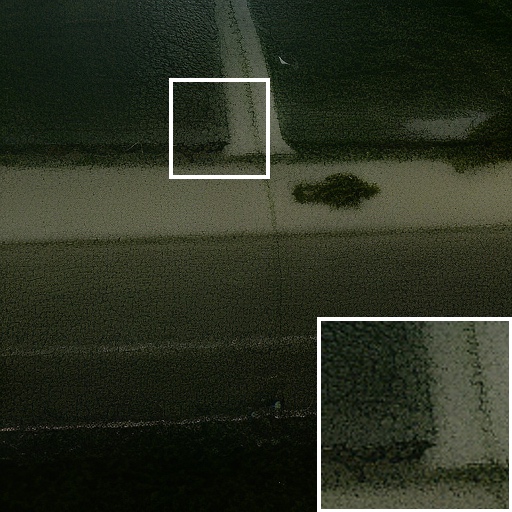} &
    \includegraphics[width=0.16\linewidth]{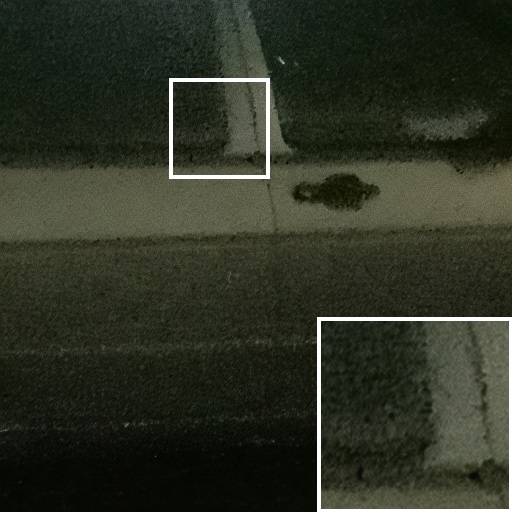} &
    \includegraphics[width=0.16\linewidth]{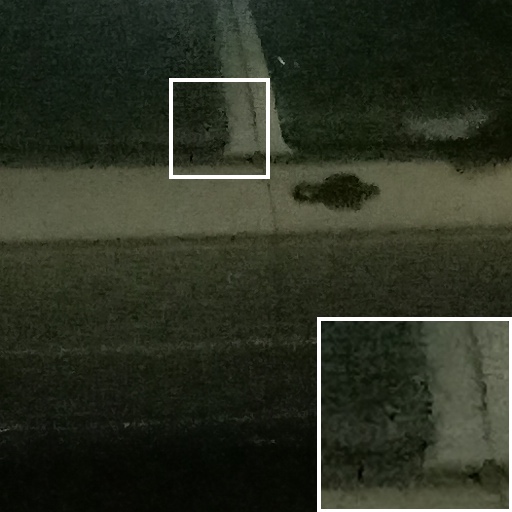} &
    \includegraphics[width=0.16\linewidth]{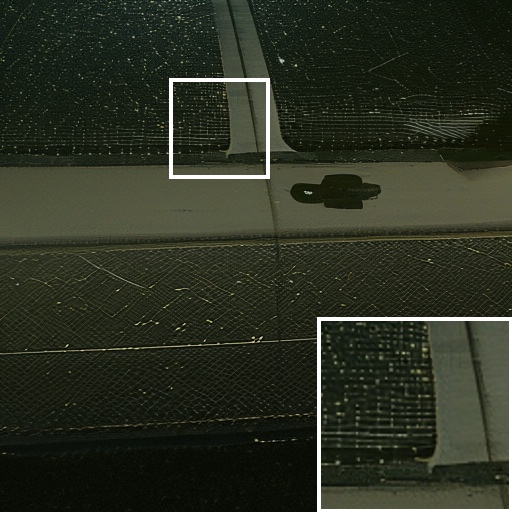} \\

    \includegraphics[width=0.16\linewidth]{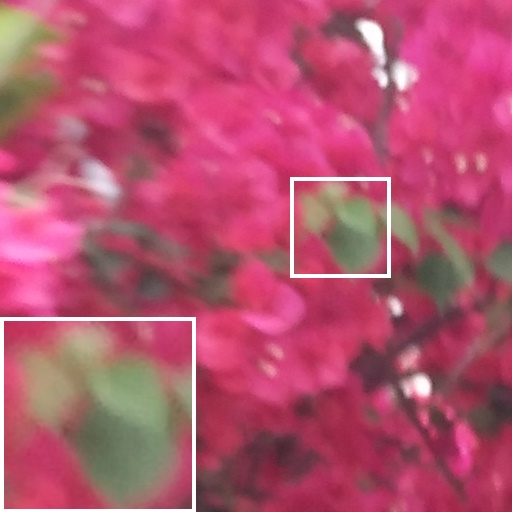} &
    \includegraphics[width=0.16\linewidth]{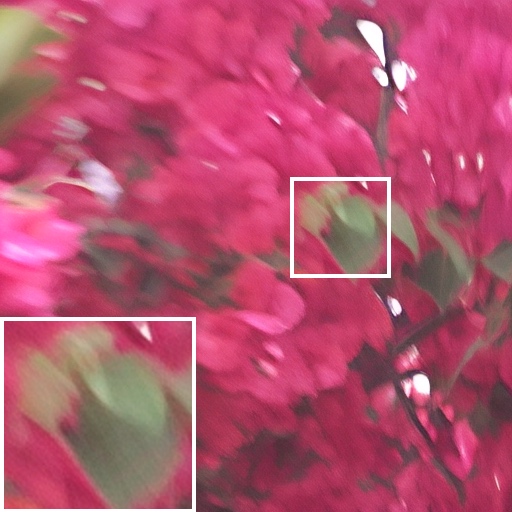} &
    \includegraphics[width=0.16\linewidth]{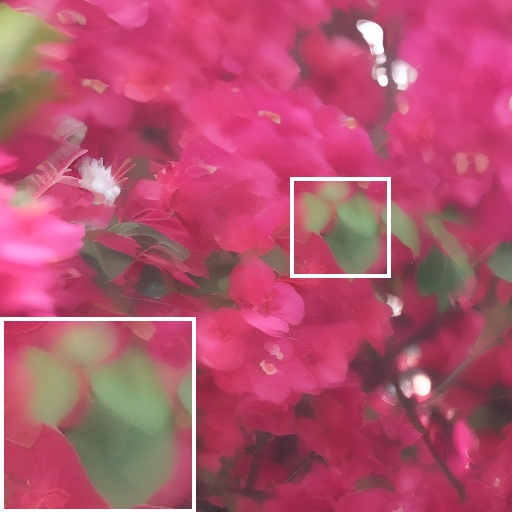} &
    \includegraphics[width=0.16\linewidth]{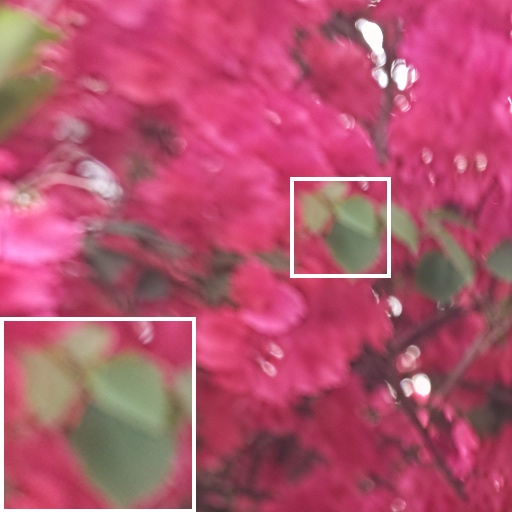} &
    \includegraphics[width=0.16\linewidth]{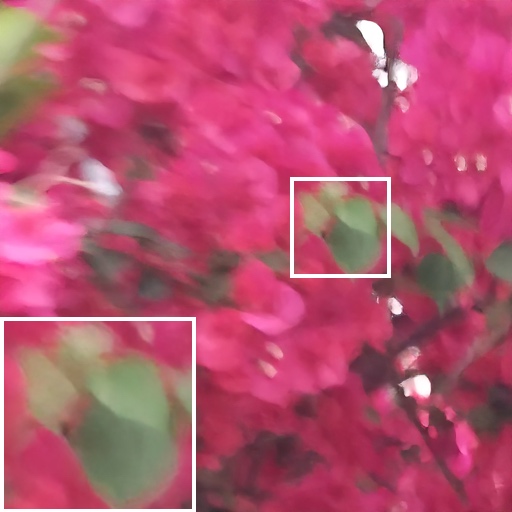} &
    \includegraphics[width=0.16\linewidth]{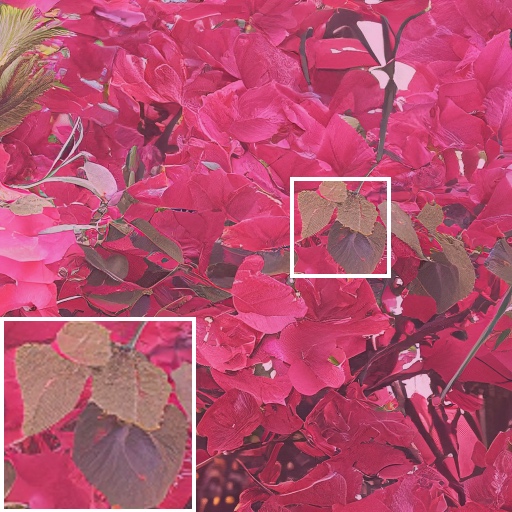} \\

    \includegraphics[width=0.16\linewidth]{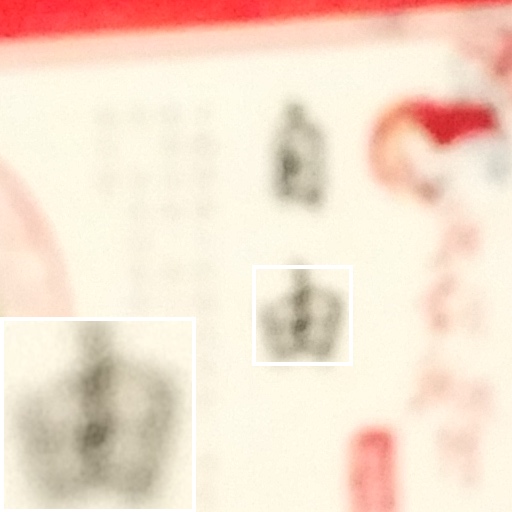} &
    \includegraphics[width=0.16\linewidth]{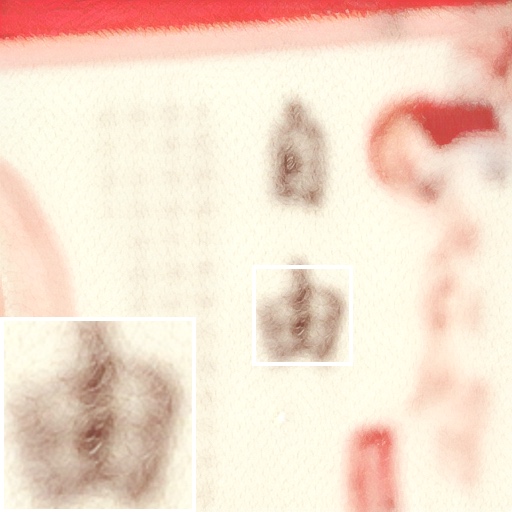} &
    \includegraphics[width=0.16\linewidth]{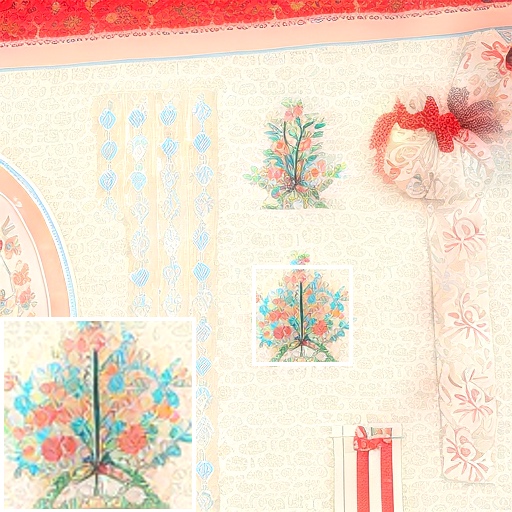} &
    \includegraphics[width=0.16\linewidth]{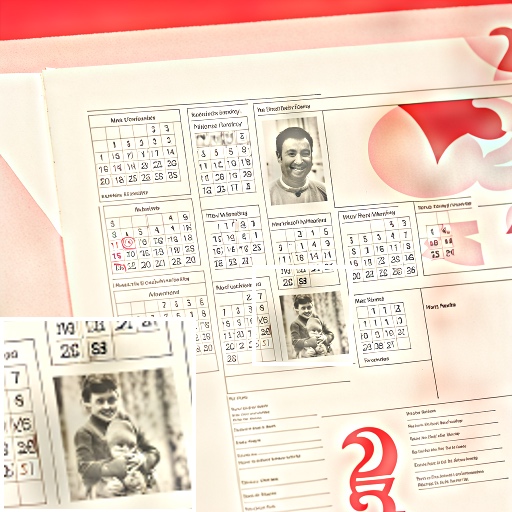} &
    \includegraphics[width=0.16\linewidth]{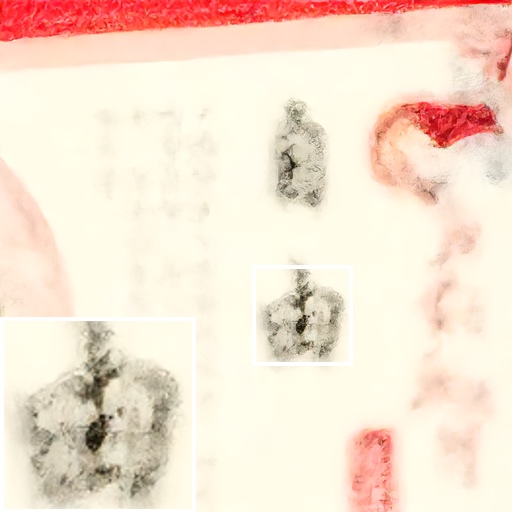} &
    \includegraphics[width=0.16\linewidth]{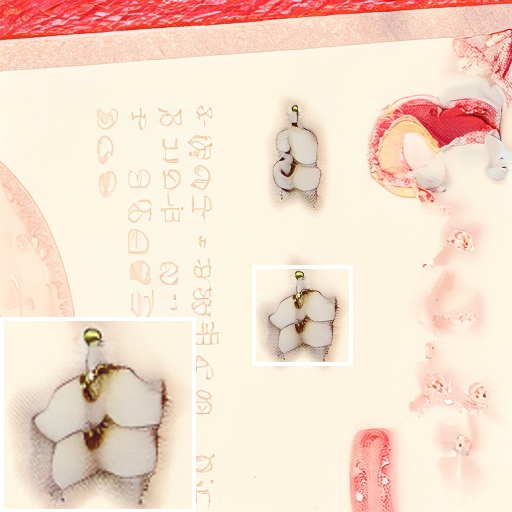} \\

    \end{tabular}
    \caption{Failure cases on the DiversePhotos$\times 1$ dataset.
    \emph{1st row}: our model does not make improvement in image details;
    \emph{2nd row}: our model (occasionally) fails to keep fidelity while improving  resolution;
    \emph{3rd row}: our model removes noise but fails to remove defocus blur;
    \emph{4th row}: our model removes noise but introduces non-existing mesh texture;
    \emph{5th row}: our model removes low resolution and motion blur, but changes the color of the leaves;
    \emph{6th row}: a hard example on which all models failed to restore.
    }
    \label{fig:failurecase}
\end{figure*}

In this section, we provide additional visualization results
on DiversePhotos$\times 1$, as shown in Fig.~\ref{fig:moredpx1},
and Fig.~\ref{fig:evenmoredpx1}. Some failure cases
are shown in Fig.~\ref{fig:failurecase}.
The failures include hallucination, color change,
artifacts, and failure to restore some degradations.
See the caption of Fig.~\ref{fig:failurecase} for details. 

\subsection{Task Weight Sensitivity}

\begin{figure*}[t]
    %\centering
    \resizebox{\linewidth}{!}{%
    \setlength{\tabcolsep}{1pt}
    \renewcommand{\arraystretch}{0.5}
    \large
    \begin{tabular}{cccc}
    
        LQ &
        SR=1.0, DN=0.0 &
        SR=0.8, DN=0.2 &
        SR=0.6, DN=0.4 \\

        \includegraphics[width=0.3\linewidth]{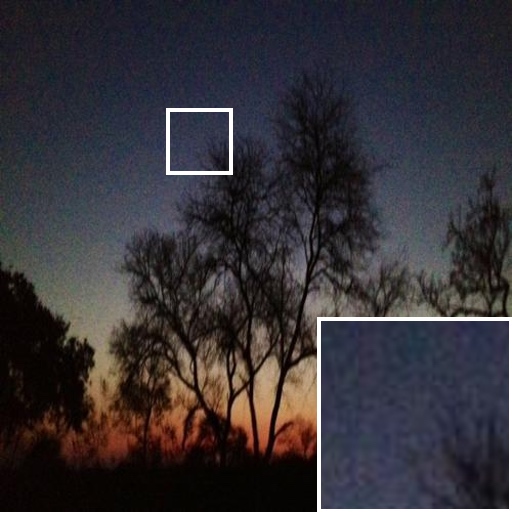} &
        \includegraphics[width=0.3\linewidth]{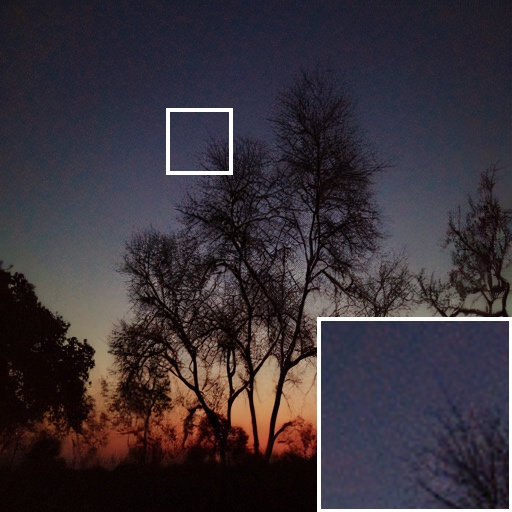} &
        \includegraphics[width=0.3\linewidth]{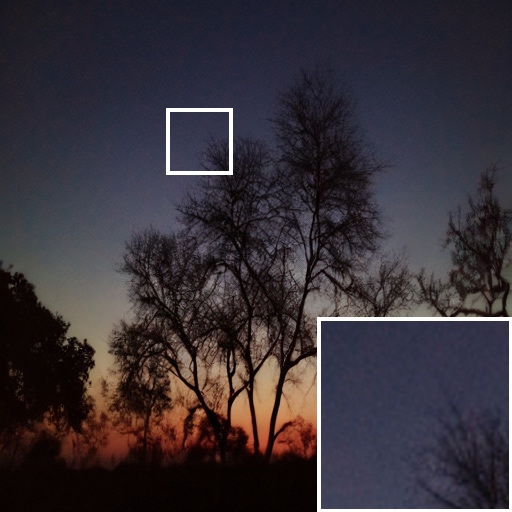} &
        \includegraphics[width=0.3\linewidth]{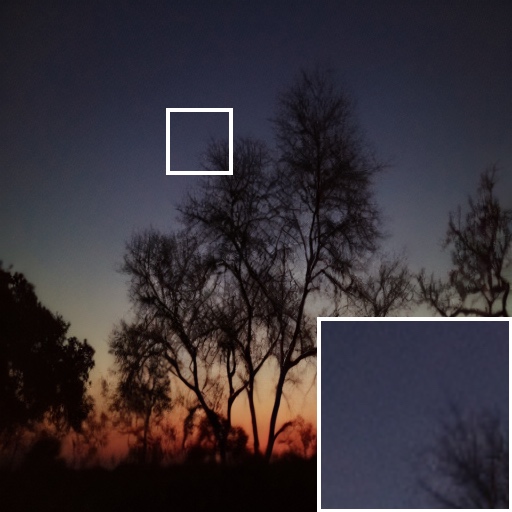} \\
        
                &
        SR=0.4, DN=0.6 &
        SR=0.2, DN=0.8 &
        SR=0.0, DN=1.0 \\
        &
        \includegraphics[width=0.3\linewidth]{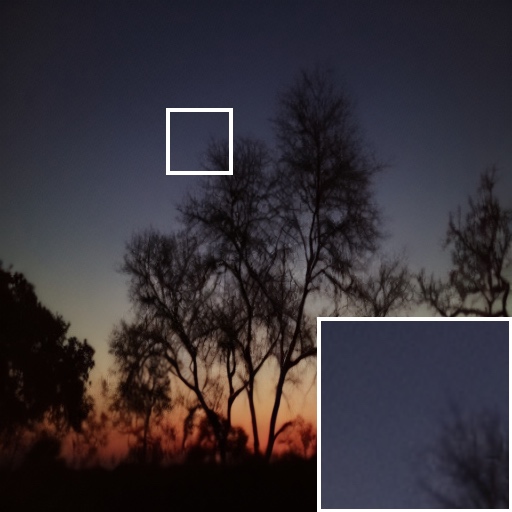} &
        \includegraphics[width=0.3\linewidth]{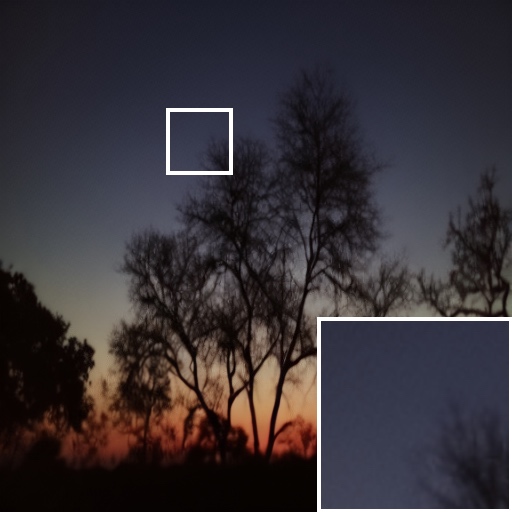} &
        \includegraphics[width=0.3\linewidth]{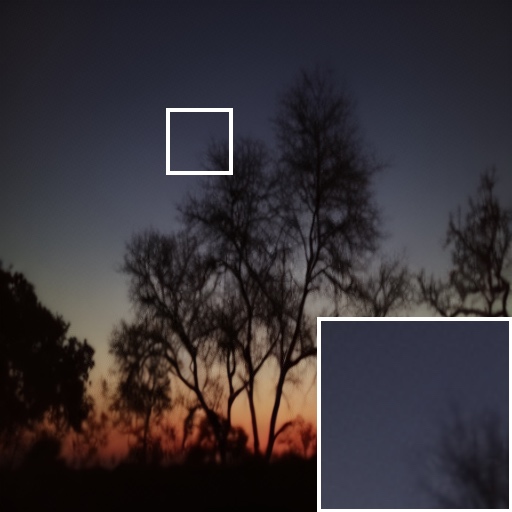} \\
        
    \end{tabular}}
    \caption{Qualitative demonstration on combination weight sensitivity. In this example, we adjust the weights for super resolution (SR) and denoise (DN), and keep the rest weights to zero. As shown from the images, the SR=1 case can improve the details
    to the tree, but does not remove all noise. In contrast, the DN=1 case can remove
    the noise, but not improve the details of the tree. By trading off the two weights,
    we can observe a smooth trade-off between the two effects. Zoom in for image details.}
    \label{fig:tradeoff2}
\end{figure*}

\begin{figure*}[t]
    %\centering
    \resizebox{\linewidth}{!}{%
    \setlength{\tabcolsep}{1pt}
    \renewcommand{\arraystretch}{0.5}
    \large
    \begin{tabular}{cccc}
    
        LQ &
        BR=1.0 &
        SR=1.0 &
        MD=1.0 \\

        \includegraphics[width=0.3\linewidth]{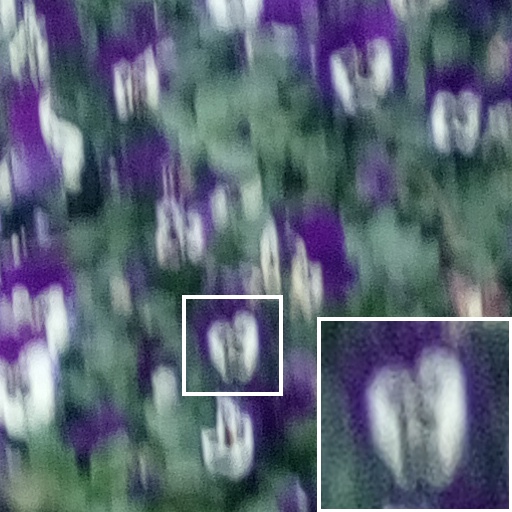} &
        \includegraphics[width=0.3\linewidth]{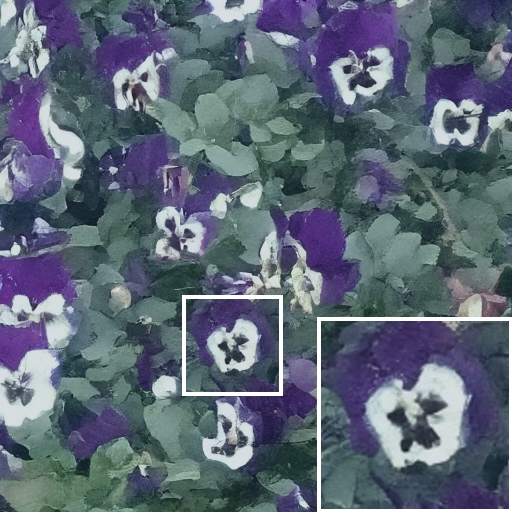} &
        \includegraphics[width=0.3\linewidth]{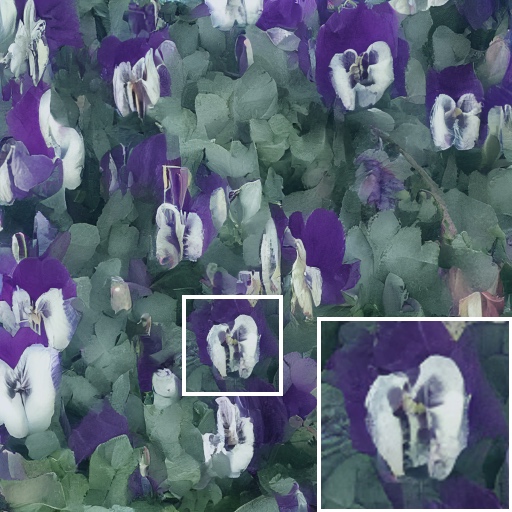} &
        \includegraphics[width=0.3\linewidth]{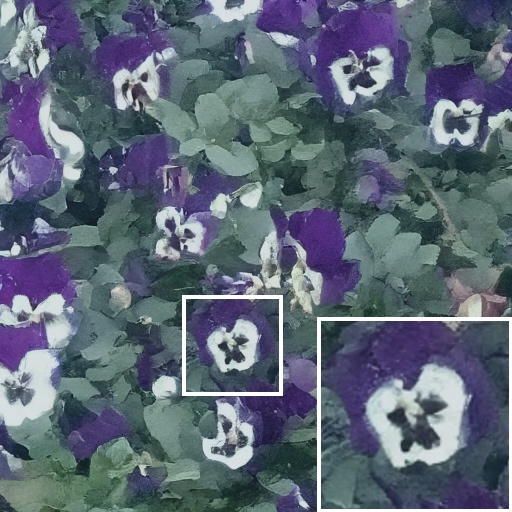} \\
        
        DD=1.0 &
        DN=1.0 &
        DownLQ=1.0 &
        Grid search\\
        \includegraphics[width=0.3\linewidth]{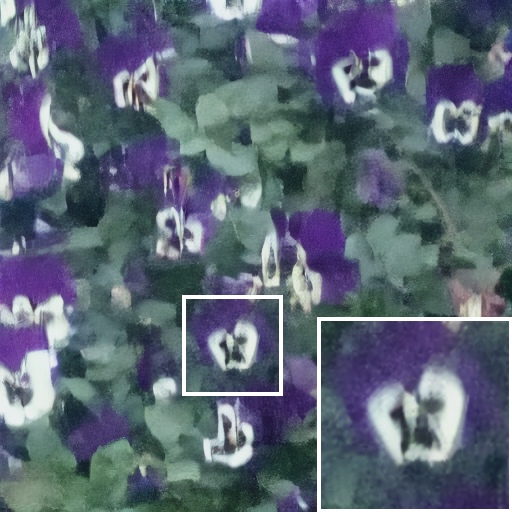} &
        \includegraphics[width=0.3\linewidth]{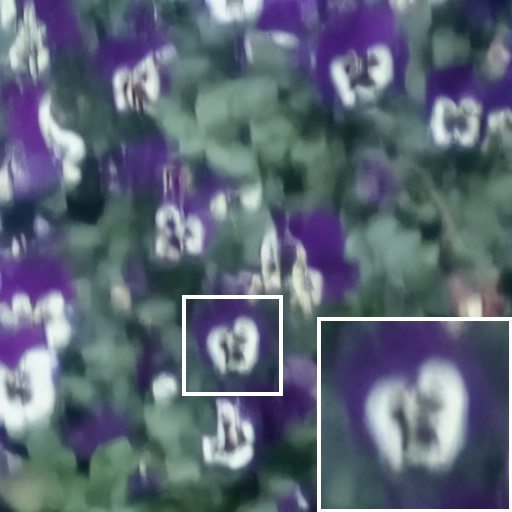} &
        \includegraphics[width=0.3\linewidth]{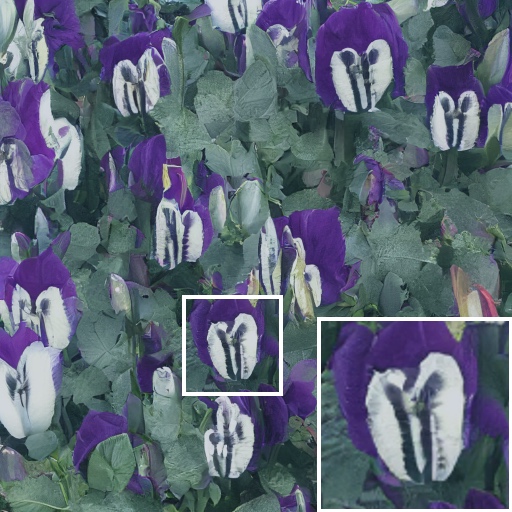} &
        \includegraphics[width=0.3\linewidth]{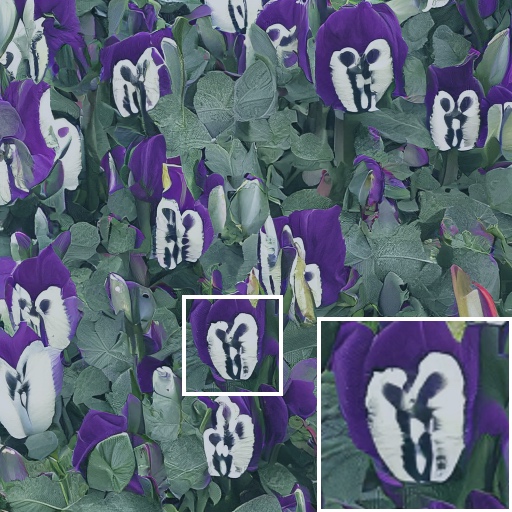} \\
        
    \end{tabular}}
    \caption{Qualitative demonstration on the same LQ input with different weights. In particular, ``SR=1.0'' means the weight for super resolution is $1.0$, while the
    rest weights are set to $0.0$. Zoom in for image details.}
    \label{fig:varweight}
\end{figure*}

\begin{figure*}[t]
    \hfill
    LQ:
    \includegraphics[width=0.25\linewidth]{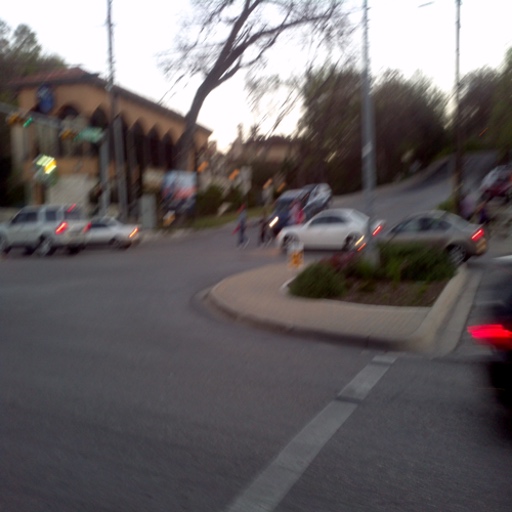}
    \hfill
    Grid search:
    \includegraphics[width=0.25\linewidth]{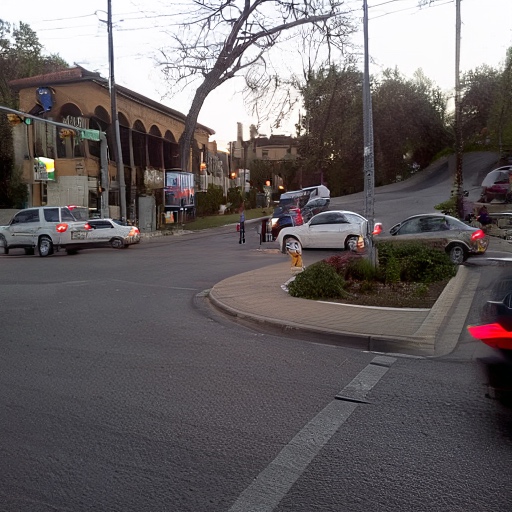}
    \hfill
    \newline
    \includegraphics[width=\linewidth]{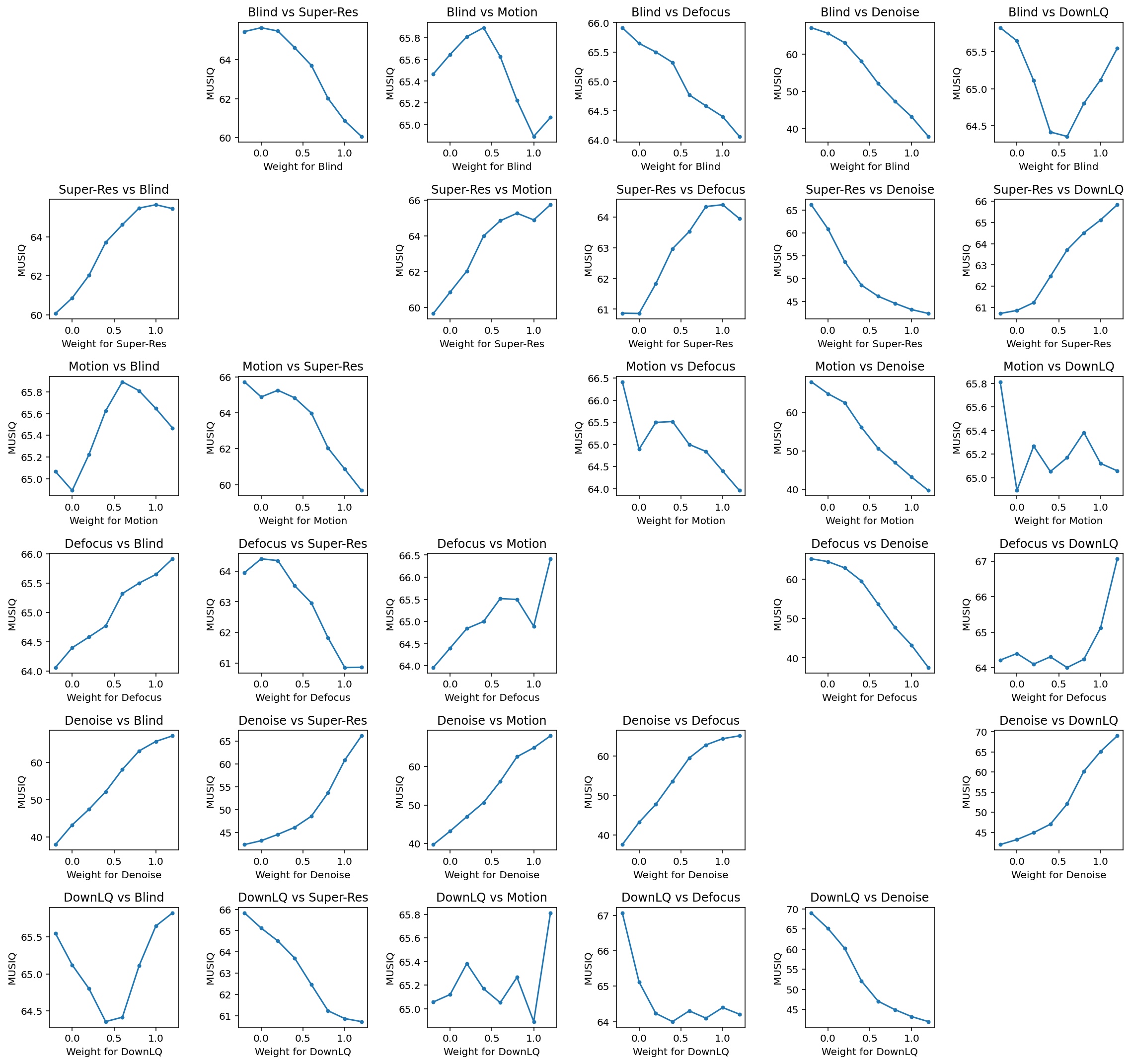}
    \caption{The trade-off curves for the example LQ image between each pair of restoration tasks: blind restoration
    (Blind), super resolution (Supre-Res), motion deblur (Motion), defocus deblur (Defocus), denoise (Denoise), and DownLQ. Here we only control the two weights for each pair
    of tasks, while keeping the rest weights as zero.
    For reference, the optimal weights (through
    grid search) for this LQ example are
    (MD=0.6,DN=-0.2,DownLQ=0.6).
    % b0.00_s0.00_m0.60_f0.00_n-0.20_h0.60
    }
    \label{fig:curve}
\end{figure*}

We provide examples to demonstrate how the changes in combination
weights in Eq.~(2) could impact the results.
Fig.~\ref{fig:tradeoff2} shows the trade-off effect between
super resolution and denoise, where a smooth trade-off between two different
effects can be observed by adjusting the combination weights.
Fig.~\ref{fig:varweight} shows different results on the same
input LQ image when applying  different combination weights.
Fig.~\ref{fig:curve} shows the MUSIQ score curves when
trading-off every pair of restoration tasks.

The average optimal weight over the DiversePhotos$\times 1$ dataset
is (BR=$0.07$, SR=$0.12$, MD=$0.07$, DD=$0.06$, DN=$-0.15$,
DownLQ=$0.83$).  % rounding error for sum to one.
%DownLQ=$0.84$).
%
The denoing task has a negative weight in average largely because the MUSIQ metric prefers
sharp images, while the denoiser (the DN=$1$ case, \emph{i.e.}, the weight for denoising is set to $1$, while the rest are set to zero) does not sharpen the given image.
So the denoiser is not preferred by MUSIQ in most cases,
and the algorithm leans towards using it as a negative
classifier-free guidance term~\cite{cfg} to push the latent
diffusion prediction to be closer to other high-quality directions.
Nevertheless, the denoiser is qualitatively effective as demonstrated
by the example in Fig.~\ref{fig:tradeoff2}.
If the average optimal weight is used for all images from DiversePhotos$\times 1$,
the results are $0.5941$, $62.10$, $0.4266$ for ClipIQA, MUSIQ, and ManIQA, respectively.
The most popular optimal weight on the dataset is
(DN=$-0.2$, DownLQ=$1.20$), where $66$ out of $160$ images 
($41.25\%$)
reach the peak MUSIQ value.
If this most popular optimal weight is used for all images,
the results are $0.6590$, $66.44$, $0.5042$ for ClipIQA, MUSIQ, and ManIQA, respectively.
%
% However, we visually observe that the DownLQ term alone cannot
% solve some defocus blur, motion blur, and fine-grained texture restoration cases.
%
%\todo{does "implicit classifier" hold in our case? how to explain}

\begin{comment}
As a proof-of-concept for demonstrating the possibility
of directly predicting the optimal combination weight (and eliminate
grid search), we analyze the correlation between some degradation-aware
features and the optimal weights.
%
In particular, the Pearson correlation and the partial distance
correlation~\cite{pdc} (ECCV2022 Best Paper) between the optimal weight
and MT-A feature~\cite{spaq} are \todo{correlation numbers};
%
The correlation values between the optimal weight
and DACLIP embedding~\cite{daclip} are \todo{correlation numbers}.
%
For reference, those correlation values between the optimal weight and ResNet-50~\cite{resnet} (ImageNet-pretrained) are \todo{correlation numbers};
and those values between the optimal weight and Gaussian noise are
\todo{correlation numbers}.
%
The correlation between the optimal weight and degradation-aware features
suggest the possibility of directly predicting the optimal weights from
them and skipping the grid search.
\end{comment}

\subsection{Evaluation on Well-Isolated Degradations}

The focus of this paper is real-world complex degradations, instead of well-isolated
degradations.
The quantitative evaluation for those well-isolated tasks, such as super-resolution, motion deblur, defocus deblur, and denoise are carried out for sanity testing purpose.
We evaluate our model on the validation sets of DIV2K~\cite{div2k}\footnote{\href{https://huggingface.co/datasets/Iceclear/StableSR-TestSets}{huggingface.co/datasets/Iceclear/StableSR-TestSets}}, GoPro~\cite{gopro}, DPDD~\cite{dpdd}, and SIDD~\cite{sidd}.
The quantitative metrics can be found in Tab.~\ref{tab:indomain}.

% Preview source code for paragraph 6

\begin{table}
\resizebox{\columnwidth}{!}{
\setlength{\tabcolsep}{2pt}

\begin{tabular}{ccccccccc}
\toprule 
\textbf{Model} & \textbf{Task Weights} & \textbf{PSNR} & \textbf{SSIM} & \textbf{LPIPS$\downarrow$} & \textbf{FID$\downarrow$} & \textbf{ClipIQA} & \textbf{MUSIQ} & \textbf{ManIQA}\tabularnewline
\midrule
\rowcolor{gblue!10}\multicolumn{9}{c}{DIV2K (3000 Crops from StableSR, size $512\times 512$)}\tabularnewline
\midrule
StableSR & - & {21.94} & {0.5343} &{0.3113} & {24.44} & 0.6771 & 65.92 & 0.4201\tabularnewline
DiffBIR & - & 21.82 & 0.5050 & 0.3670 & 32.72 & {0.7300} & {69.87} & {0.5667}\tabularnewline
SUPIR & - & 20.85 & 0.4945 & 0.3904 & 31.60 & 0.7134 & 63.69 & 0.5477\tabularnewline
DACLIP-IR & - & 21.93 & 0.4864 & 0.4881 & 71.93 & 0.3295 & 48.68 & 0.2654\tabularnewline
\hline
SR-Only & - & 22.12 & 0.5352 & 0.3028 & 21.17 & 0.6083 & 66.45 & 0.4366 \tabularnewline
% 21.62	0.5136	0.3386	27.21	0.6291	67.68	0.4551
% UniRes & SR=1 & 21.62 & 0.5136 & 0.3386 & 27.21 & 0.6291 & 67.68 & 0.4551\tabularnewline -- unires webli
UniRes & SR=1 & 21.41 & 0.5127 & 0.3325 & 25.99 & 0.6308 & 67.29 & 0.4567\tabularnewline
\midrule 
\rowcolor{ggreen!10}\multicolumn{9}{c}{GoPro (1111 Images)}\tabularnewline
\midrule
UniRes & MD=1 & 25.04 & 0.7629 & 0.1604 & 11.83 & 0.3073 & 58.65 & 0.2630\tabularnewline
\midrule 
\rowcolor{gyellow!10}\multicolumn{9}{c}{DPDD ($74$ Images)}\tabularnewline
\midrule 

% FID 31.57
UniRes & DD=1 & 24.03 & 0.6980 & 0.1678 & - & 0.5088 & 63.42 & 0.4198\tabularnewline
\midrule 
\rowcolor{red!10}\multicolumn{9}{c}{SIDD ($1280$ Crops, size $256\times 256$)}\tabularnewline
\midrule 

% FID 45.54
UniRes & DN=1 & 26.94 & 0.8120 & 0.1821 & - & 0.2945 & 22.19 & 0.2603\tabularnewline
\bottomrule
\end{tabular}

}

\caption{Evaluation on Well-Isolated Restoration Tasks. In this paper, we focus on complex degradations, instead of these well-isolated degradations.}

\label{tab:indomain}
\end{table}

%\todo{run the evaluations on GoPro, DPDD, SIDD val sets}

\subsection{Positive and Negative Prompts}

Recent works~\cite{supir,diffbir} demonstrate the effectiveness of positive
and negative prompts (\emph{e.g.}, ``blur'', ``low-quality'', \emph{etc}.).
To make the model correctly understand the negative-quality concepts, \cite{supir}
explicitly introduce negative-quality images to the training samples.
Similarly, to extend our proposed method with positive and negative prompt words,
we need to modify the training data pipeline.

In particular, after sampling each training tuple with (LQ image, text prompt, HQ image), there is (1) $1\%$ probability that
the text prompt will be replaced with positive-quality words: \emph{``photorealistic, clean, high-resolution, ultra-high definition, 4k detail, 8k resolution, masterpiece, cinematic, highly detailed.''};
(2) $1\%$ probability that the text prompt will be replaced with the negative-quality words: \emph{``oil painting, cartoon, blur, dirty, messy, low quality, deformation, low resolution, over-smooth.''}, and meanwhile swap the LQ image and HQ image;
(3) $98\%$ probability that the training tuple is left intact.
This modification allows the model to properly understand the
concept of ``positive quality'' and ``negative quality'',
which is similar to the observation in \cite{supir}.

Then we validate the impact of those positive and negative words on the DiversePhotos$\times 1$ dataset.
In particular, based on the optimal weights obtained by
grid search, if we add the diffusion latent prediction
for the positive words 
$\bm{\epsilon}_\theta(\bm{z}_t, \bm{z}_\text{LQ}, \bm{s}_\text{positive})$ with weight $+1.0$, and that for the
negative words
$\bm{\epsilon}_\theta(\bm{z}_t, \bm{z}_\text{LQ}, \bm{s}_\text{negative})$ with weight $-1.0$, the results
will be $0.6748$, $69.70$, $0.5354$ for ClipIQA, MUSIQ, and
ManIQA, respectively.
Comparing to the UniRes results under the default setting (\emph{i.e.}, $0.6519$, $68.22$, $0.5021$), the positive and
negative words leads to a slight performance gain.
%
%If we further change the weights for the positive and negative words to $+4.0$ and $-4.0$, the results will be
%$0.6780$, $71.31$, $0.5748$,
%for ClipIQA, MUSIQ, and ManIQA, respectively.
%
Further increasing the absolute values for their weights
may occasionally lead to artifacts according to our observation.
Extending our proposed method
with positive words and negative words is effective.

%\begin{comment}
%For instance, the fusion of positive prediction and the negative prediction
%in \cite{supir} can be reorganized as $w_1 \bm{\epsilon}_\theta(\bm{z}_t,
%\bm{x}, \bm{s}_\text{text+pos}) + w_2 \bm{\epsilon}_\theta(
%\bm{z}_t, \bm{x}, \bm{s}_\text{neg})$, where the $\bm{s}_\text{text+pos}$ 
%and $\bm{s}_\text{neg}$ are the positive text prompt (contains both
%semantic caption and positive words about image quality) and negative prompt
%(negative words about image quality, such as ``blur''),
%respectively.
%
%Inspired by the positive and negative prompts, and the triplet loss,
%we propose to further adjust the noise direction, pulling it closer
%to the ``positive'', and pushing it away from the ``negative'':
%
%\begin{align}
%\tilde{\bm{\epsilon}}_\theta \leftarrow
%\tilde{\bm{\epsilon}}_\theta
%+ w_\text{pos} \bm{\epsilon}_\theta(\bm{z}_t,
%\bm{x}, \bm{s}_\text{pos})
%- w_\text{neg} \bm{\epsilon}_\theta(
%\bm{z}_t, \bm{x}, \bm{s}_\text{neg})
%\end{align}
%
%And the two additional terms can always be merged into the notations
%in Eq.~\ref{eq:core} while increasing $K$ by $2$.
%\end{comment}

\subsection{Limitation of Non-Reference Metrics}

\begin{figure*}[h]
\setlength{\tabcolsep}{1pt}
\renewcommand{\arraystretch}{0.2} % Adjust the value as needed
\centering
    \begin{tabular}{ccc}
    LQ & MD=1 & Grid search\\
    %\midrule
    
    \includegraphics[width=0.33\linewidth]{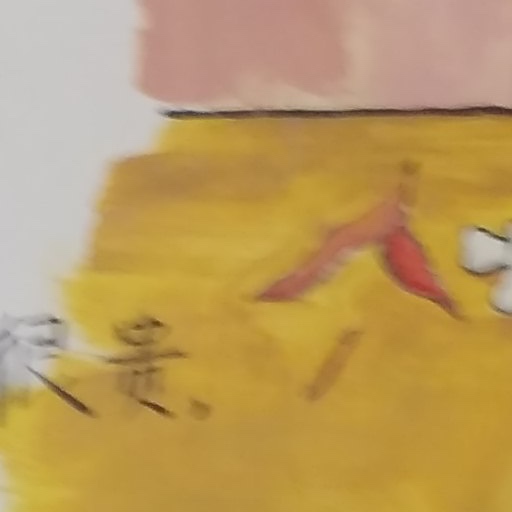} &
    \includegraphics[width=0.33\linewidth]{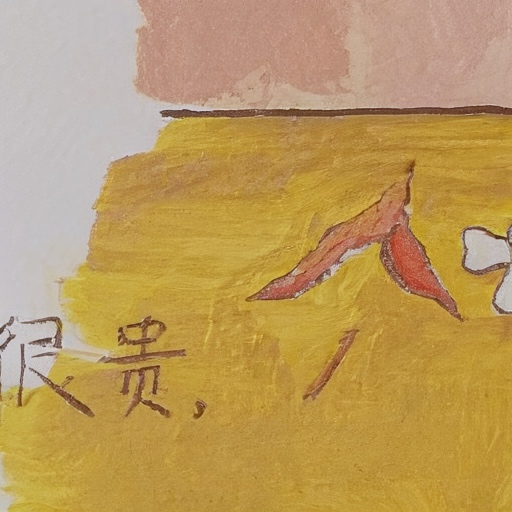} &
    \includegraphics[width=0.33\linewidth]{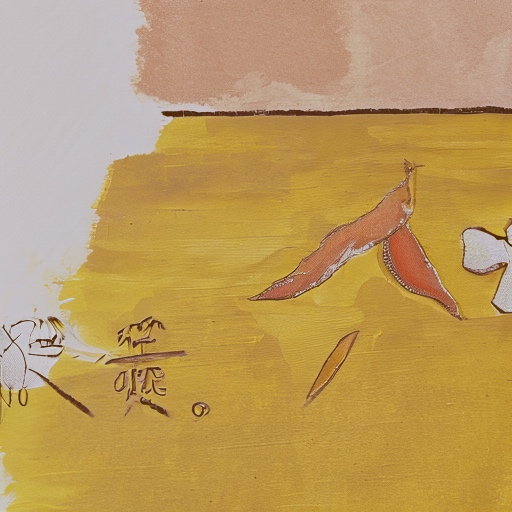}
    \\

    \end{tabular}
    \caption{Demonstration of the occasional discrepancy between human preference and
    non-reference metric.
    The grid search result (right) removes motion blur from the LQ (left), but also impacts fidelity.
    However, by manually setting the weight for motion deblur (MD) to $1$ and the rest to zero,
    a visually better result can be obtained (middle). This is an example where non-reference
    metric is not fully aligned with human preference.}
    \label{fig:nrdiscrepancy}
\end{figure*}

Our method employs MUSIQ~\cite{musiq} as an approximation to human
perceptual preference for grid search.
However, MUSIQ is not fully aligned with human, and can
lead to some discrepancies where the grid search result is not visually the best.
An example for such discrepancy is shown in Fig.~\ref{fig:nrdiscrepancy}.
Potential future work may involve incorporating better image quality metrics.

\section{Dataset Details}

\subsection{DiversePhotos}

\begin{table}[h]
\resizebox{1.0\columnwidth}{!}{%
\begin{tabular}{@{}crrrr|r@{}}
\toprule
\textbf{} & \textbf{Low Resolution} & \textbf{Motion Blur} &
\textbf{Defocus Blur} & \textbf{Noise} & \emph{sum} \\ \midrule
SPAQ~\cite{spaq}      & 20                                  & 17                                  & 6                                    & 21                                 & 64                      \\
KONIQ~\cite{koniq}     & 14                                  & 4                                   & 14                                   & 8                                  & 40                      \\
LIVE~\cite{live}      & 6                                   & 19                                  & 20                                   & 11                                 & 56                      \\
\hline
\emph{sum}     & 40                                  & 40                                  & 40                                   & 40                                 & 160                     \\ \bottomrule
\end{tabular}}
\caption{DiversePhotos$\times 1$ Dataset Statistics. It contains $160$ images in total,
dedicating $40$ images for each of the dominating degradation types: low resolution, motion blur, defocus blur, and noise. The table shows the number of images we curated from each public dataset for each degradation.}
\label{tab:dpx1stats}
\end{table}

The ``DiversePhotos'' dataset is our curation of test images, curated from
SPAQ~\cite{spaq}, KONIQ~\cite{koniq}, and LIVE~\cite{live}.
The images in DiversePhotos collectively cover multiple mobile devices and DLSR cameras, as well as
a wide range of degradations.

\textbf{DiversePhotos$\times 1$}.
The DiversePhotos$\times 1$ image set involves $160$ images,
with $40$ images for each dominating degradation: low-resolution, motion blur,
defocus blur, and noise.
Each image is in $512\times 512$ resolution. See Tab.~\ref{tab:dpx1stats} for the statistics.

\textbf{DiversePhotos$\times 4$}.
This set of test images are the $128\times 128$ center crops of the DiversePhotos$\times 1$ images.

\textbf{Steps for reproducing ``DiversePhotos$\times 1$'':}
\begin{enumerate}
    \item Download SPAQ~\cite{spaq}, KONIQ~\cite{koniq}, and LIVE~\cite{live} datasets.
    \item Gather images whose file names are mentioned in the following 12 listings.
    \item Center-crop all images from SPAQ and KONIQ datasets to $512\times 512$ resolution.
    \item Resize (bicubic) all images from LIVE dataset (from $500\times 500$) to $512\times 512$ resolution.
\end{enumerate}

%\todo{double check the file names}
% koniq: file name OK
% spaq: file name ok
% live: file name OK

{\small%

\vspace{.3em}
\noindent
\hl{(SPAQ, low resolution as dominating degradation, with other degradations)}:
%SPAQ images with low resolution as dominating degradation:
00019, 00025, 00033, 00109, 00192,  00226,  00251, 00381, 00414, 00559, 00561, 00585, 00743, 03973, 04085, 04136, 04270, 04317, 04334, 06682.

\vspace{.3em}
\noindent
\hl{(SPAQ, motion blur as dominating degradation, with other degradations)}: % SPAQ images with motion blur as dominating degradation:
00043, 00075, 00121, 00161, 00175, 00178, 00236, 01868, 03513, 04089, 04272, 04380, 06341, 06863, 10388, 10391, 10495.

\vspace{.3em}
\noindent
\hl{(SPAQ, defocus blur  as dominating degradation, with other degradations)}: % SPAQ images with defocus blur as dominating degradation:
00125, 00212, 00282, 04379, 06727, 09121.

\vspace{.3em}
\noindent
\hl{(SPAQ, noise as dominating degradation, with other degradations)}: %SPAQ images with noise as dominating degradation:%
00077, 00086, 00096, 00143, 00187, 00199, 00292, 00365, 00450, 04337, 04345, 06485, 06703, 07121, 07162, 07394, 07494, 07866, 07903, 08108, 09682.

\vspace{.3em}
\noindent
\hl{(KONIQ, low resolution as dominating degradation, with other degradations)}: %KONIQ images with low resolution as dominating degradation: 
1755366250, 187640892, 2096424103, 2443117568, 2 6393826, 2704811, 2836089223, 2956548148, 3015139450, 3435545140, 3551648026, 4378419360, 527633229, 86243803.

\vspace{.3em}
\noindent
\hl{(KONIQ, motion blur as dominating degradation, with other degradations)}: %KONIQ images with motion blur as dominating degradation: 
2367261033, 3147416579, 331406867, 62480371.

\vspace{.3em}
\noindent
\hl{(KONIQ, defocus blur as dominating degradation, with other degradations)}: %KONIQ images with motion blur as dominating degradation: 
1306193020, 315889745, 55711788, 1807195948, 206294085, 2166503846, 2214729676, 23371433, 2360058082, 2950983139, 3149433848, 324339500, 427196028, 518080817.

\vspace{.3em}
\noindent
\hl{(KONIQ, noise as dominating degradation, with other degradations)}: %KONIQ images with noise as dominating degradation: 
1317678723, 1987196687, 218457399, 2593384818, 2837843986, 2867718050, 3727572481, 4410900135,

\vspace{.3em}
\noindent
\hl{(LIVE, low resolution as dominating degradation, with other degradations)}: %LIVE images with low resolution as dominating degradation:
110, 723, 760, 805, 819, 875.

\vspace{.3em}
\noindent
\hl{(LIVE, motion blur as dominating degradation, with other degradations)}: %LIVE images with motion blur as dominating degradation: 
1017, 104, 1156, 12, 154, 239, 270, 283, 29, 429, 458, 460, 468, 659, 663, 700, 732, 810, 856.

\vspace{.3em}
\noindent
\hl{(LIVE, defocus blur as dominating degradation, with other degradations)}: %LIVE images with defocus blur as dominating degradation:
337, 550, 592, 698, 713, 714, 717, 731, 737, 750, 751, 787, 788, 855, 862, 873, 874, 876, 884, 887.

\vspace{.5em}
\noindent
\hl{(LIVE, noise as dominating degradation, with other degradations)}: %LIVE images with noise as dominating degradation: 
1001, 1011, 1024, 1037, 1055, 1079, 1098, 1149, 370, 443, 5.
}

We will provide public download links to the resulting images
in the future.

\subsection{OID-Motion}

\begin{figure*}[t]
\setlength{\tabcolsep}{1pt}
\renewcommand{\arraystretch}{0.2} % Adjust the value as needed
\centering
    \begin{tabular}{cc|cc|cc}
    LQ & HQ & LQ & HQ & LQ & HQ\\
    \midrule
    
    \includegraphics[width=0.16\linewidth]{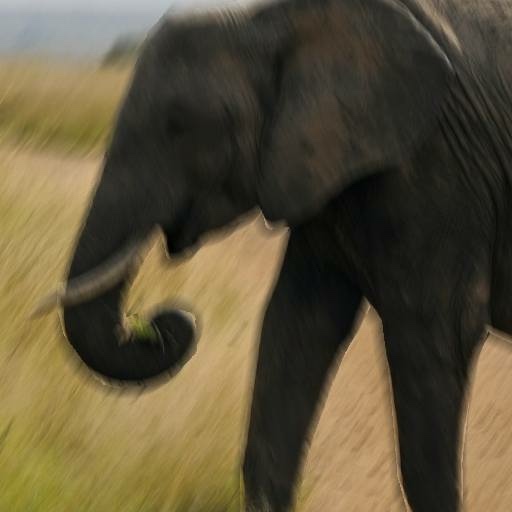} &
    \includegraphics[width=0.16\linewidth]{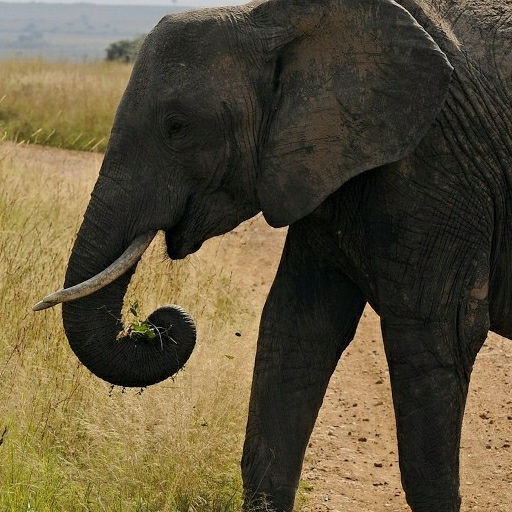} &
    \includegraphics[width=0.16\linewidth]{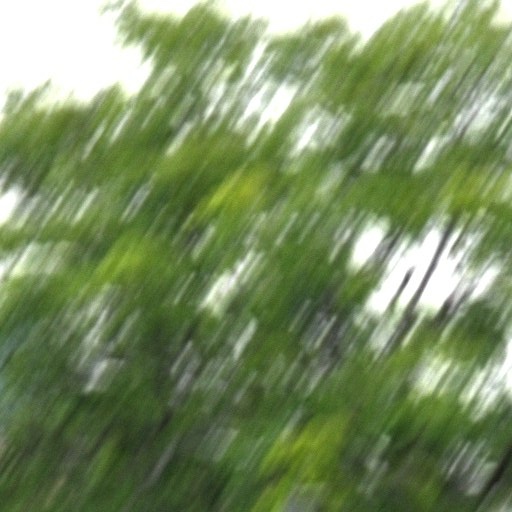} &
    \includegraphics[width=0.16\linewidth]{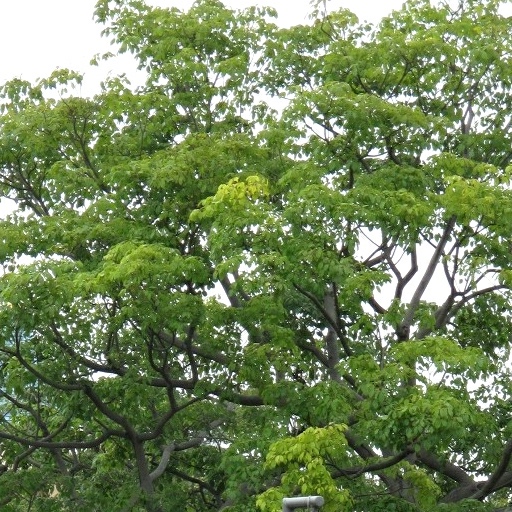} &
    \includegraphics[width=0.16\linewidth]{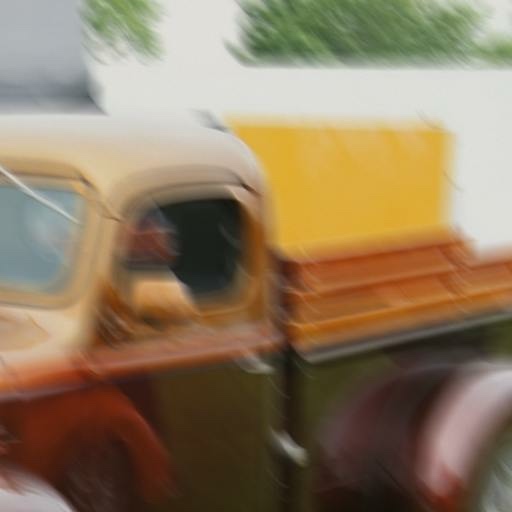} &
    \includegraphics[width=0.16\linewidth]{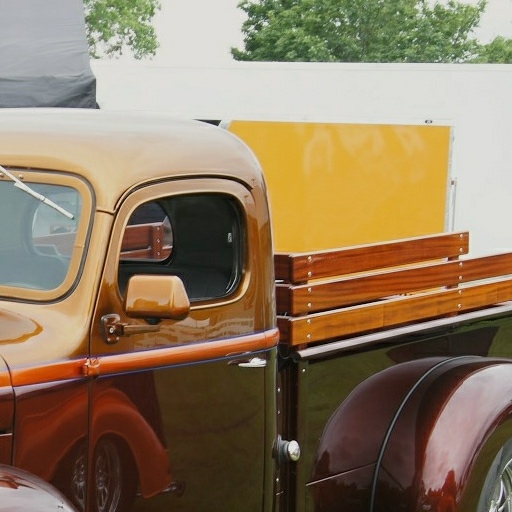} 
    \\
    
    \includegraphics[width=0.16\linewidth]{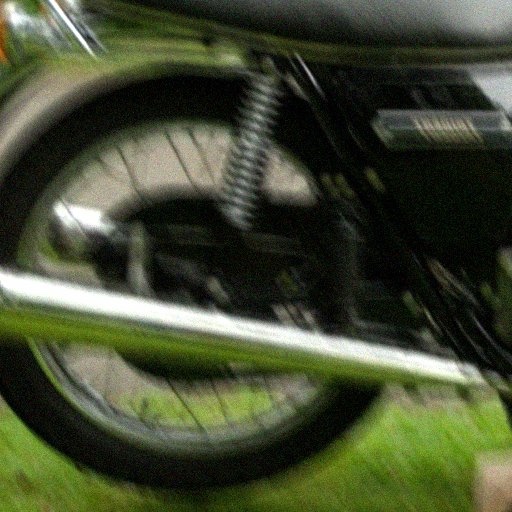} &
    \includegraphics[width=0.16\linewidth]{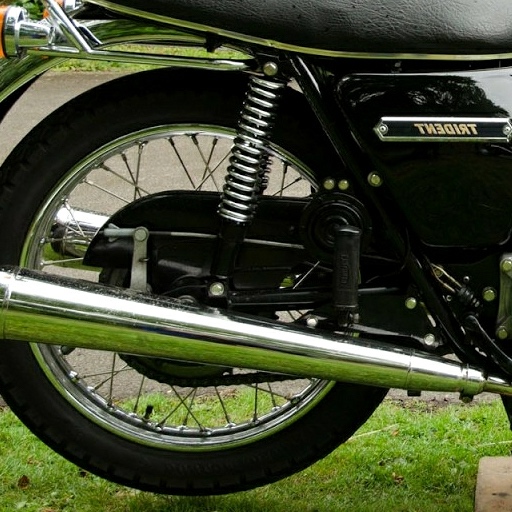} &
    \includegraphics[width=0.16\linewidth]{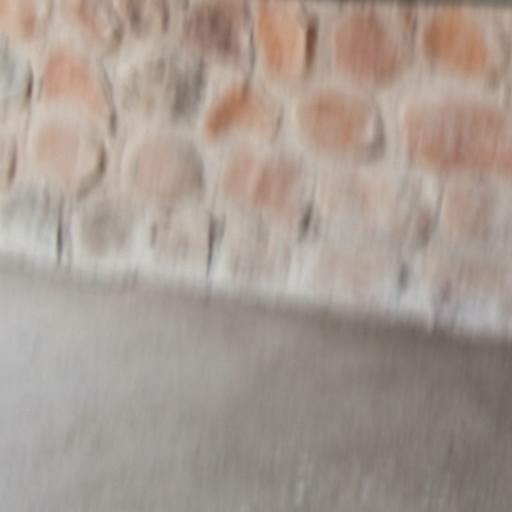} &
    \includegraphics[width=0.16\linewidth]{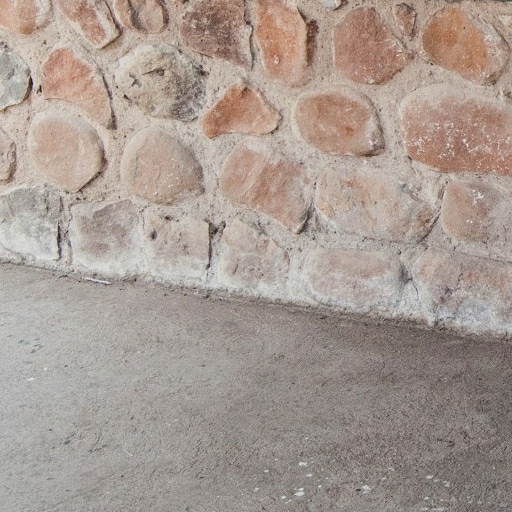} &
    \includegraphics[width=0.16\linewidth]{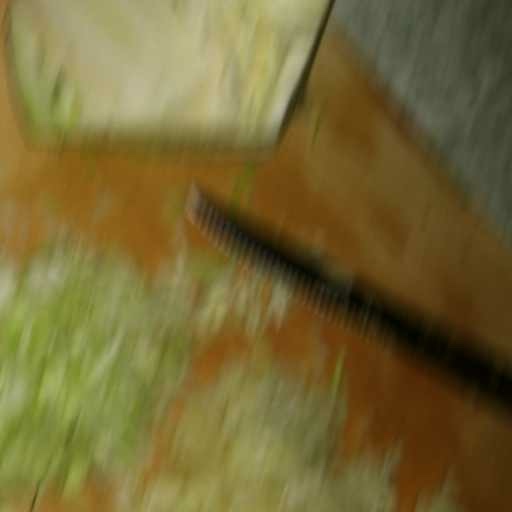} &
    \includegraphics[width=0.16\linewidth]{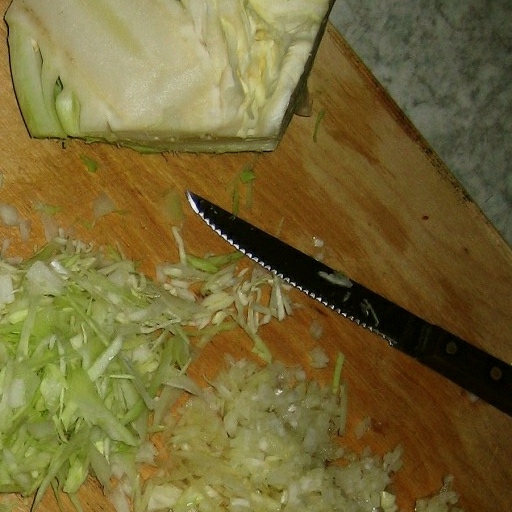} 
    \\
    
    \includegraphics[width=0.16\linewidth]{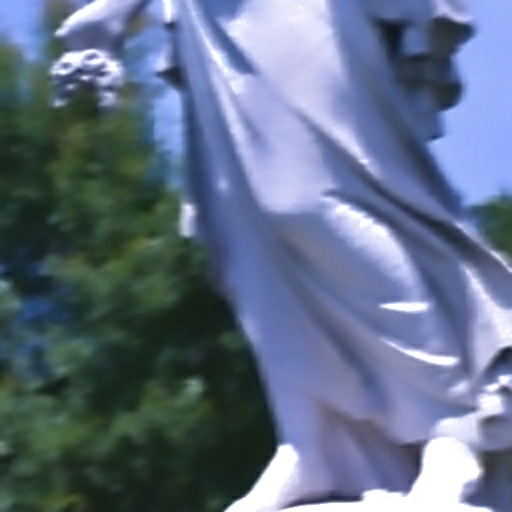} &
    \includegraphics[width=0.16\linewidth]{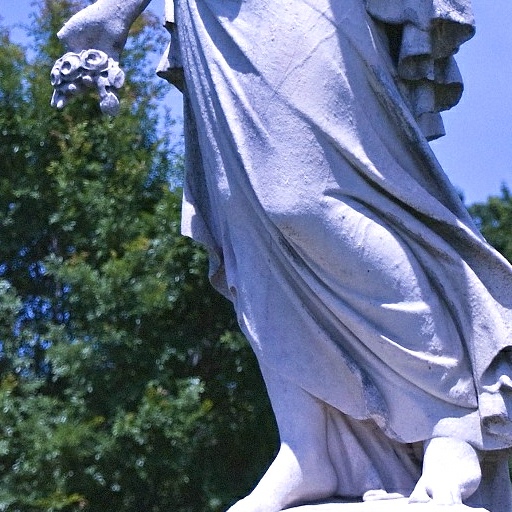} &
    \includegraphics[width=0.16\linewidth]{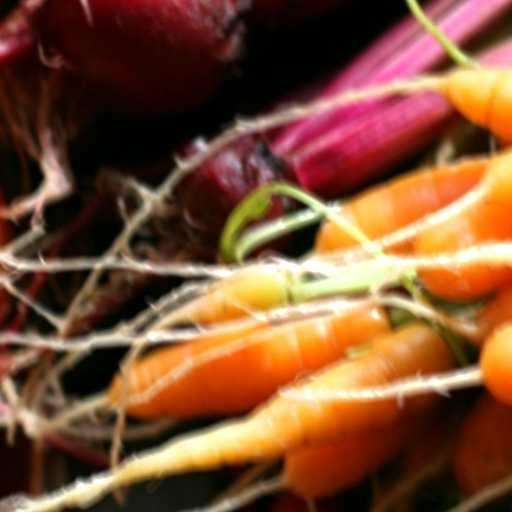} &
    \includegraphics[width=0.16\linewidth]{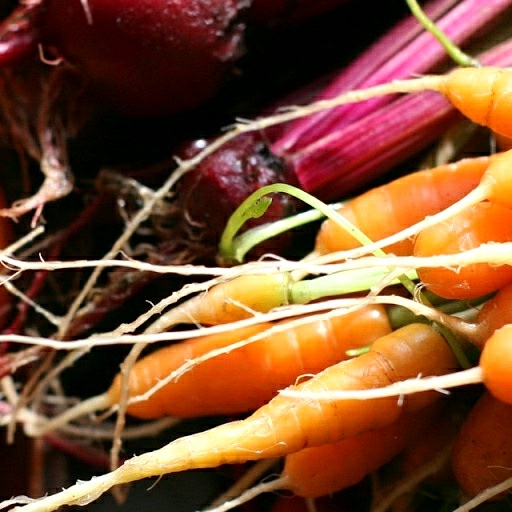} &
    \includegraphics[width=0.16\linewidth]{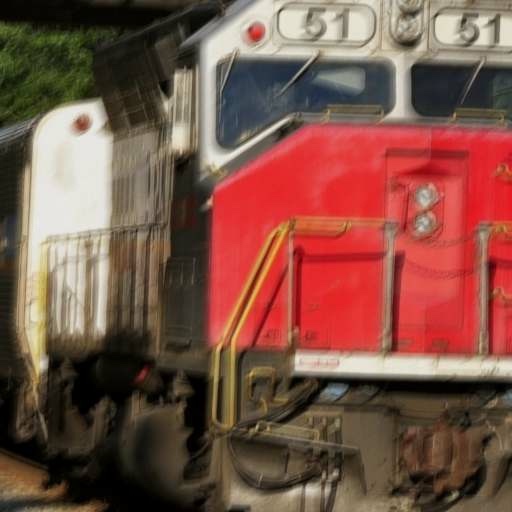} &
    \includegraphics[width=0.16\linewidth]{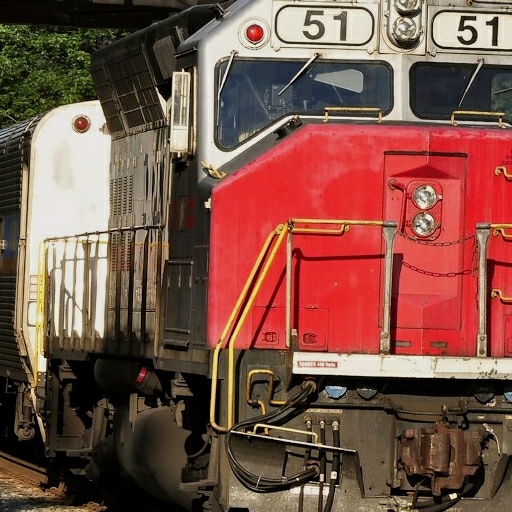} 

    \end{tabular}
    \caption{Samples from the OID-Motion training dataset. It is simulated with the camera shake blur~\cite{shakeblur} on the Open Image Dataset~\cite{oid}.}
    \label{fig:oidmotion}
\end{figure*}

To create a diverse dataset of degraded images, we simulated camera shake blur as described in~\cite{shakeblur}. This involves generating random blur kernels with a range of intensities and sizes, which were then applied to high-quality images from the Open Image Dataset~\cite{oid} to simulate per-object motion blur.  We further degraded these images by introducing lens blur (using Gaussian blur kernels), shot noise, read-out noise, and JPEG compression. By randomly sampling the parameters for each degradation, we created a dataset that encompasses a wide spectrum of image quality, from heavily degraded to almost no degradation.
Some OID-Motion sample images are shown in Fig.~\ref{fig:oidmotion}.

\end{document}